\documentclass{article}

\usepackage{microtype}
\usepackage{graphicx}
\usepackage{caption}
\usepackage{subcaption}
\usepackage{booktabs} 
\usepackage{booktabs} 

\usepackage{enumitem}
\usepackage{hyperref}
\usepackage{amsmath}  
\usepackage{svg}
\usepackage{makecell}
\usepackage{amssymb}
\usepackage{svg}
\usepackage{tcolorbox}
\usepackage{color}
\usepackage{xcolor}

\newcommand{\x}{{\mathbf{x}}}
\newcommand{\y}{{\mathbf{y}}}
\newcommand{\z}{{\mathbf{z}}}
\newcommand{\m}{{\mathbf{m}}}
\newcommand{\pr}{{\mathbf{p}}}
\newcommand{\A}{{\mathbf{A}}}
\newcommand{\e}{\mathbf{\epsilon}}
\newcommand{\p}{\boldsymbol{\theta}}
\newcommand{\vr}{\boldsymbol{v}}
\newcommand{\F}{{\mathcal{F}}}

 \newcommand{\Se}{{\mathbf{S}}}
\newcommand{\M}{{\mathbf{M}}}

\usepackage[accepted]{icml2024}

\usepackage{amsmath}
\usepackage{amssymb}
\usepackage{mathtools}
\usepackage{bm}
\usepackage{amsthm}

\usepackage[capitalize,noabbrev]{cleveref}

\theoremstyle{plain}

\theoremstyle{definition}

\theoremstyle{remark}

\usepackage[textsize=tiny]{todonotes}

\icmltitlerunning{Optimal Eye Surgeon}

\begin{document}

\twocolumn[
\icmltitle{Optimal Eye Surgeon: Finding image priors through sparse generators at initialization}



\icmlsetsymbol{equal}{*}

\begin{icmlauthorlist}
\icmlauthor{Avrajit Ghosh}{yyy}
\icmlauthor{Xitong Zhang}{yyy}
\icmlauthor{Kenneth Sun}{comp}
\icmlauthor{Qing Qu}{comp}
\icmlauthor{Saiprasad Ravishankar}{yyy,bb}
\icmlauthor{Rongrong Wang}{yyy}
\end{icmlauthorlist}

\icmlaffiliation{yyy}{Department of Computational Mathematics, Science and Engineering (CMSE), Michigan State University, MI, USA.}
\icmlaffiliation{bb}{Department of Biomedical Engineering, Michigan State University, MI, USA}
\icmlaffiliation{comp}{Department of Electrical Engineering and Computer Science (EECS), University of Michigan – Ann Arbor, MI, USA.}

\icmlcorrespondingauthor{Avrajit Ghosh}{ghoshavr@msu.edu}

\icmlkeywords{Machine Learning, ICML}

\vskip 0.3in
]



\printAffiliationsAndNotice{}  

\begin{abstract}

   We introduce \textit{Optimal Eye Surgeon} (OES), a framework for pruning and training deep image generator networks. Typically, untrained deep convolutional networks, which include image sampling operations, serve as effective image priors \citep{ulyanov2018deep}. However, they tend to overfit to noise in image restoration tasks due to being overparameterized. OES addresses this by adaptively pruning networks at random initialization to a level of underparameterization. This process effectively captures low-frequency image components \textit{even without training, by just masking}. When trained to fit noisy image, these pruned subnetworks, which we term \textit{Sparse-DIP}, resist overfitting to noise. This benefit arises from underparameterization and the regularization effect of masking, constraining them in the manifold of image priors (Figure-\ref{fig:concept_diag}). We demonstrate that subnetworks pruned through OES surpass other leading pruning methods, such as the Lottery Ticket Hypothesis, which is known to be suboptimal for image recovery tasks~\citep{wu2023chasing}. Our extensive experiments demonstrate the transferability of OES-masks and the characteristics of sparse-subnetworks for image generation. Code is available at \href{https://github.com/Avra98/Optimal-Eye-Surgeon.git}{https://github.com/Avra98/Optimal-Eye-Surgeon.git}.
\end{abstract}

\begin{figure}[h]
  \centering
    \centering
     \includegraphics[width=\linewidth]{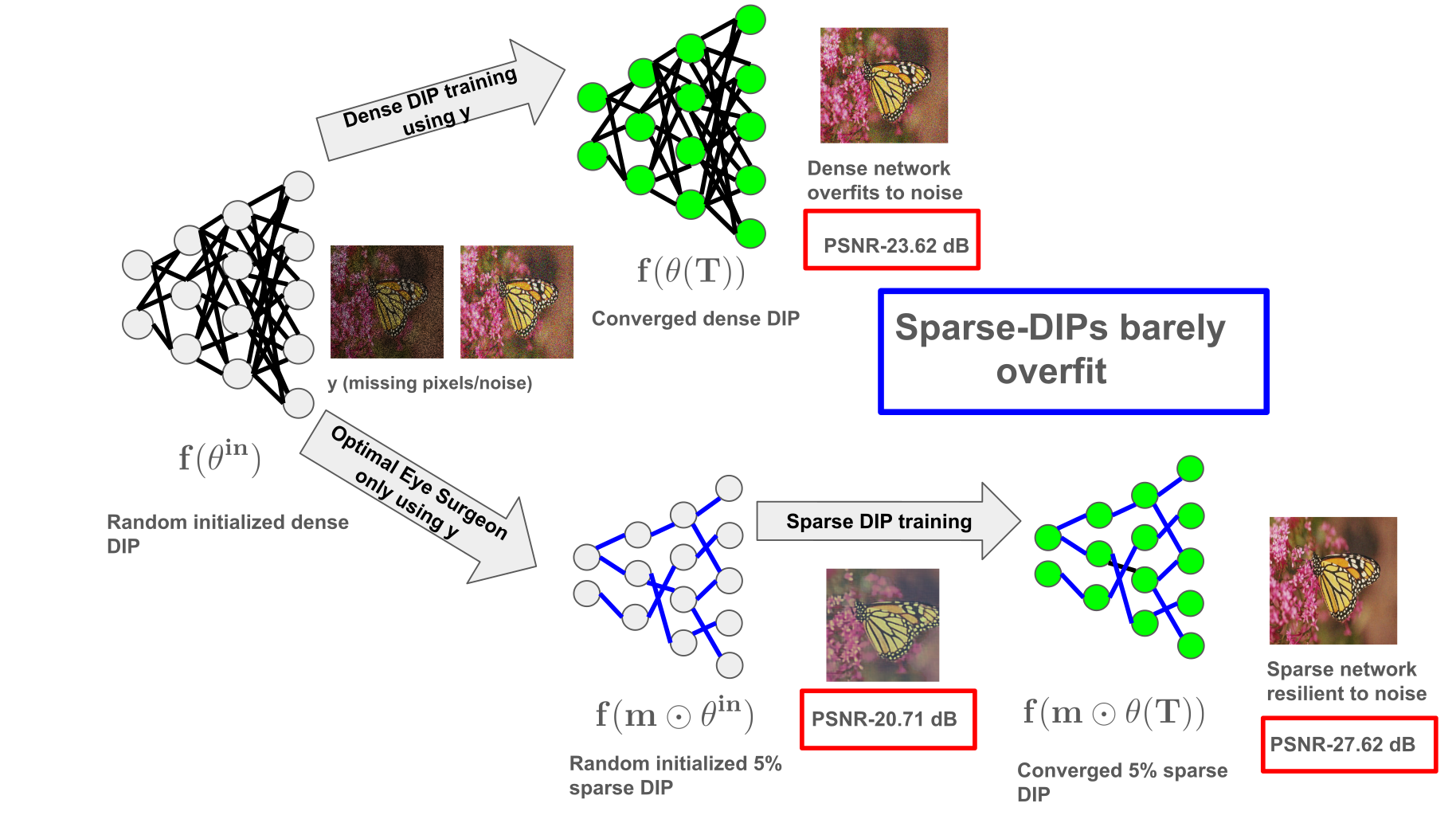}
    \caption{Sparse-DIP lessens overfitting}
    \label{fig:flow} 
\end{figure}  
\begin{figure}[h]
  \begin{minipage}[t]{0.48\textwidth}
    \centering
    \includegraphics[width=1\linewidth]{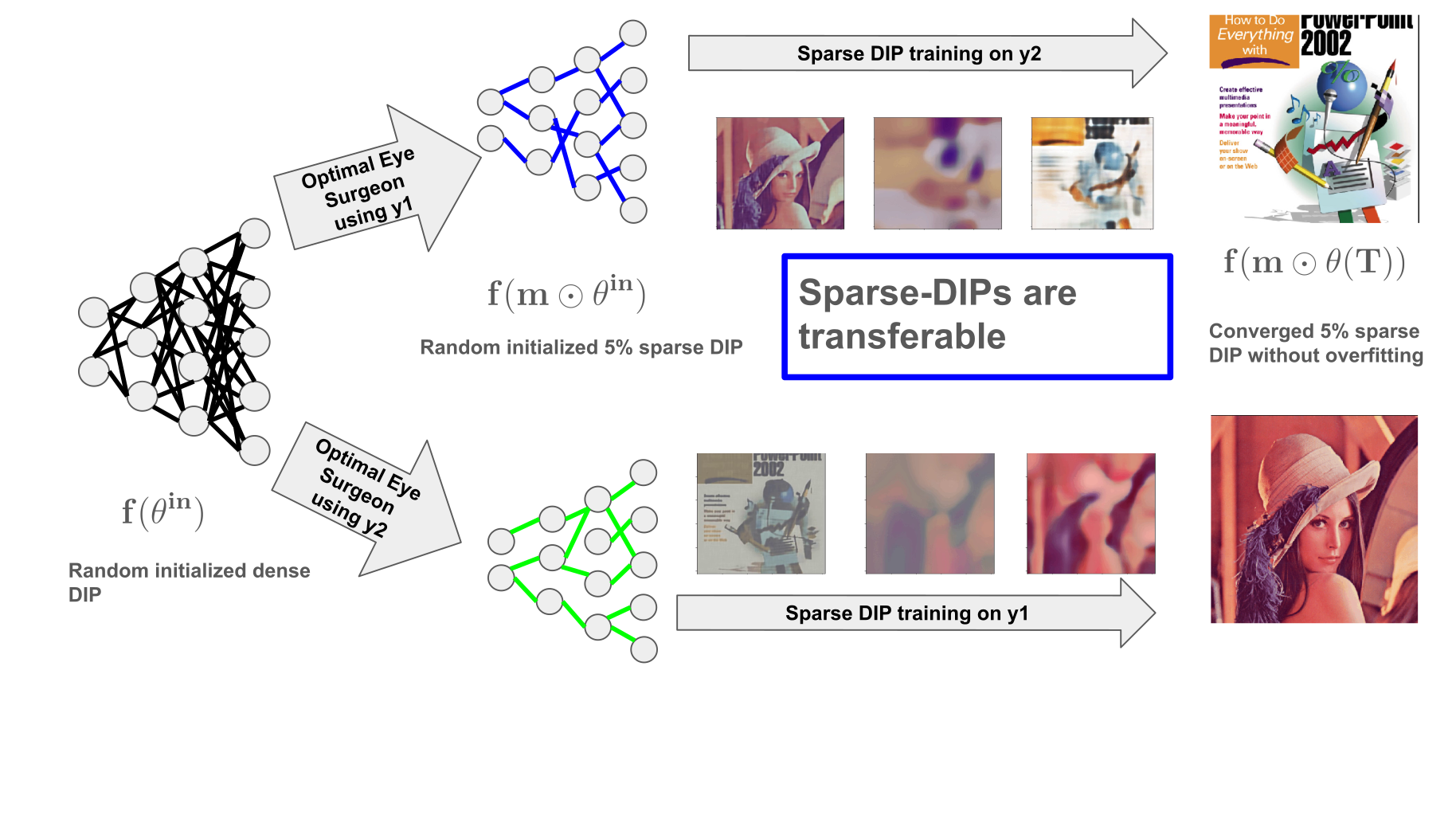}
    \vspace{-0.4in}
    \caption{Sparse-DIPs are transferrable}
    \label{fig:transfer}
  \end{minipage}
\end{figure}
\vspace{-0.3in}
\section{Introduction}

Overparameterization has been central to the success of deep learning especially in image classification tasks. Empirically it is observed that bigger models (at scale) generalize better. Eventually, it was found that sufficiently sparse subnetworks can be found within these deep dense networks that can reach as high test accuracy as their dense counterparts. These sparse networks are called matching subnetworks. This led researchers to further study neural network pruning.
However, the impact of overparameterization in deep convolutional neural networks (CNNs) hasn't been thoroughly studied for image reconstruction and inversion tasks although overparameterization is important in many image recovery tasks. \citet{jin2017deep} empirically showed that trained deep CNNs are better substitutes to regularized iterative algorithms and direct inversion~\citep{katsaggelos1989iterative}. 
The initial works further led to deploying deep convolutional networks inside the typical iterative image reconstruction framework, where it is fused with the physics or the forward model of the image generation problem \citep{venkatakrishnan2013plug}. \\
       Going one step ahead, \citet{ulyanov2018deep} showed that \textit{untrained} deep convolutional networks can recover images directly from corrupted measurements. Hourglass architectures like Unet/Skipnet having downsampling, upsampling and convolutional operations are natural image priors, as they bias the output towards the prior distribution of natural images. These networks are known as Deep Image Prior (DIP). When trained to reconstruct a corrupted image, DIPs first learn the natural image component of the corrupted image and then overfit to the noise as they are highly overparameterized. This phenomenon is known as spectral bias 
 \citep{chakrabarty2019spectral}. Hence, some early stopping time criteria need to be adopted before these models overfit to the noise or artifacts in the image. Finding an estimate of early stopping time typically requires knowledge of the clean image and noise-corruption level, which are usually unknown, making this an active area of research \citep{wang2021early}. 

Underparameterized models\footnote{In image restoration, underparameterized models are defined as networks with fewer parameters than the number of image pixels. So, the image fit loss $\| G(\p,\z) -\y\|_{2}^2$ may not be zero at convergence.} emerged as a good substitute to deep Unets as means to prevent overfitting. \citet{heckel2018deep} proposed deep decoder which consists of only upsampling layers and convolutional layers with kernel size $1 \times 1$. Deep decoders prevent overfitting to a large extent but as they are sufficiently underparameterized, they are not rich image priors. They fail to capture detailed image information \citep{wu2023chasing}.  \\
In this work, we bridge this gap between overparameterized models like deep image prior and underparameterized models like deep decoder. We aim to find a sparse sub-network within a dense DIP network that can act as an image prior and doesn't overfit to noise because of underparameterization. Our main contributions are as follows:

\begin{enumerate}
    \item We propose a principled approach of pruning a deep image prior network at \textit{ random initialization} with only the \textit{corrupted measurement for a single image} and train the pruned network till convergence (Figure-\ref{fig:flow}). 
    \item We show that the masked subnetwork output gives a low frequency approximation of the clean image by just masking at random network initialization. Further training these subnetworks to convergence alleviates overfitting. 
    \item We show that these sparse networks are transferable, i.e., masks learned on one image are transferable for recovering a different image (Figure-\ref{fig:transfer}). 
\end{enumerate}
 
\vspace{-0.1in}
\section{Image reconstruction with DIP}
The general framework for image reconstruction involves corrupted measurements $\y$ produced from a clean image $\x$ undergoing a corruption process $\y = \A(\x) + \e$, where $\A(.)$ represents the corruption operation and $\e$ is a noise vector drawn from any standard normalized  
distribution (e.g., Gaussian). The objective is to determine $\x$ given $\y$. Image reconstruction entails finding the MAP (Maximum A Posteriori) solution, which maximizes the posterior distribution $p(\x|\y) \propto p(\y|\x)p(\x)$. Assuming Gaussian noise, the likelihood term $p(\y|\x)$ focuses on minimizing $||\y - \A(\x)||_{2}^2$ to identify the optimal fit. However, since the forward operator $\A(.)$ typically has a large null space, making the inverse problem ill-posed, additional insight into the prior distribution $p(\x)$ is required. 

Deep image prior proposed by \citet{ulyanov2018deep} showed that by reparameterizing the reconstruction variable $\x$ as the output of an untrained deep Unet $\x = G(\p,\z)$, we can regularize the solution-space of the output to look like natural images. For example, $G(.)$ denotes the hourglass convolutional architecture, $\p$ are the model parameters and $\z$ is a random input to the network. Here, the image prior is implicit, as the output space of $G(\p,\z)$ inherently encapsulates the unique characteristics of a natural image. For image denoising, we minimize the loss  $||\y - G(\p,\z)||_{2}^2$ w.r.t network parameters $\p$, the target of the network being the corrupted image $\y$. Early in the training, the deep Unet architecture regularizes solutions towards natural images, giving an estimate of the clean image. However, as the model is highly overparameterized, the training loss $||\y - G(\p,\z)||_{2}^2$ will be driven to 0, essentially ensuring $G(\p,\z)$ fits the noisy image $\y$. Hence, some early-stopping strategy is required to obtain the clean image, which is difficult without the knowledge of the ground-truth clean image $\x$. 

Several works in recent years have approached the challenge of finding the early-stopping time or preventing overfitting to noise, which broadly falls into two classes as discussed next. 
\subsection{Through regularization}
\citet{cheng2019bayesian} considers a Bayesian approach to inference, by conducting posterior inference using stochastic gradient Langevin dynamics which delays overfitting. \citet{jo2021rethinking,shi2022measuring,metzler2018unsupervised} control the deep network capacity by regularizing the layer-weights or the Jacobian of the network. These methods incur an additional computational and backpropagation cost. \citet{liu2019image,mataev2019deepred,sun2020solving,cascarano2021combining,10096631} use additional regularizers on the deep, dense models such as the total-variation norm or trained denoiser or external guidance. These methods require the right regularization level which depends on the noise-type, level, and image class to avoid overfitting. \citet{you2020robust} model sparse additive noise as an explicit term in the optimization. \citet{ding2021rank} explore subgradient methods with diminishing step size schedules for impulse noise with $\ell_{1}$ loss. These methods are limited to the types and the levels of noise they target. Finally, \citet{wang2021early} develop a general-purpose early-stopping detection criterion for all of the above methods. Their approach to detecting the transition from clean to noisy reconstruction is by estimating the running variance of the reconstructed image over the iteration window. However, in certain cases, their detection peak is sometimes off by certain iterations, as acknowledged by the authors. All of these works attempt to avoid overfitting while optimizing overparameterized dense models which incurs additional cost on storage and computational time. 

\subsection{Through underparameterization}
On the contrary, the performance of under-parameterized networks for image recovery is significantly less approached. \citet{heckel2018deep} first proposed Deep-decoder, an underparameterized network consisting only of the decoder part of the Unet architecture. Underparameterization provides a barrier to overfitting, allowing the deep decoder to denoise without much overfitting. However, due to the same reason, deep decoders slightly underperform when the underlying  ground-truth image has fine-grained texture details. Hence, for images with rich detail information, deep decoders underperform. However, their ability to prevent overfitting for most denoising problems makes them attractive for image restoration problems compared to typical DIP networks and their variants. 
The recent success of under-parameterized networks like deep decoder motivates the question:\vspace{-0.2in}
\paragraph{Q1}\label{q1}: \textit{Can under-parameterization prevent overfitting and at the same time recover high-quality images?}
If the answer to question Q1 is yes, then the next question is how to build these underparameterized networks. As a first step to this question, we start with an overparameterized Unet and attempt to study a principled pruning strategy to obtain an underparameterized network. Thus, we study the second and more interesting question: \vspace{-0.2in}
\paragraph{Q2}\label{q2}: \textit{Can we design a principled pruning method with only the corrupted measurements $\y$ to obtain an underparameterized network that satisfies Q1? }
Our answers to both questions are positive and our findings reveal some interesting phenomena on overparameterization, initialization and their relation to capturing image priors.  



\vspace{-0.1in}
\section{Optimal Eye Surgeon: Pruning image generators at initialization}

Neural network weight pruning dates back to as early as the early 90's \citep{lecun1989optimal,hassibi1993optimal}. Pruning can be broadly classified into three classes based on when networks are pruned: 
1) \textit{Pruning at Initialization (PAI)} methods prune deep networks at random initialization. The resultant sparse sub-network at initialization is then trained to convergence at inference time. Pruning at Initialization (PAI) techniques, like SNIP \citep{lee2018snip}, GraSP \citep{wang2020picking}, and SynFlow \citep{tanaka2020pruning}, focus on effective weight pruning in neural networks at random initialization. SNIP removes weights minimally impacting loss, GRASP preserves information flow, and SynFlow, a data-free method, maintains total synaptic flow under sparsity constraints. Our proposed method falls under this category. 
2) \textit{Pruning while Training (PWT)} takes a randomly initialized dense network and jointly trains and prunes a neural network by updating weights and masking the weights during training. Different strategies can be adopted for determining importance scores like random dropout, magnitude, or back-and-forth pruning \citep{evci2020rigging,zhao2019variational,he2018soft}. The benefits of pruning early in training were also shown in \citet{you2019drawing}. 3) \textit{Pruning After Training (PAT)} involves a Pretrain-Prune-Retrain cycle and is essential for obtaining matching subnetworks at non-trivial sparsity levels. The Lottery Ticket Hypothesis (LTH) \cite{frankle2018lottery} advocates for Iterative Magnitude Pruning (IMP), which removes a percentage of weights based on the magnitude from a pretrained network, then retrains the remaining weights from their original initialization. For complex networks and large datasets, weight rewinding to an early-epoch state \citep{frankle2019stabilizing} and learning-rate rewinding \citep{renda2020comparing} were deemed essential to obtain matching subnetworks. The weight magnitudes at the end of training are crucial, as highlighted in ongoing research \citep{paul2022unmasking}. 

Overparameterization seems to be a crucial factor for finding sparse matching subnetworks. \citet{zhou2019deconstructing,ramanujan2020s} showed that when a network is sufficiently large, even learning a mask at random initialization (termed as \textit{supermasks}) can show competitive performance like training a network. This phenomenon is termed as strong lottery ticket hypothesis, and was recently proved by \citet{malach2020proving,da2021proving} under certain network assumptions. Supermasks were also used to generate different subnetworks for various tasks from the same dense network \citep{wortsman2020supermasks,mallya2018piggyback}, with applications also in graph networks \citep{huang2022you}. Our work is the first to show the existence and effectiveness of supermasks for image reconstruction. We further highlight the notable diffrences of pruning for image classification and image reconstruction in Table-\ref{comparingclassvsrec}.

\vspace{-0.1in}

\subsection{Suboptimality of LTH for DIP}
\label{subopt_lth_dip}
\begin{figure}[htb]
  \centering
  \includegraphics[width=\linewidth]{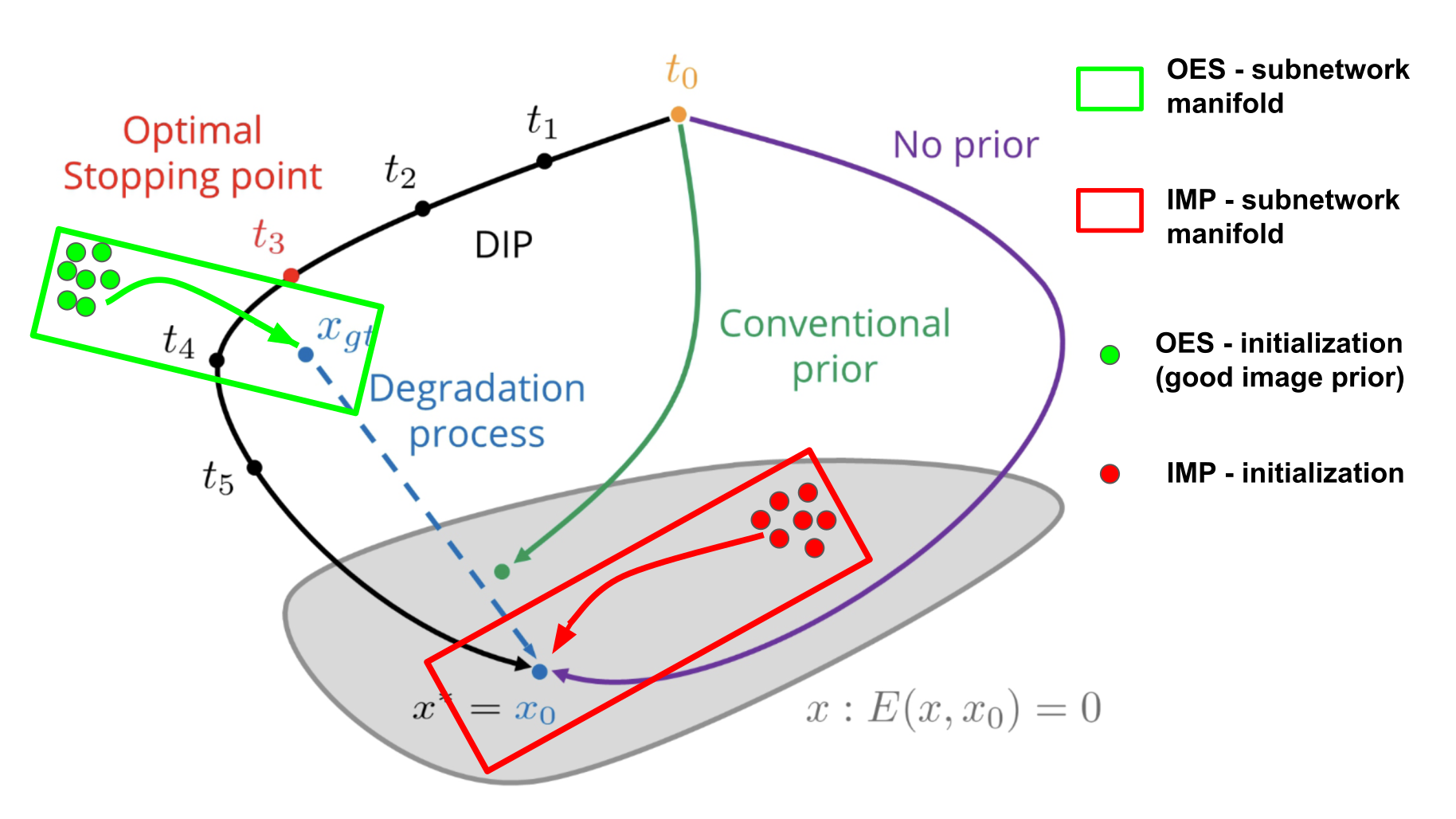}
  \caption{Subnetworks learned by OES are image generators with good image priors. On the contrary, the range of subnetworks learned by LTH is close to the overfitted noisy image, which is far from being an image prior. Image adapted from \cite{ulyanov2018deep}}
  \label{fig:concept_diag}
\end{figure}
LTH-based methods are very reliable to obtain matching sparse subnetworks at non-trivial sparsities for various ML tasks, a feat unachieved by other pruning methods. Given the success of LTH on a variety of machine learning tasks, at first-sight, it might be tempting to apply LTH on image reconstruction based tasks involving deep image prior. However, for unsupervised learning schemes like \textit{DIP which overfit to noise at convergence, using the magnitudes at convergence, to determine which weights to prune, is in fact detrimental (Figure-\ref{fig:concept_diag}).} 

 \begin{tcolorbox}
For image reconstruction, network output overfits to noise at convergence. Subnetworks obtained by LTH at convergence (without early stopping time) perform poorly on denoising tasks. 
\end{tcolorbox}
\begin{figure}[h]
    \centering
    \includegraphics[width=0.4\textwidth]{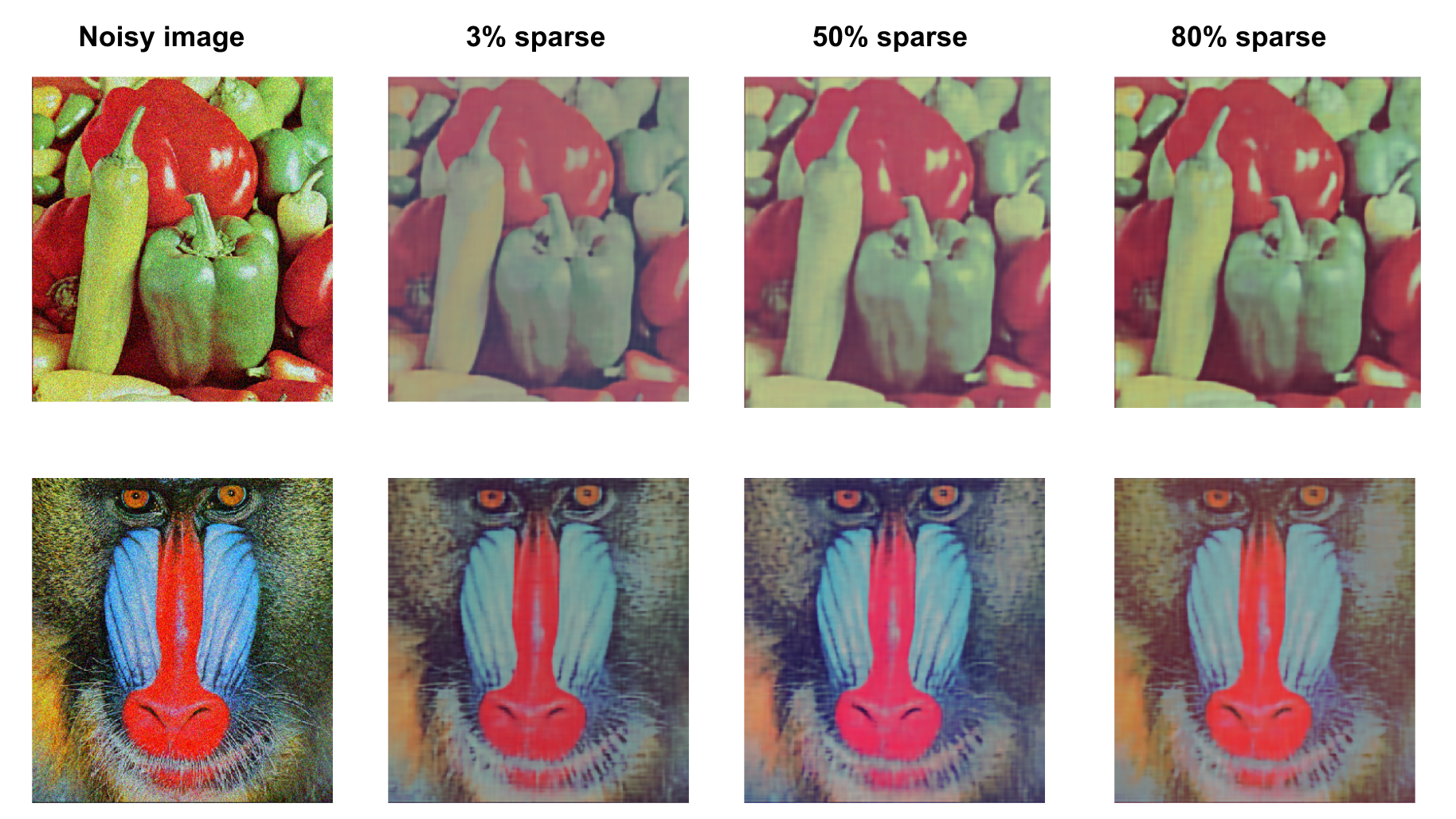}
    \caption{Masking network at initialization induces image prior. Figures show the images after masking image generator parameters at random-initialization $G(\p_{in} \circ \m^{*}(\y),\z)$. The mask $\m^{*}$ was learned using the OES algorithm. Images corresponding to several sparsity levels are shown. We show that Strong Lottery Ticket Hypothesis also holds for image reconstruction. }
    \label{fig:set-14_masked}
\end{figure}

Two possible ways to apply LTH to DIP for image reconstruction tasks are: a) using the clean image $\x$ to train the DIP model which will not require any early-stopping and b) the early stopping (ES) time can be obtained from the knowledge of $\x$ and the weight magnitude at ES can be used to obtain the mask. \citet{wu2023chasing}, adopted method a) to obtain the mask, which might not be practical for most image reconstruction problems (see Section \ref{lottery_image} for detailed comparison) as we do not have knowledge of the clean image $\x$ nor an assumption of an early-stopping time (Figure-\ref{fig:early_var}). We show the effect of using LTH-based methods (with loss involving $\y$) as the mask in Figure-\ref{impeffects} and in Section \ref{imp-den} (Appendix). Further, we study in detail, the architecture of the pruned network derived from LTH pruning in Figure-\ref{fig:subhist1} which sheds light on why IMP-masks may underperform in image reconstruction tasks. In our work, we propose Optimal Eye Surgeon (OES), a framework to prune image generators at random initialization which is optimal for pruning image generator networks. 
\vspace{-0.1in}
\subsection{Masking at initialization}
Let $G(\p,\z)$ be a dense and deep image generator network with random input $\z$. Let the random input $\z \in \mathbb{R}^q$, and let $\p$ be vectorized parameters of a dense Unet, $\p \in \mathbb{R}^d$. Let $\x$, $\y$ be the clean and noisy/corrupted RGB image such that $\x,\y \in  \mathbb{R}^{3\times H \times W}$, where $H$ and $W$ are respectively the height and width of the image. As DIPs are sufficiently overparameterized, i.e., the number of parameters is much more than the number of image pixels, it is usually the case that $d \gg 3HW$. Let $\p_{in}$ be the random initialized neural network, where the uniform Kaiming initialization is used. Let $\m \in \{0,1\}^d$ be the binary mask that we aim to learn at initialization. To learn an s-sparse mask, i.e., with only $s$ non-zero parameters out of $d$, we would have to solve an integer problem:
\begin{align}
\label{MI}
    \m^{*}(\y) &= \arg \min_{\m \in \{0,1\}^d } || G(\p_{in} \circ \m,\z) - \y||_{2}^2 \nonumber \\
    &\quad \text{such that} \quad ||\m||_{0} \leq s. 
\end{align}
Equation~\eqref{MI} involves discrete optimization for deep networks, where $d$ is very large (in millions). To get around this difficulty, we propose a Bayesian relaxation of \eqref{MI} that is differentiable and unconstrained and can be solved by a local iterative algorithm such as gradient descent. We attempt this by reformulating \eqref{MI} as learning Bernoulli dropout probability parameters $\pr$ with the mask $\m$ being sampled from the Bernoulli distribution with mean $\pr \in \mathbb{R}^d$. 
\vspace{-0.3in}

\begin{equation}
\begin{aligned}
\m^{*}(\y) &= C(\pr^{*})  \quad \text{such that} \quad \\
& \pr^{*} = \arg \min_{\pr} \underbrace{\mathbb{E}_{\m \sim Ber(\pr) }  \left[ || G(\p_{in} \circ \m,\z) - \y||_{2}^2 \right]}_{R(\pr)} \\
& + \lambda KL(Ber(\pr)||Ber(\pr_{0})).
\end{aligned}
\label{ber_mask}
\end{equation}

\vspace{-0.1in}
The deterministic inequality constraint $||\m||_{0} \leq s $ is changed into an unconstrained penalty which ensures that the learned Bernoulli distribution $Ber(\pr)$ is close to a prior Bernoulli distribution $Ber(\pr_{0})$, the known prior distribution depends on the desired sparsity level $s$. We fix $p_{0} =\frac{s}{d}$. For Bernoulli distributions, the distance measure between two distributions as given by Kulbick-Luiber divergence has a closed form and is given by $KL(Ber(\pr)||Ber(\pr_{0}(s))) = \sum_{i} \left( p_{i} \log \frac{p_{i}}{p_{0i}} + (1 - p_{i} ) \log \frac{1 - p_{i}}{1 - p_{0i}} \right)$, where $p_{i}$ and $p_{0i}$ denotes the Bernoulli mean probability corresponding to the $i^{th}$ weight parameter of $\mathbf{p}$ and $\mathbf{p}_0$. We solve this optimization problem by learning $\pr$ via the Gumbel-softmax trick. We delay the details of the algorithm to the Appendix section \ref{gumb_softmax_det}. After obtaining the converged $\pr$, we prune the weights based on the ranking/ordering of $\pr$ to obtain the desired sparsity level, which is denoted by the $C(.)$ function. We discuss the importance of KL regularization compared to $L_{1}$ regularization \cite{sreenivasan2022rare} or no regularization \cite{zhou2019deconstructing} in Section \ref{l1reg} of Appendix. Previous work on Bernoulli mask learning and pruning on network initialization only focused on image-classification tasks, whereas our work applies it to image-reconstruction tasks and develops many new findings that might provide important insight into new structure design for DIP.

\begin{figure}[h]
    \centering
    \includegraphics[width=0.6\textwidth]{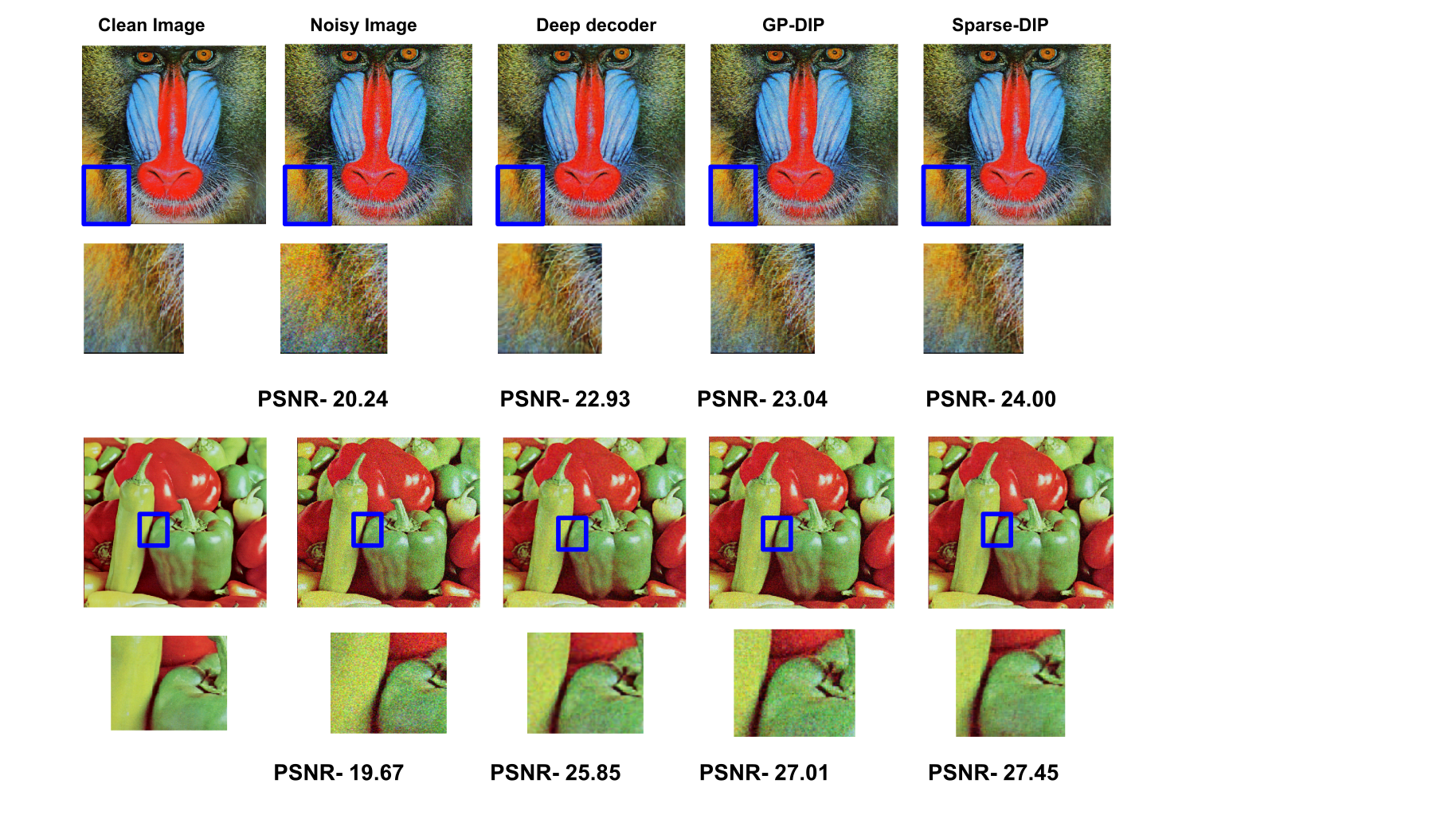}
      \caption{Comparative Analysis of Denoising Performance on 'Baboon' and 'Pepper' Images at $\sigma=25$ dB. }
    \label{fig:baboon_pepper}
\end{figure}

 

Our proposed OES (Optimal Eye Surgeon) algorithm consists of two steps:
\begin{itemize}[leftmargin=*,noitemsep,topsep=0pt,parsep=0pt,partopsep=0pt]
\item Solve the optimization problem in \eqref{ber_mask} to learn the mask $\m^{*}(\y)$ using OES.
\item Train the sparse subnetwork $G(\p \circ \m^{*}(\y),\z)$ to convergence to fit the corrupted image $\y.$
\end{itemize}

We summarize the important observations from applying our algorithm to DIP for image reconstruction. These findings will be supported by extensive numerical experiments in the next section and appendix.
\begin{itemize}[leftmargin=*,noitemsep,topsep=0pt,parsep=0pt,partopsep=0pt]
 \setlength\itemsep{0.25em}
    \item \textbf{Finding-1}: Masks learned by Step 1 of OES when applied at initialization induce a relatively good image prior (Figure-\ref{fig:set-14_masked}).  We term the sparse subnetwork $G(\p_{in} \circ \m^{*}(\y),\z)$ at initialization as \textit{Sparse-DIP}.  It gives a good low-frequency approximation of the clean image by just the masking network. 
        \item \textbf{Finding-2}: 
OES effectively recovers the clean image and exhibits minimal or no overfitting for denoising problems (Figure-\ref{fig:combined_denoise}).
    \item \textbf{Finding-3}: 
On image recovery tasks, the training of subnetworks identified by OES is much more effective than those discovered by the current best Pruning At Initialization (PAI) methods. Furthermore, masks created by methods based on the Lottery Ticket Hypothesis (LTH) are not ideal for reconstructing images, a point we explore in detail in Section \ref{finding-3}. 
    \item \textbf{Finding-4}: Sparse-DIPs are transferable across images, datasets and corruption processes. More specifically, a mask learned by OES from one image can be used to successfully reconstruct other images, from completely different datasets.
        \item \textbf{Finding-5}: The encoder part of DIP is more compressible (prunable) than the decoder part. (Section~\ref{arch-prune})
     \item \textbf{Finding-6} (Appendix): The irregularly pruned sparse-DIP is better than the regular deep decoder of a similar size. (Figure-\ref{set14-den-figures}).  
    \item  \textbf{Finding-7} (Appendix): Mask trained based on the initial weights is more transferrable than that based on the magnitude of the final trained weights like LTH. (Section-\ref{imp-den} in Appendix).  
\end{itemize}
\vspace{-0.1in}

\section{Experimental support of the findings}
Through extensive experiments, we confirm our findings. We use images from three popular datasets: the Set-14 dataset \citep{zeyde2012single}, the standard image dataset \cite{ulyanov2018deep} and the Face dataset \citep{bevilacqua2012low}. In Finding-1 (\ref{finding-1}), and Figure-\ref{fig:set-14_masked}, we study the quality of images that are produced by just masking. In Finding-2 (\ref{finding-2}), we compare the denoising performance of OES with overparameterized DIP, Gaussian process DIP \citep{cheng2019bayesian} and underparameterized deep decoder \citep{heckel2018deep}. In Finding-3 (\ref{finding-3}), we show results of OES against state-of-the-art pruning methods. Finally, we compare the transferability of OES and IMP across various combinations of images and datasets in Finding-4 (Section \ref{finding-4}).



\vspace{-0.1in}
\subsection{Finding-1: Masking at initialization induces image prior}
\label{finding-1}
Masking at initialization with masks learned by OES inherently captures low frequency components of the image. In Figure-\ref{fig:set-14_masked}, we display the results of $G(\p_{in} \circ \m^{*}(\y),\z)$ alongside the original corrupted image $\y$ for images in the Set-14 dataset. Images across three different levels of sparsities $3\%, 50\%$, and $80\%$ are shown. OES-masked images for other datasets are shown in Figure-\ref{fig:both-images} in the appendix. 
We observe that OES can effectively reconstruct the simpler, low-frequency parts of an image from the corrupted version $\y$, but it struggles with the more intricate details. This means that while OES can denoise an image, some information is lost in the process.
    Masking has its limitations compared to regular training. It can only represent a limited number of functions, up to $2^{d}$, where $d$ is the number of parameters in the network. Consequently, due to this limitation in function representation, the training loss with masking cannot reach zero. In our study, we found that OES primarily reconstructs the simpler, low-frequency parts of images. Since natural images usually contain more low-frequency elements, focusing on these parts allows for the greatest reduction in loss. Additionally, because the model described in \eqref{MI} lacks sufficient function representation capability, it never achieves a training loss of zero. However, even in this limited setting, OES is effective in finding a mask that represents the image $\y$ as closely as possible. For experiments in the manuscript, $\y$ is the Gaussian noise corrupted image.

\subsection{Finding-2: Sparse DIPs prevent overfitting}
\label{finding-2}
\begin{figure}[ht]
    \centering
    \begin{subfigure}[b]{0.5\textwidth} 
        \centering
        \includegraphics[width=\textwidth]{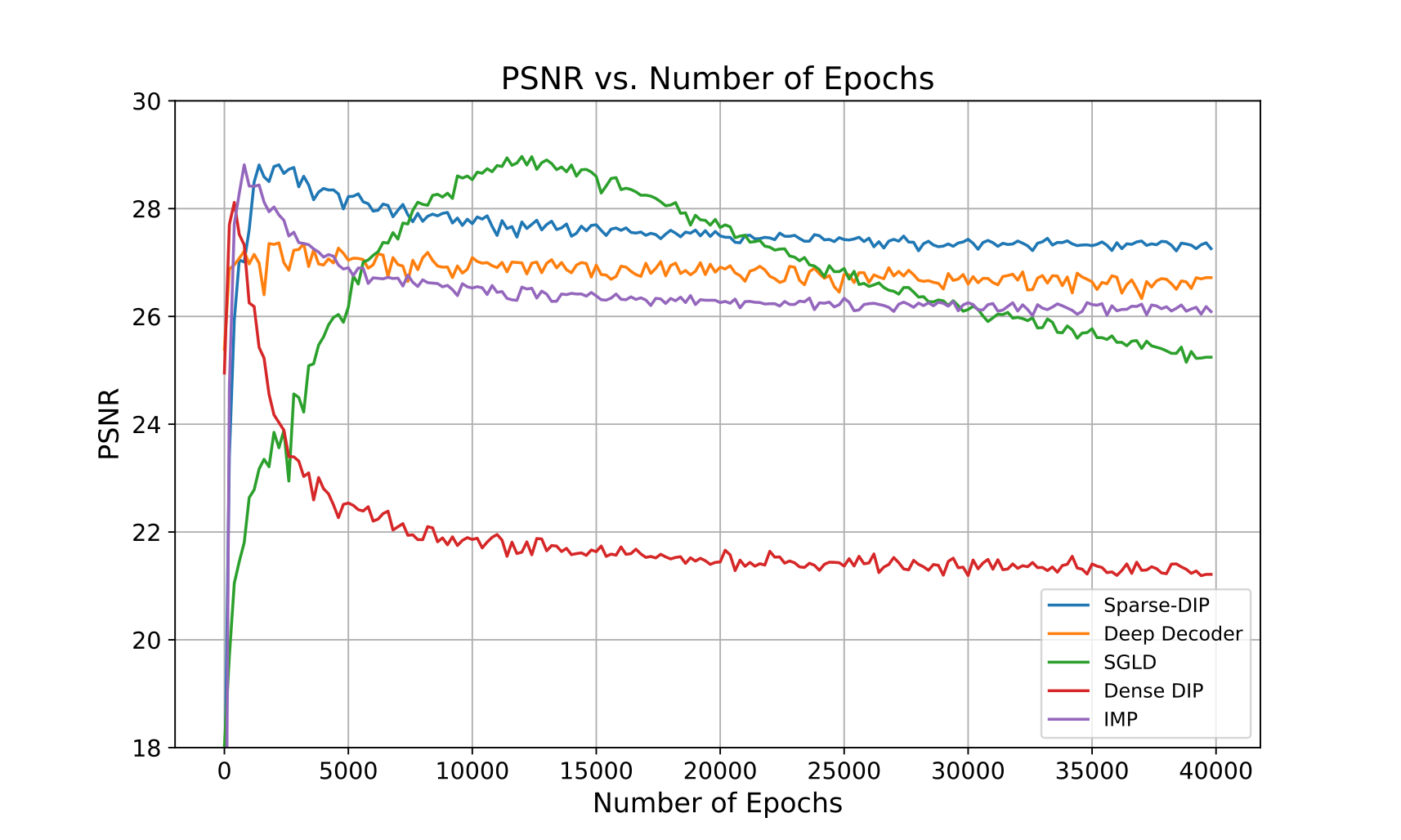}
             \caption{Pepper (Set-14 dataset)}
        \label{fig:sub1}
    \end{subfigure}
    \begin{subfigure}[b]{0.5\textwidth} 
        \centering
        \includegraphics[width=\textwidth]{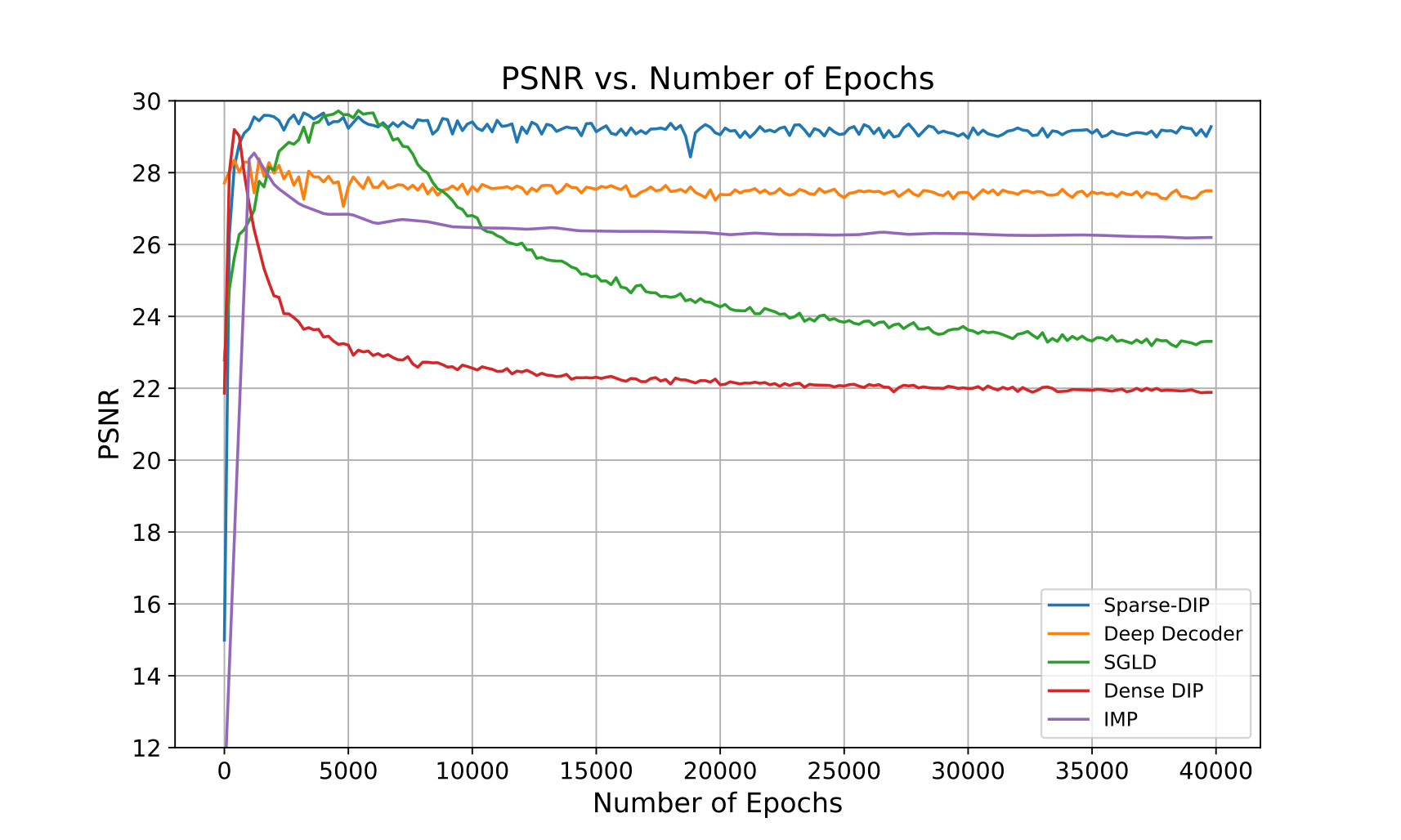}
        \caption{Door (Standard Dataset)}
        \label{fig:sub4}
    \end{subfigure}
    \caption{Denoising results of various methods on noisy images ($\sigma=25$ dB) across 3 popularly used datasets.}
    \label{fig:combined_denoise}
\end{figure}

In OES, we further train the remaining subnetwork within the obtained mask  ($G(\p_{in} \circ \m^{*}(\y),\z)$) till convergence to perform the image reconstruction task, image denoising. We conduct experiments on the denoising capabilities of these subnetworks over several noise levels and various images across 3 popularly used datasets, which we report in Figure-\ref{fig:combined_denoise}. We perform a comparison with the following enumerated network based denoising methods. 1) Dense DIP which is the overparameterized network originally proposed by \citet{ulyanov2018deep}. The encoder part has 6 layers (\textit{Conv} $\rightarrow$ \textit{ReLU} $\rightarrow$ \textit{Batchnorm} $\rightarrow$ \textit{Downsample}) followed by 6 layers of decoder (\textit{Upsample} $\rightarrow$ \textit{ReLU} $\rightarrow$ \textit{Batchnorm}). The convolution patch size in both the encoder and decoder parts is $3 \times 3$. The input $\z$ is fixed to be a random tensor drawn from the Gaussian distribution of dimension $ H \times W \times 32 \times 3$. The total number of parameters in Dense-DIP is 3008867 (3 million) and the image dimension ($\y$) is $3*512*512 = 786432$ (0.7 million). The network is overparameterized. 2) Gaussian Process-DIP (GP-DIP) is the network trained by SGLD and proposed by \citet{cheng2019bayesian} to alleviate overfitting to an extent, and 3) Deep Decoder, proposed by \citet{heckel2018deep} is an underparameterized network that prevents overfitting. Deep decoder contains only the decoder part of Unet. It has $1 \times 1$ convolution layer and upsampling layers ( $1 \times 1$ \textit{Conv} $\rightarrow$ \textit{Up sample} $\rightarrow$  \textit{ReLU} $\rightarrow$ \textit{channelnorm}). Standard decoder architecture proposed by \citet{heckel2018deep} uses channel dimension of $128$ with 6 layers as optimal denoising architecture. For this architecture, deep decoder has 100224 (0.1 million) parameters \footnote{ Further reduction of number of layers to 5 makes the denoising performance poor as mentioned by the authors and also confirmed by our experiments. We use the 6-layer deep decoder as the standard for our experiments in the paper, unless specified otherwise}. \textit{Sparse-DIP} is the pruned architecture obtained at initialization by our OES method. We perform denoising with a $3\%$ sparse subnetwork found by OES which has approximately 90217 parameters (0.09 million), slightly less than the number of parameters in deep decoder. We use the ADAM optimizer with learning rate $10^{-2}$ (as reported in \citet{ulyanov2018deep}) in all our experiments for training both the dense and sparse networks. In Figure-\ref{fig:combined_denoise} and Table-\ref{main_table_denoise}, we report the results without applying early stopping and running the optimization procedure for a large number of iterations (40k). Sparse-DIP outperforms deep decoders and the overparameterized models (with regularization) for majority of the images. In Figure-\ref{fig:baboon_pepper}, we plot the denoising results on Baboon and Pepper images. When closely zoomed in the area of focus, we observe that the deep decoder suffers from oversmoothing the edges, while GP-DIP overfits to noise due to overparameterization. We study this phenomenon in detail in Section \ref{noise-imped}. The OES framework can also be extended to general noisy inverse problem settings involving a forward operator (with a non-trivial nullspace). We extend our framework to MRI reconstruction from undersampled k-space measurements in Appendix \ref{mri_recon}. 
\vspace{-0.2in}

 \begin{figure*}[ht]
    \centering
    \begin{minipage}{0.24\textwidth}
        \centering
        \includegraphics[width=\textwidth]{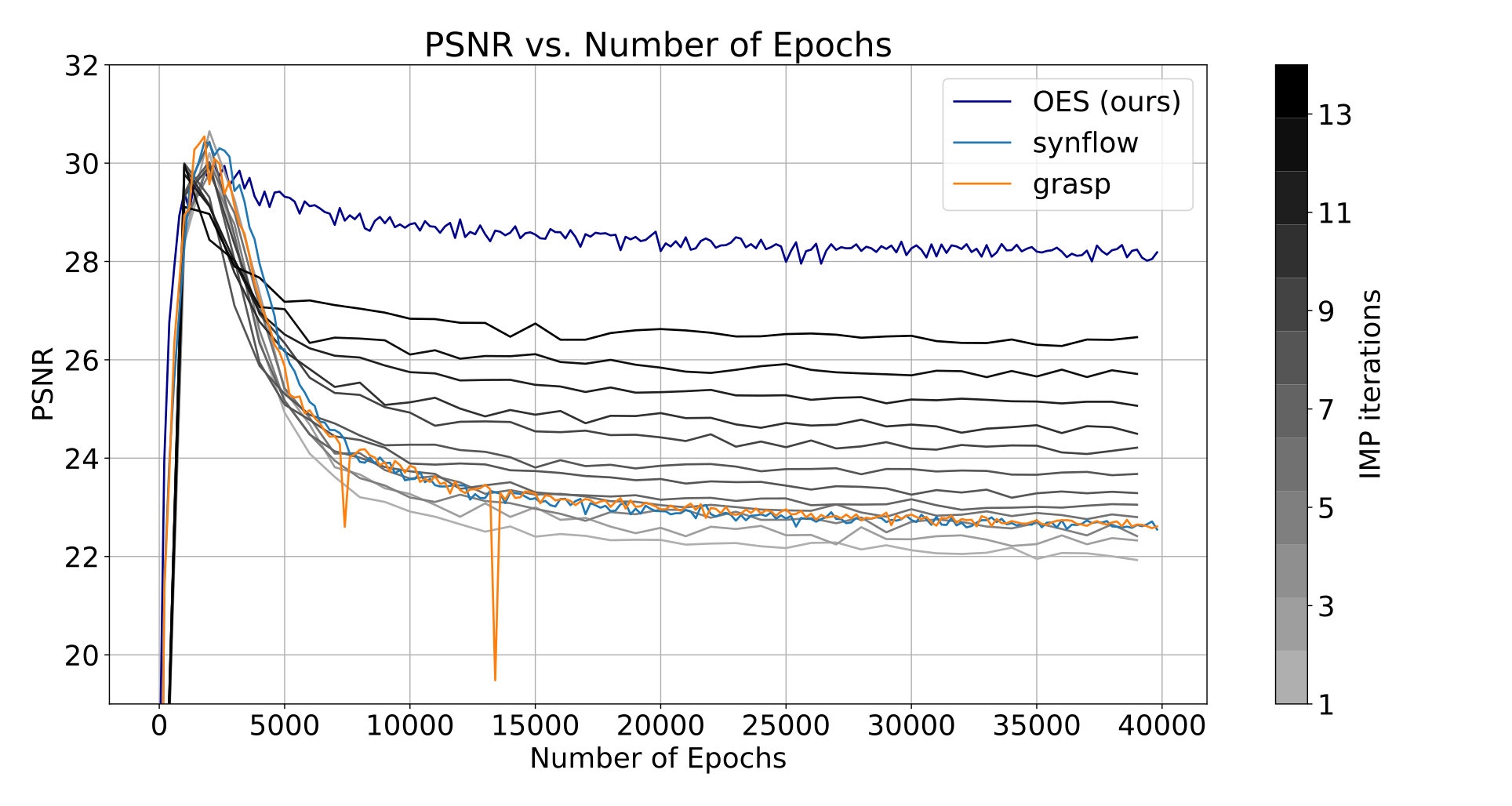}
        \subcaption{Dataset-3 (Standard dataset)}
    \end{minipage}\hfill
    \begin{minipage}{0.24\textwidth}
        \centering
        \includegraphics[width=\textwidth]{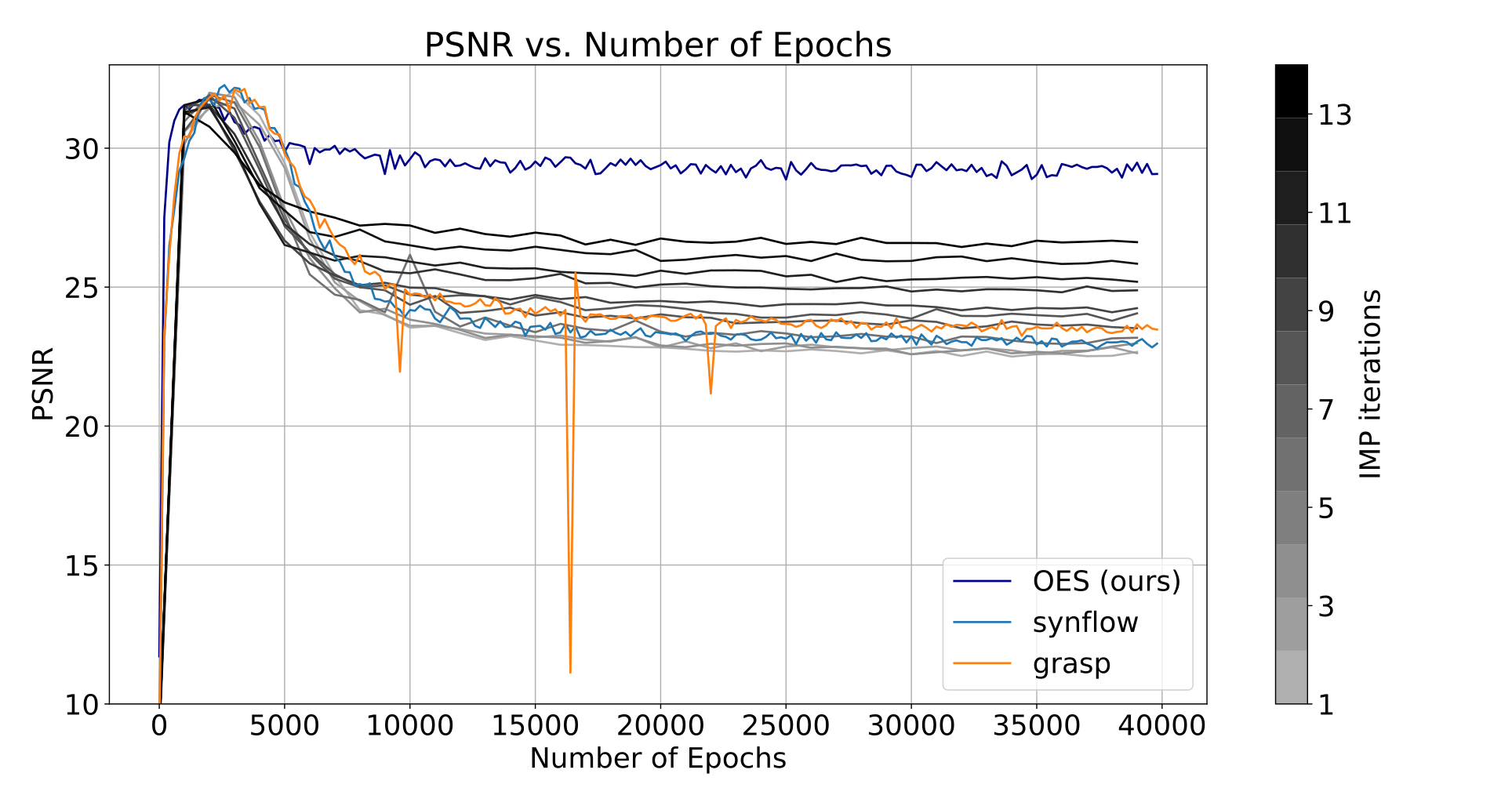}
        \subcaption{Face-1 (Face dataset)}
    \end{minipage}\hfill
    \begin{minipage}{0.24\textwidth}
        \centering
        \includegraphics[width=\textwidth]{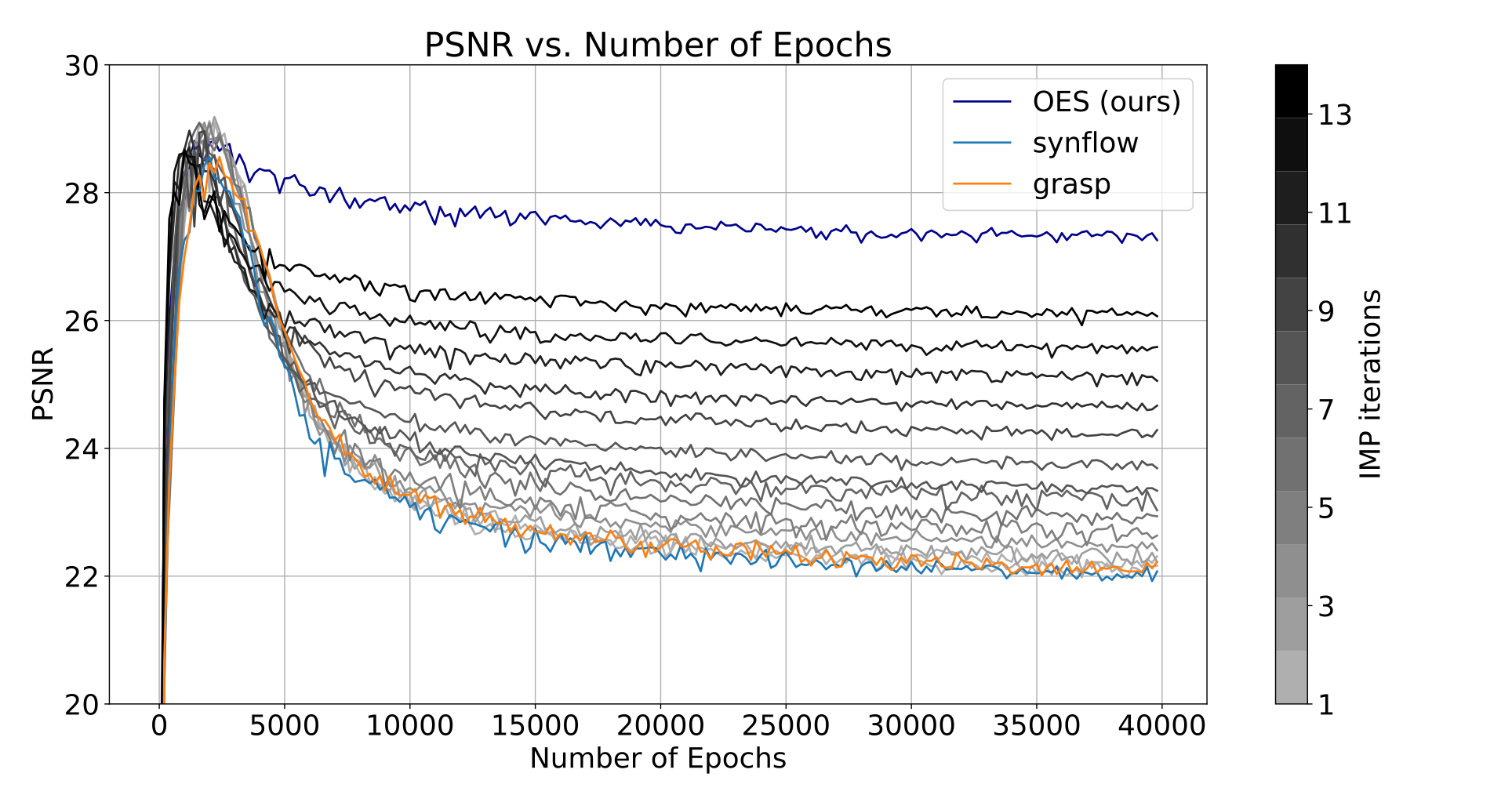}
        \subcaption{Pepper (Set-14 dataset)}
    \end{minipage}\hfill
    \begin{minipage}{0.24\textwidth}
        \centering
        \includegraphics[width=\textwidth]{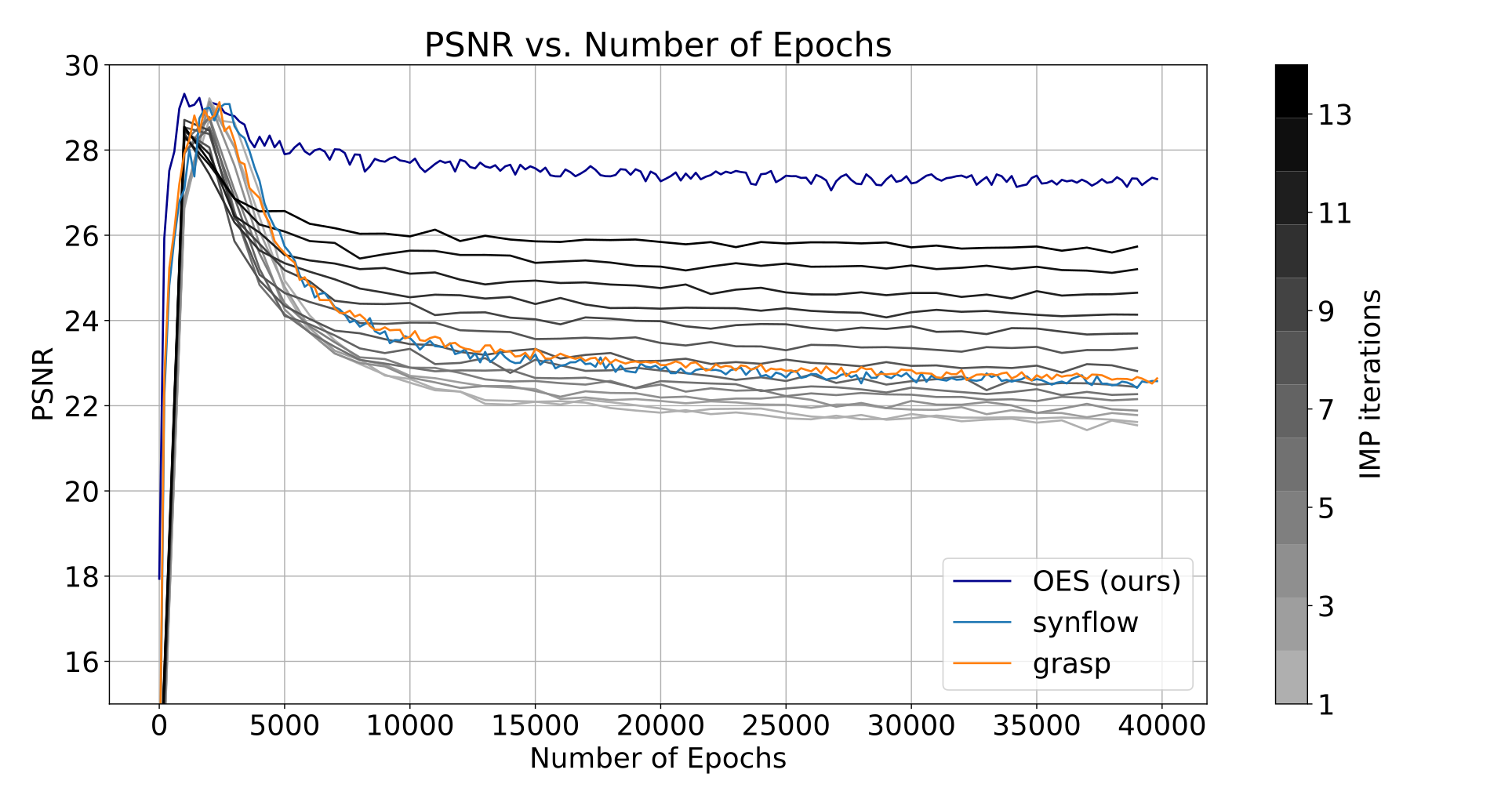}
        \subcaption{Lena (Set-14 dataset)}
    \end{minipage}
    \caption{Comparison of denoising performances for subnetworks found by various pruning methods (GRASP, Synflow, IMP, and OES). IMP utilizes 14 pruning iterations with $20\%$ weight reduction at iterations. All the masks are $5\%$ sparse. IMP undergoes an additional 14 steps of training and pruning before obtaining the final mask. The gray curves indicate the progression of IMP iterations, with darker shades representing higher iteration counts. Each IMP iteration is shown. The detailed result for all images in 3 datasets can be found in Table-\ref{pruning_methods} in the Appendix.   }
    \label{impeffects}
\end{figure*}
\vspace{-0.2in}

\begin{figure}[ht]
    \centering
    \begin{subfigure}{0.4\textwidth} 
        \includegraphics[width=\textwidth]{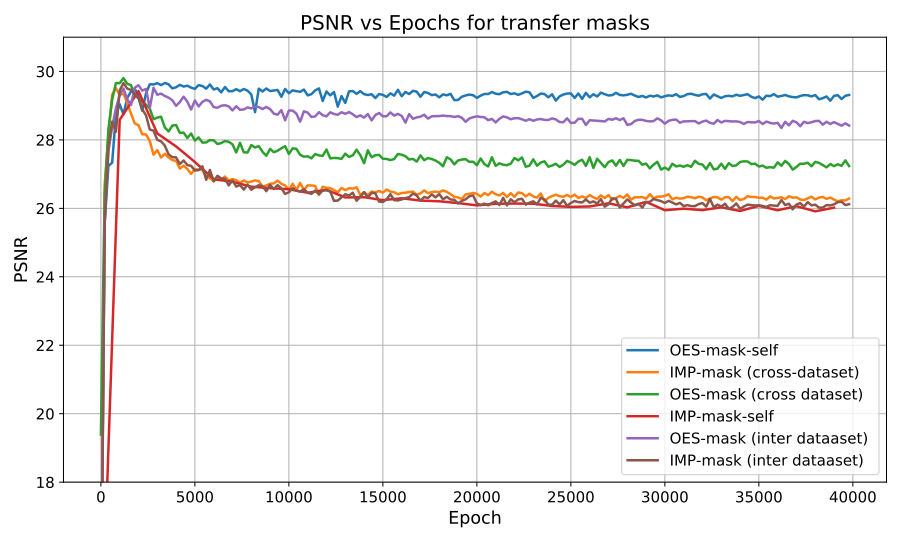}
        \caption{Building. ($\y_{target}$)}
        \label{fig:sub3}
    \end{subfigure}\hfill 
    \begin{subfigure}{0.4\textwidth} 
        \includegraphics[width=\textwidth]{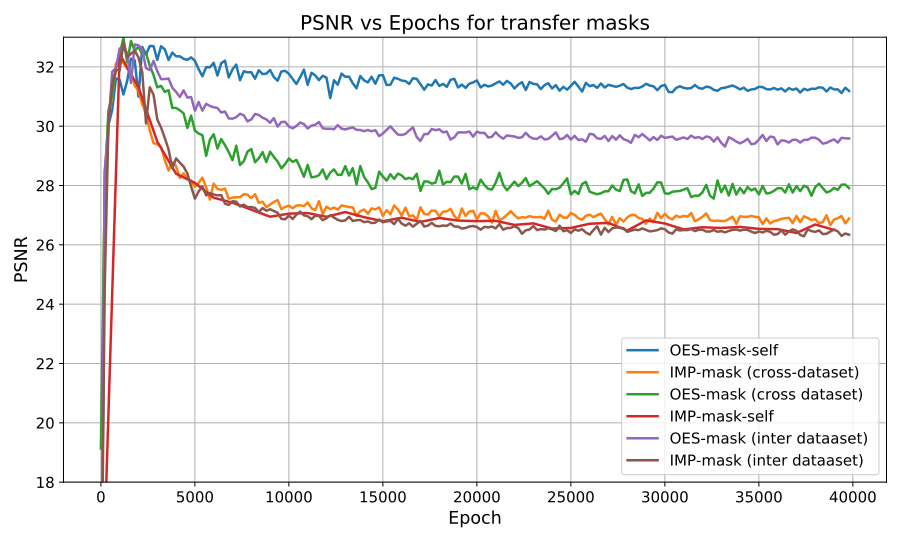}
        \caption{Face-3. ($\y_{target}$)}
        \label{fig:sub4}
    \end{subfigure}
    \caption{Performance of masks trained from several images for denoising with noise level ($\sigma=25$ dB). Self denotes a mask learned from the same image. Inter-dataset denotes mask learned from images from the same dataset. Cross-dataset denotes mask learned from images of different dataset. The standard dataset and the face dataset were used in this experiment. For Figure-a) Inter-datset mask ($\y_{source}$) is House, Cross-dataset mask ($\y_{source}$) is Face-0. For Figure-b) Inter-datset mask ($\y_{source}$) is Face-0, Cross-dataset mask ($\y_{source}$) is House. All the masks used in this figure are $5\%$ sparse.}
    \label{fig:cross_dat_combined}  
\end{figure}
\vspace{-0.1in}
\subsection{Finding-3: OES is superior to other pruning methods}
\label{finding-3}
We compare OES with the state-of-the-art pruning methods. For pruning at initialization, we compare with GraSP,  SynFlow\footnote{Performance of SNIP is not upto par with GraSP and SynFlow and hence we don't report it.}, magnitude and random scores in Table-\ref{pruning_methods}. For LTH based pruning, we report for two pruning schedules and observe that gradual pruning for larger pruning iterations yields better performance. We evaluate the denoising performance of these subnetworks at $5\%$ sparsity level. In Figure-\ref{impeffects}, we observe that at $5\%$ sparsity level, OES masks shows minimal to no overfitting. LTH masks are obtained based on ranking the magnitude of the weights at convergence (at 40k epochs) and subnetworks obtained by LTH show overfitting, when masks are at the same level of sparsity. We demonstrate the adverse effect of LTH on DIP in Figure-\ref{impeffects}, which we orignally motivated in Section~\ref{subopt_lth_dip}. However, when PSNR curves with LTH masks are plotted at every pruning iteration in Figure-\ref{impeffects}, we observe that the effect of overfitting becomes less severe when networks become more sparse. Both Synflow and Grasp show signs of overfitting for image denoising. In magnitude and random pruning methods applied \textit{at initialization}, it is often observed that layers with a large number of parameters (large width) and those with fewer parameters (small width) are respectively at a higher risk of being entirely pruned. We consistently observe that with magnitude and random pruning at initialization, at $5\%$ sparsity level, there is \textit{layer-collapse}. This phenomenon occurs when an entire layer gets pruned and the output is a constant image.

\vspace{-0.1in}

\subsection{Finding-4: OES masks are transferable}
\label{finding-4}
We perform experiments on transferring the masks obtained by Step 1 of OES on one image and show the masked subnetwork can be used for denoising a different image. We compare the transferability of the OES masks with IMP masks at the same level of sparsity ($5\%$). We also show that OES masks can be transferred not only to images within the same dataset, but also to those from a different dataset. In Figure-\ref{fig:cross_dat_combined}, we compare the denoising performance for different sets of learned masks for both IMP and OES. Say there are two image datasets: Dataset-A (face) and Dataset-B (standard dataset), each of which contains noisy images. Also, let us term the image that is used to learn the mask as $\y_{source}$ and the image on which denoising is performed as $\y_{target}$. Then we explore three possibilities: 1) \textbf{self-masking}: $\y_{source} = \y_{target}$, the same corrupted image is used to learn the mask, and the mask is used for denoising; 2) \textbf{inter-dataset masking}: $\y_{source} \neq \y_{target}$, but both $\y_{source}$ and $ \y_{target}$ belong to the same dataset; and 3) \textbf{cross-dataset masking}: $\y_{source} \neq \y_{target}$ and both of them belong to different datasets (say $\y_{source} \in $ Dataset-A and $\y_{target} \in $ Dataset-B or vice-versa). In the experiments, we use images from a standard image dataset \cite{ulyanov2018deep} and the face-dataset \cite{bevilacqua2012low} to show the extent of transferability between inter and cross datasets. We note that the images in this dataset are visually diverse as face images have different characteristics than those in the standard image dataset. We observe in most cases, self-masking by OES provides the best performance. IMP masks provide the worst performance irrespective of the source and target image. Inter-dataset masking and cross-dataset masking by OES also gives good PSNR at convergence but the performance slightly degrades when compared to self-masking. More experiments comparing IMP based masking with OES are provided in Section-\ref{impvsoes-lena} in the Appendix.

\begin{figure}[htb]
    \centering
    \begin{subfigure}[t]{0.24\textwidth}
        \centering
        \includegraphics[width=\textwidth]{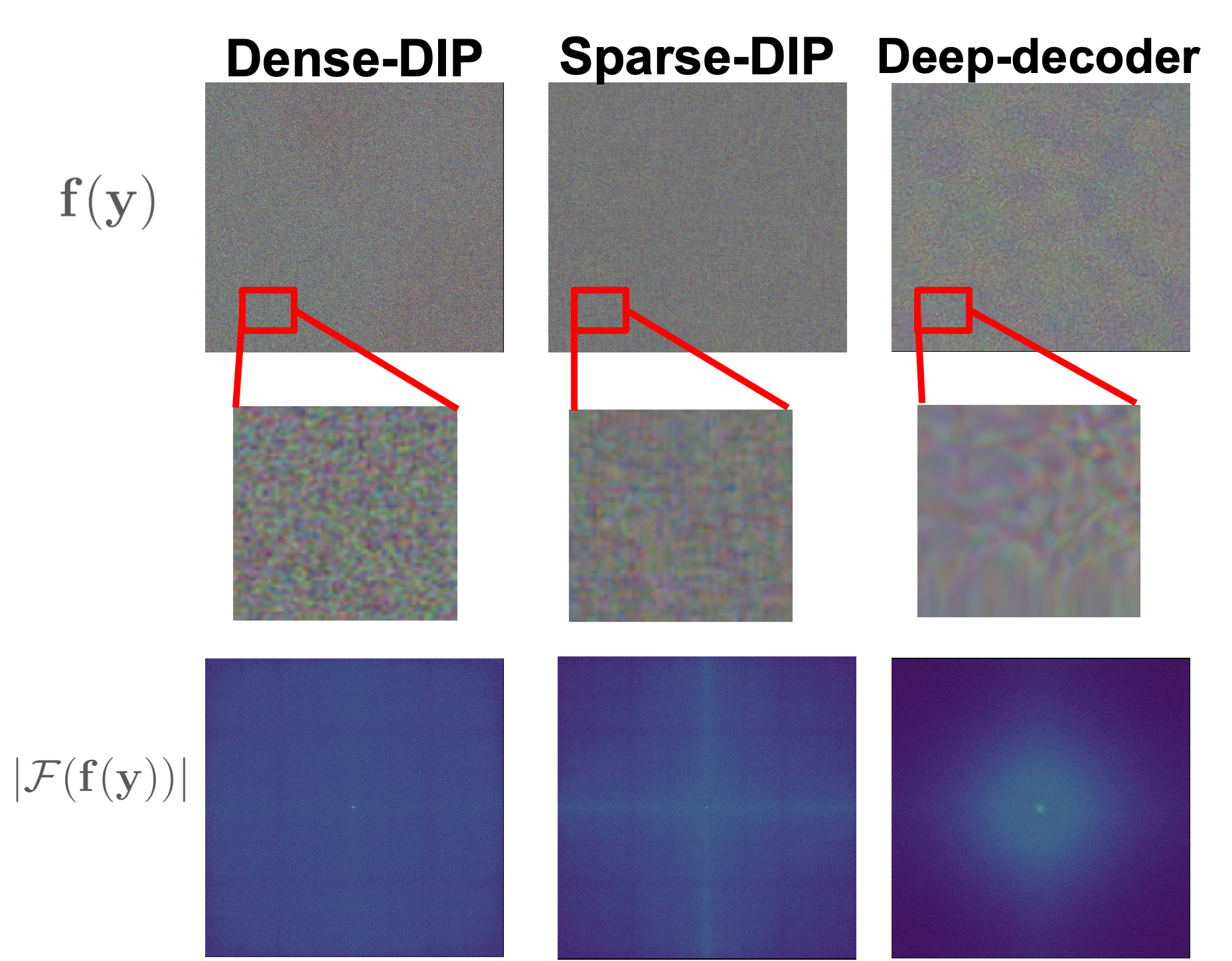}
        \subcaption{Noise impedance}
        \label{fig:fig1}
    \end{subfigure}\hfill
    \begin{subfigure}[t]{0.23\textwidth}
        \centering
        \includegraphics[width=\textwidth]{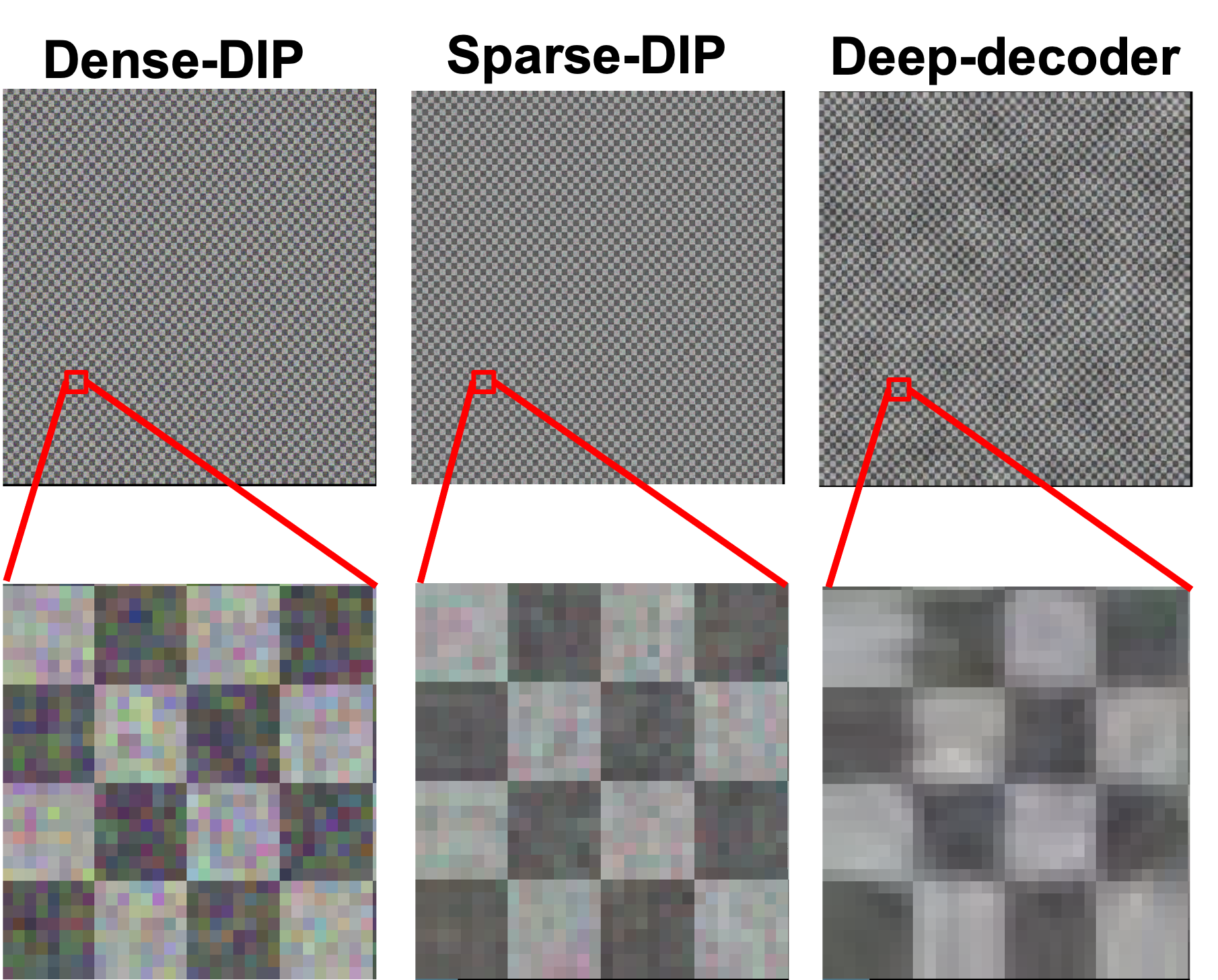}
        \subcaption{Recovery of edges}
        \label{fig:fig2}
    \end{subfigure}
    \caption{a) shows the ability of networks to fit noise. $f(\y)$ is the network output and $|\mathcal{F}(\y)|$ is the magnitude of Fourier coefficients. b) shows quality of recovering edges. }
    \vspace{-0.1in}
\end{figure}

\begin{figure*}[ht]
    \centering
    \begin{subfigure}[t]{0.48\textwidth}
        \centering
        \includegraphics[width=1.0\linewidth]{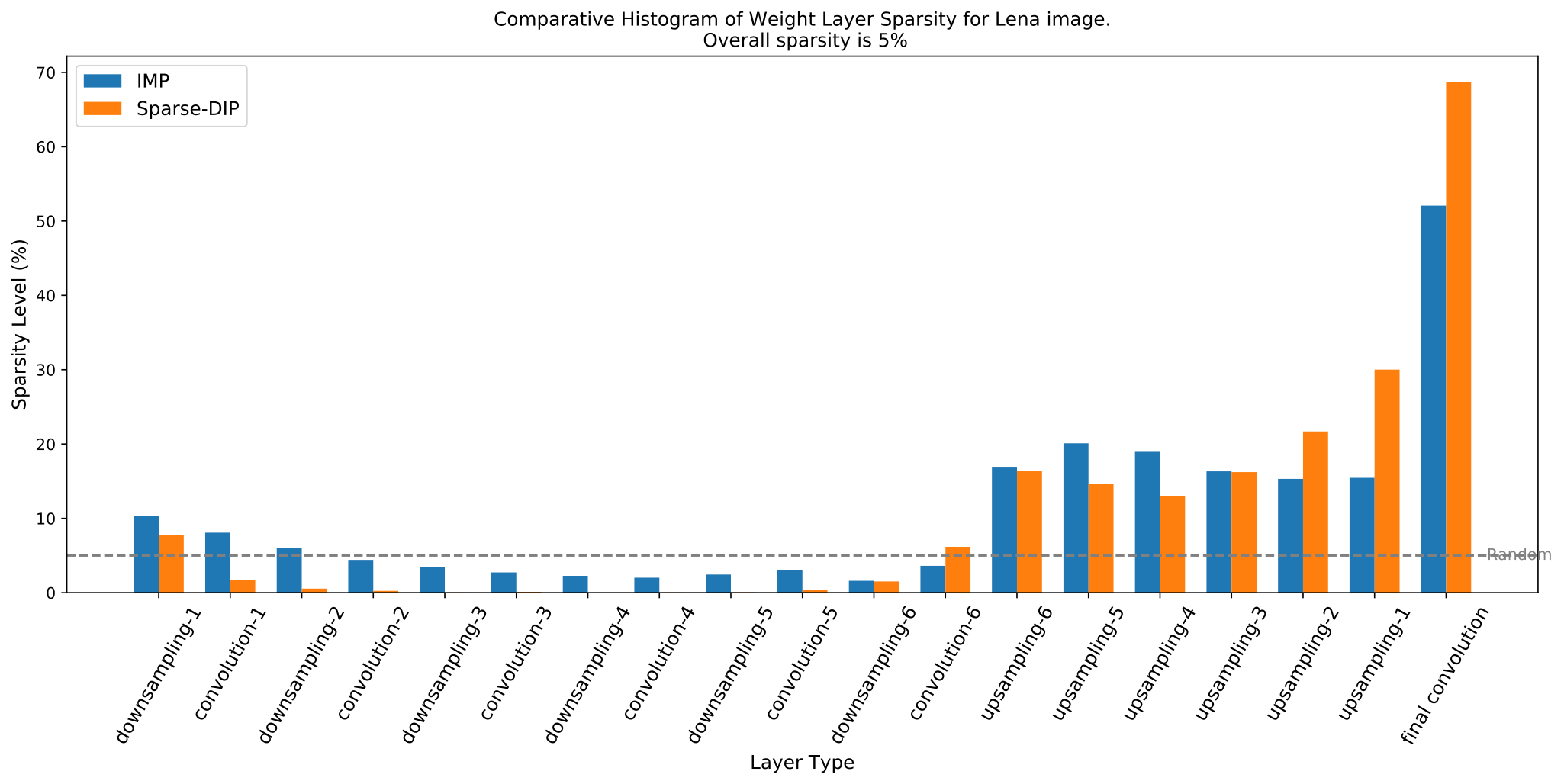}
        \caption{OES vs IMP pruned network}
        \label{fig:subhist1}
    \end{subfigure}%
    \quad
    \begin{subfigure}[t]{0.48\textwidth}
        \centering
        \includegraphics[width=1.0\linewidth]{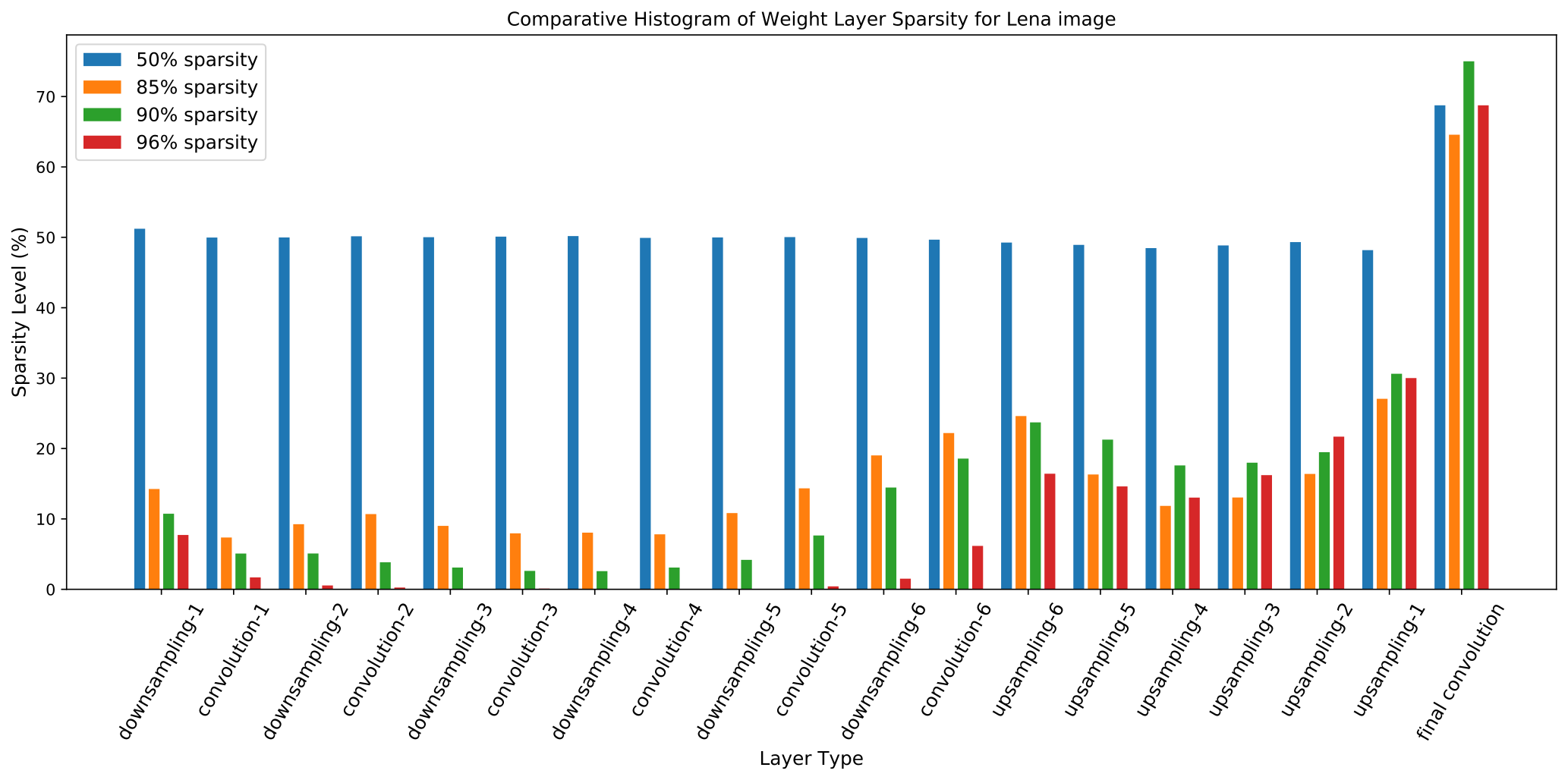}
        \caption{Layer-wise sparsity for pruned Unet for various sparsity levels using OES.}
        \label{fig:subhist2}
    \end{subfigure}
    \caption{a) Layerwise sparsity in Unet architecture for pruning methods IMP and OES. b) Distribution of layerwise parameters for various sparsity levels using OES. The corrupted image $\y$ used was Lena. The overall sparsity in the architecture is $5\%$.}
    \label{fig:combined}
\end{figure*}

\vspace{-0.1in}

\section{Noise impedance of sparse-DIP}
\label{noise-imped}


Sparse-DIPs often outperform deep decoder even with lower levels of parameter count. Based on our experiments, we observe that with images having edges, this difference becomes prominent. To further investigate, we study the noise impedance of the network (denoted as $f(\y)$) when trained to fit random Gaussian noise by minimizing the loss $\| G(\p,\z) - \y \|_{2}^2$ w.r.t. the parameters of the network (dense or sparse), where $\y \sim \mathcal{N}(\mathbf{0}, \sigma^2\mathbf{I})$. This is to see how each network has the capacity to fit white Gaussian noise.
\textit{Dense DIP} fits the noise in the image perfectly with zero training loss and in Fourier domain $\mathcal{F} [f(\y)]$ shows a constant wide-band spectrum which is quite typical for Gaussian white noise. \textit{Deep Decoder} is underparameterized and the training loss does not go to 0 indicating that noise $\y$ lies outside of the output range space of the network. Deep decoder smoothens out the noise to a large extent. The magnitude of the Fourier Transform of the output shows that the cut-off frequency is small essentially making it act like a low-pass filter with small bandwidth. \textit{Sparse-DIP} is also underparameterized and obtained by masking $97\%$ weights of a dense DIP. The magnitude spectrum of $\mathcal{F} [f(\y)]$ shows that x-axis and y-axis of the spectra have much higher magnitude than that of deep decoder, hence it can reconstruct directional edges better than deep decoder. To further explore the image representation and noise impedance capacity, we fit these three networks to a noisy chessboard image, where the strip frequency is very high. We observe from Figure-\ref{fig:fig2} that Dense DIPs recover the high-frequency edges but overfit to noise. Deep decoder has very low cut-off frequency (Figure-\ref{fig:fig1}). It fails to recover the vertical and horizontal edges of the chessboard, although it does smoothen out the noise. Sparse-DIP recovers the edges and does not overfit to noise. The ability of Sparse-DIP to reconstruct high-frequency edges better than deep decoder (with similar number of parameters) explains why it showed superior denoising performance in Figure-\ref{fig:combined_denoise} and Table-\ref{main_table_denoise}.

\vspace{-0.1in}

\section{Pruned architecture study - Finding 5}
\label{arch-prune}
Throughout all the experiments, we used Unet without skip connection as the Dense-DIP architecture. In Figure-\ref{fig:subhist1}, we show how the different layers of Unet are pruned with OES and IMP. These may shed light on the superior performance of OES when compared to IMP. In Figure-\ref{fig:subhist2}, we show the pruning pattern for OES masking for various levels of sparsity. We make the following observations: 1) \textit{Importance of first and last layer}: The first layer of the encoder (\textit{convolution+dowsampling}) layer and the last layer of decoder (\textit{convolution} layer) have large number of remaining weights. The final reconstructed image is formed after convolution in the last layer, so it justifies the observation that the final convolution layer has the least amount of pruned weights. 2) \textit{Towards the emergence of deep-decoder}: In Figure-\ref{fig:subhist2}, we observe that for various levels of sparsity, the decoder part of the architecture is pruned the least. This leads to the observation that for image generation, the upsampling layers play a crucial role, also observed in \citet{liu2023devil}. This further justifies the use of Deep decoder proposed by \citet{heckel2018deep}, where the authors only use the decoder part of the Unet. 3) \textit{Encoder layers play a role in overfitting}: When comparing the architecture of IMP-pruned vs OES-pruned networks, we observe that IMP prefers the layers in the encoder much more than OES. 4) \textit{Importance of encoder-decoder junction}: The junction between the encoder-decoder is important as it has lot of non-zero parameters after pruning. This part is responsible for the generation of the low-frequency information of the image, which composes the majority of the information for natural images. This is because the spatial feature in this layer (because of simultaneous downsampling) is comparable to convolutional patch filter size, making its receptive field larger. 5) \textit{Pruning pattern for various sparsities}: We observe a similarity in the sparsity pattern across different layers in the shape of 'W'. For various pruning percentages $85\%,90\%$ and $96\%$, we observe a similarity in the sparsity pattern across different layers. The three most important layers for the Unet seem to be the first layer (also the first layer of the encoder), the encoder-decoder junction and the last layer (final convolution). 

\section{Limitations}
While our work presents a novel method to prevent overfitting, it is essential to acknowledge few limitations:

\begin{itemize}
[leftmargin=*,noitemsep,topsep=0pt,parsep=0pt,partopsep=0pt]
 \setlength\itemsep{0.25em}
\item Sparse networks tend to overfit slightly for transferring across different domains (Figure-\ref{fig:cross_dat_combined}).
\item Finding the mask adds computational overhead due to the Gumbel Softmax reparameterization. Since the masks are transferable, this overhead is not significant.
\item Specialized tasks, such as MRI image processing, require unique architectures (e.g., two-channel Unet), limiting the transferability of OES subnetworks across different tasks with different architectures.
\end{itemize}

\vspace{-0.1in}
\section{Conclusion}

In this work, we demonstrate for the first time that in a dense deep image generator network, there exists a hidden subnetwork (sparse DIP) at initialization that shows potential of reconstructing low-frequency information of an image from only its noisy measurements. Sparse DIPs show significant potential for image reconstruction and transferability, surpassing traditional pruning methods. We believe that the connection between sparsity in the generator network and the low-dimensionality of the image output (situated in the manifold of images) prompts further theoretical investigation. We aim to further explore the role of these sparse networks within diffusion model-based generative frameworks, aiming to expedite the process and enhance the quality of generated images.

\section*{Acknowledgment}
AG acknowledges Shijun Liang (PhD student, Michigan State University (MSU)) for his help with the MRI reconstruction experiments and Ismail Alkhouri and Evan Bell from MSU for helpful discussions. AG, SR, RR, and KS acknowledge support from NSF CCF-2212065.
QQ acknowledges support from NSF CAREER CCF-2143904, NSF CCF-2212066, NSF CCF-2212326, ONR N00014-22-1-2529, and a gift grant from KLA. The authors also acknowledge the comments of the anonymous reviewers, which helped improve the manuscript.

\section*{Impact Statement}
 Our research aims to enhance image reconstruction efficiency using sparse generator networks. There are some potential societal impacts of this advancement, particularly in the domain of image generation and recovery; however, we do not feel that any specific impacts need to be highlighted here.
\nocite{langley00}

\bibliography{sparse_dip}

\begin{thebibliography}{48}
\providecommand{\natexlab}[1]{#1}
\providecommand{\url}[1]{\texttt{#1}}
\expandafter\ifx\csname urlstyle\endcsname\relax
  \providecommand{\doi}[1]{doi: #1}\else
  \providecommand{\doi}{doi: \begingroup \urlstyle{rm}\Url}\fi

\bibitem[Arican et~al.(2022)Arican, Kara, Bredell, and Konukoglu]{arican2022isnas}
Arican, M.~E., Kara, O., Bredell, G., and Konukoglu, E.
\newblock Isnas-dip: Image-specific neural architecture search for deep image prior.
\newblock In \emph{Proceedings of the IEEE/CVF Conference on Computer Vision and Pattern Recognition}, pp.\  1960--1968, 2022.

\bibitem[Bell et~al.(2023)Bell, Liang, Qu, and Ravishankar]{10096631}
Bell, E., Liang, S., Qu, Q., and Ravishankar, S.
\newblock Robust self-guided deep image prior.
\newblock In \emph{ICASSP 2023 - 2023 IEEE International Conference on Acoustics, Speech and Signal Processing (ICASSP)}, pp.\  1--5, 2023.
\newblock \doi{10.1109/ICASSP49357.2023.10096631}.

\bibitem[Bevilacqua et~al.(2012)Bevilacqua, Roumy, Guillemot, and Alberi-Morel]{bevilacqua2012low}
Bevilacqua, M., Roumy, A., Guillemot, C., and Alberi-Morel, M.~L.
\newblock Low-complexity single-image super-resolution based on nonnegative neighbor embedding.
\newblock 2012.

\bibitem[Cascarano et~al.(2021)Cascarano, Sebastiani, Comes, Franchini, and Porta]{cascarano2021combining}
Cascarano, P., Sebastiani, A., Comes, M.~C., Franchini, G., and Porta, F.
\newblock Combining weighted total variation and deep image prior for natural and medical image restoration via admm.
\newblock In \emph{2021 21st International Conference on Computational Science and Its Applications (ICCSA)}, pp.\  39--46. IEEE, 2021.

\bibitem[Chakrabarty \& Maji(2019)Chakrabarty and Maji]{chakrabarty2019spectral}
Chakrabarty, P. and Maji, S.
\newblock The spectral bias of the deep image prior.
\newblock \emph{arXiv preprint arXiv:1912.08905}, 2019.

\bibitem[Chen et~al.(2020)Chen, Gao, Robb, and Huang]{chen2020dip}
Chen, Y.-C., Gao, C., Robb, E., and Huang, J.-B.
\newblock Nas-dip: Learning deep image prior with neural architecture search.
\newblock In \emph{Computer Vision--ECCV 2020: 16th European Conference, Glasgow, UK, August 23--28, 2020, Proceedings, Part XVIII 16}, pp.\  442--459. Springer, 2020.

\bibitem[Cheng et~al.(2019)Cheng, Gadelha, Maji, and Sheldon]{cheng2019bayesian}
Cheng, Z., Gadelha, M., Maji, S., and Sheldon, D.
\newblock A bayesian perspective on the deep image prior.
\newblock In \emph{Proceedings of the IEEE/CVF Conference on Computer Vision and Pattern Recognition}, pp.\  5443--5451, 2019.

\bibitem[da~Cunha et~al.(2021)da~Cunha, Natale, and Viennot]{da2021proving}
da~Cunha, A., Natale, E., and Viennot, L.
\newblock Proving the lottery ticket hypothesis for convolutional neural networks.
\newblock In \emph{International Conference on Learning Representations}, 2021.

\bibitem[Ding et~al.(2021)Ding, Jiang, Chen, Qu, and Zhu]{ding2021rank}
Ding, L., Jiang, L., Chen, Y., Qu, Q., and Zhu, Z.
\newblock Rank overspecified robust matrix recovery: Subgradient method and exact recovery.
\newblock \emph{arXiv preprint arXiv:2109.11154}, 2021.

\bibitem[Evci et~al.(2020)Evci, Gale, Menick, Castro, and Elsen]{evci2020rigging}
Evci, U., Gale, T., Menick, J., Castro, P.~S., and Elsen, E.
\newblock Rigging the lottery: Making all tickets winners.
\newblock In \emph{International Conference on Machine Learning}, pp.\  2943--2952. PMLR, 2020.

\bibitem[Frankle \& Carbin(2018)Frankle and Carbin]{frankle2018lottery}
Frankle, J. and Carbin, M.
\newblock The lottery ticket hypothesis: Finding sparse, trainable neural networks.
\newblock \emph{arXiv preprint arXiv:1803.03635}, 2018.

\bibitem[Frankle et~al.(2019)Frankle, Dziugaite, Roy, and Carbin]{frankle2019stabilizing}
Frankle, J., Dziugaite, G.~K., Roy, D.~M., and Carbin, M.
\newblock Stabilizing the lottery ticket hypothesis.
\newblock \emph{arXiv preprint arXiv:1903.01611}, 2019.

\bibitem[Hassibi et~al.(1993)Hassibi, Stork, and Wolff]{hassibi1993optimal}
Hassibi, B., Stork, D.~G., and Wolff, G.~J.
\newblock Optimal brain surgeon and general network pruning.
\newblock In \emph{IEEE international conference on neural networks}, pp.\  293--299. IEEE, 1993.

\bibitem[He et~al.(2018)He, Kang, Dong, Fu, and Yang]{he2018soft}
He, Y., Kang, G., Dong, X., Fu, Y., and Yang, Y.
\newblock Soft filter pruning for accelerating deep convolutional neural networks.
\newblock \emph{arXiv preprint arXiv:1808.06866}, 2018.

\bibitem[Heckel \& Hand(2018)Heckel and Hand]{heckel2018deep}
Heckel, R. and Hand, P.
\newblock Deep decoder: Concise image representations from untrained non-convolutional networks.
\newblock \emph{arXiv preprint arXiv:1810.03982}, 2018.

\bibitem[Huang et~al.(2022)Huang, Chen, Fang, Menkovski, Zhao, Yin, Pei, Mocanu, Wang, Pechenizkiy, et~al.]{huang2022you}
Huang, T., Chen, T., Fang, M., Menkovski, V., Zhao, J., Yin, L., Pei, Y., Mocanu, D.~C., Wang, Z., Pechenizkiy, M., et~al.
\newblock You can have better graph neural networks by not training weights at all: Finding untrained gnns tickets.
\newblock \emph{arXiv preprint arXiv:2211.15335}, 2022.

\bibitem[Jin et~al.(2017)Jin, McCann, Froustey, and Unser]{jin2017deep}
Jin, K.~H., McCann, M.~T., Froustey, E., and Unser, M.
\newblock Deep convolutional neural network for inverse problems in imaging.
\newblock \emph{IEEE transactions on image processing}, 26\penalty0 (9):\penalty0 4509--4522, 2017.

\bibitem[Jin et~al.(2022)Jin, Carbin, Roy, Frankle, and Dziugaite]{jin2022pruning}
Jin, T., Carbin, M., Roy, D., Frankle, J., and Dziugaite, G.~K.
\newblock Pruning’s effect on generalization through the lens of training and regularization.
\newblock \emph{Advances in Neural Information Processing Systems}, 35:\penalty0 37947--37961, 2022.

\bibitem[Jo et~al.(2021)Jo, Chun, and Choi]{jo2021rethinking}
Jo, Y., Chun, S.~Y., and Choi, J.
\newblock Rethinking deep image prior for denoising.
\newblock In \emph{Proceedings of the IEEE/CVF International Conference on Computer Vision}, pp.\  5087--5096, 2021.

\bibitem[Katsaggelos(1989)]{katsaggelos1989iterative}
Katsaggelos, A.~K.
\newblock Iterative image restoration algorithms.
\newblock \emph{Optical engineering}, 28\penalty0 (7):\penalty0 735--748, 1989.

\bibitem[LeCun et~al.(1989)LeCun, Denker, and Solla]{lecun1989optimal}
LeCun, Y., Denker, J., and Solla, S.
\newblock Optimal brain damage.
\newblock \emph{Advances in neural information processing systems}, 2, 1989.

\bibitem[Lee et~al.(2018)Lee, Ajanthan, and Torr]{lee2018snip}
Lee, N., Ajanthan, T., and Torr, P.~H.
\newblock Snip: Single-shot network pruning based on connection sensitivity.
\newblock \emph{arXiv preprint arXiv:1810.02340}, 2018.

\bibitem[Liu et~al.(2019)Liu, Sun, Xu, and Kamilov]{liu2019image}
Liu, J., Sun, Y., Xu, X., and Kamilov, U.~S.
\newblock Image restoration using total variation regularized deep image prior.
\newblock In \emph{ICASSP 2019-2019 IEEE International Conference on Acoustics, Speech and Signal Processing (ICASSP)}, pp.\  7715--7719. Ieee, 2019.

\bibitem[Liu et~al.(2023)Liu, Li, Pang, Nie, and Yap]{liu2023devil}
Liu, Y., Li, J., Pang, Y., Nie, D., and Yap, P.-T.
\newblock The devil is in the upsampling: Architectural decisions made simpler for denoising with deep image prior.
\newblock In \emph{Proceedings of the IEEE/CVF International Conference on Computer Vision}, pp.\  12408--12417, 2023.

\bibitem[Maddison et~al.(2016)Maddison, Mnih, and Teh]{maddison2016concrete}
Maddison, C.~J., Mnih, A., and Teh, Y.~W.
\newblock The concrete distribution: A continuous relaxation of discrete random variables.
\newblock \emph{arXiv preprint arXiv:1611.00712}, 2016.

\bibitem[Malach et~al.(2020)Malach, Yehudai, Shalev-Schwartz, and Shamir]{malach2020proving}
Malach, E., Yehudai, G., Shalev-Schwartz, S., and Shamir, O.
\newblock Proving the lottery ticket hypothesis: Pruning is all you need.
\newblock In \emph{International Conference on Machine Learning}, pp.\  6682--6691. PMLR, 2020.

\bibitem[Mallya et~al.(2018)Mallya, Davis, and Lazebnik]{mallya2018piggyback}
Mallya, A., Davis, D., and Lazebnik, S.
\newblock Piggyback: Adapting a single network to multiple tasks by learning to mask weights.
\newblock In \emph{Proceedings of the European conference on computer vision (ECCV)}, pp.\  67--82, 2018.

\bibitem[Mataev et~al.(2019)Mataev, Milanfar, and Elad]{mataev2019deepred}
Mataev, G., Milanfar, P., and Elad, M.
\newblock Deepred: Deep image prior powered by red.
\newblock In \emph{Proceedings of the IEEE/CVF International Conference on Computer Vision Workshops}, pp.\  0--0, 2019.

\bibitem[Mehta(2019)]{mehta2019sparse}
Mehta, R.
\newblock Sparse transfer learning via winning lottery tickets.
\newblock \emph{arXiv preprint arXiv:1905.07785}, 2019.

\bibitem[Metzler et~al.(2018)Metzler, Mousavi, Heckel, and Baraniuk]{metzler2018unsupervised}
Metzler, C.~A., Mousavi, A., Heckel, R., and Baraniuk, R.~G.
\newblock Unsupervised learning with stein's unbiased risk estimator.
\newblock \emph{arXiv preprint arXiv:1805.10531}, 2018.

\bibitem[Paul et~al.(2022)Paul, Chen, Larsen, Frankle, Ganguli, and Dziugaite]{paul2022unmasking}
Paul, M., Chen, F., Larsen, B.~W., Frankle, J., Ganguli, S., and Dziugaite, G.~K.
\newblock Unmasking the lottery ticket hypothesis: What's encoded in a winning ticket's mask?
\newblock \emph{arXiv preprint arXiv:2210.03044}, 2022.

\bibitem[Ramanujan et~al.(2020)Ramanujan, Wortsman, Kembhavi, Farhadi, and Rastegari]{ramanujan2020s}
Ramanujan, V., Wortsman, M., Kembhavi, A., Farhadi, A., and Rastegari, M.
\newblock What's hidden in a randomly weighted neural network?
\newblock In \emph{Proceedings of the IEEE/CVF conference on computer vision and pattern recognition}, pp.\  11893--11902, 2020.

\bibitem[Renda et~al.(2020)Renda, Frankle, and Carbin]{renda2020comparing}
Renda, A., Frankle, J., and Carbin, M.
\newblock Comparing rewinding and fine-tuning in neural network pruning.
\newblock \emph{arXiv preprint arXiv:2003.02389}, 2020.

\bibitem[Shi et~al.(2022)Shi, Mettes, Maji, and Snoek]{shi2022measuring}
Shi, Z., Mettes, P., Maji, S., and Snoek, C.~G.
\newblock On measuring and controlling the spectral bias of the deep image prior.
\newblock \emph{International Journal of Computer Vision}, 130\penalty0 (4):\penalty0 885--908, 2022.

\bibitem[Sreenivasan et~al.(2022)Sreenivasan, Sohn, Yang, Grinde, Nagle, Wang, Xing, Lee, and Papailiopoulos]{sreenivasan2022rare}
Sreenivasan, K., Sohn, J.-y., Yang, L., Grinde, M., Nagle, A., Wang, H., Xing, E., Lee, K., and Papailiopoulos, D.
\newblock Rare gems: Finding lottery tickets at initialization.
\newblock \emph{Advances in Neural Information Processing Systems}, 35:\penalty0 14529--14540, 2022.

\bibitem[Sun et~al.(2020)Sun, Sanchez, Latorre, and Cevher]{sun2020solving}
Sun, Z., Sanchez, T., Latorre, F., and Cevher, V.
\newblock Solving inverse problems with hybrid deep image priors: the challenge of preventing overfitting.
\newblock \emph{arXiv preprint arXiv:2011.01748}, 2020.

\bibitem[Tanaka et~al.(2020)Tanaka, Kunin, Yamins, and Ganguli]{tanaka2020pruning}
Tanaka, H., Kunin, D., Yamins, D.~L., and Ganguli, S.
\newblock Pruning neural networks without any data by iteratively conserving synaptic flow.
\newblock \emph{Advances in neural information processing systems}, 33:\penalty0 6377--6389, 2020.

\bibitem[Ulyanov et~al.(2018)Ulyanov, Vedaldi, and Lempitsky]{ulyanov2018deep}
Ulyanov, D., Vedaldi, A., and Lempitsky, V.
\newblock Deep image prior.
\newblock In \emph{Proceedings of the IEEE conference on computer vision and pattern recognition}, pp.\  9446--9454, 2018.

\bibitem[Venkatakrishnan et~al.(2013)Venkatakrishnan, Bouman, and Wohlberg]{venkatakrishnan2013plug}
Venkatakrishnan, S.~V., Bouman, C.~A., and Wohlberg, B.
\newblock Plug-and-play priors for model based reconstruction.
\newblock In \emph{2013 IEEE global conference on signal and information processing}, pp.\  945--948. IEEE, 2013.

\bibitem[Wang et~al.(2020)Wang, Zhang, and Grosse]{wang2020picking}
Wang, C., Zhang, G., and Grosse, R.
\newblock Picking winning tickets before training by preserving gradient flow.
\newblock \emph{arXiv preprint arXiv:2002.07376}, 2020.

\bibitem[Wang et~al.(2021)Wang, Li, Zhuang, Chen, Liang, and Sun]{wang2021early}
Wang, H., Li, T., Zhuang, Z., Chen, T., Liang, H., and Sun, J.
\newblock Early stopping for deep image prior.
\newblock \emph{arXiv preprint arXiv:2112.06074}, 2021.

\bibitem[Wortsman et~al.(2020)Wortsman, Ramanujan, Liu, Kembhavi, Rastegari, Yosinski, and Farhadi]{wortsman2020supermasks}
Wortsman, M., Ramanujan, V., Liu, R., Kembhavi, A., Rastegari, M., Yosinski, J., and Farhadi, A.
\newblock Supermasks in superposition.
\newblock \emph{Advances in Neural Information Processing Systems}, 33:\penalty0 15173--15184, 2020.

\bibitem[Wu et~al.(2023)Wu, Chen, Jiang, and Wang]{wu2023chasing}
Wu, Q., Chen, X., Jiang, Y., and Wang, Z.
\newblock Chasing better deep image priors between over-and under-parameterization.
\newblock \emph{Transactions on Machine Learning Research}, 2023.

\bibitem[You et~al.(2020)You, Zhu, Qu, and Ma]{you2020robust}
You, C., Zhu, Z., Qu, Q., and Ma, Y.
\newblock Robust recovery via implicit bias of discrepant learning rates for double over-parameterization.
\newblock \emph{Advances in Neural Information Processing Systems}, 33:\penalty0 17733--17744, 2020.

\bibitem[You et~al.(2019)You, Li, Xu, Fu, Wang, Chen, Baraniuk, Wang, and Lin]{you2019drawing}
You, H., Li, C., Xu, P., Fu, Y., Wang, Y., Chen, X., Baraniuk, R.~G., Wang, Z., and Lin, Y.
\newblock Drawing early-bird tickets: Towards more efficient training of deep networks.
\newblock \emph{arXiv preprint arXiv:1909.11957}, 2019.

\bibitem[Zeyde et~al.(2012)Zeyde, Elad, and Protter]{zeyde2012single}
Zeyde, R., Elad, M., and Protter, M.
\newblock On single image scale-up using sparse-representations.
\newblock In \emph{Curves and Surfaces: 7th International Conference, Avignon, France, June 24-30, 2010, Revised Selected Papers 7}, pp.\  711--730. Springer, 2012.

\bibitem[Zhao et~al.(2019)Zhao, Ni, Zhang, Zhao, Zhang, and Tian]{zhao2019variational}
Zhao, C., Ni, B., Zhang, J., Zhao, Q., Zhang, W., and Tian, Q.
\newblock Variational convolutional neural network pruning.
\newblock In \emph{Proceedings of the IEEE/CVF conference on computer vision and pattern recognition}, pp.\  2780--2789, 2019.

\bibitem[Zhou et~al.(2019)Zhou, Lan, Liu, and Yosinski]{zhou2019deconstructing}
Zhou, H., Lan, J., Liu, R., and Yosinski, J.
\newblock Deconstructing lottery tickets: Zeros, signs, and the supermask.
\newblock \emph{Advances in neural information processing systems}, 32, 2019.

\end{thebibliography}
\bibliographystyle{icml2024}

\newpage
\appendix
\onecolumn

\section*{\large Appendix}
In the appendix section, we provide further extensive results to support our findings presented in the manuscript. We present the following sections sequentially:
\begin{enumerate}
    \item Section-\ref{denoising} presents denoising results which were reported in Finding-2 of the manuscript (\ref{finding-2}). Here, we present denoising performance for several noise levels. 
    \item Section-\ref{prune_comp_tab} contains the performance of OES method with standard pruning methods in literature over images in three different datasets. 
    \item Section-\ref{gumb_softmax_det} summarizes the details of the Gumbel softmax reparameterization trick that was utitlized in learning the mask by OES. 
    \item Sectioin-\ref{related_works} summarizes related works and confirms similar findings with related works. Here, we also highlight the difference of our work and show how OES is a more generalizable approach compared to the related works. We highlight that \textit{no clean image is needed or no prior assumption on architecuture is required for finding a good subnetwork.} 
    \item Section-\ref{imp-den} highlights the difficulty in using IMP for pruning networks for image reconstruction tasks. We also consider the oracle case, where \textit{clean image is used for IMP and we show that it has poor transferrability compared to OES.} 
    \item Section-\ref{inpaint_task} shows transfer to a different task (here inpainting). We test OES masks learned on inpainting and denoising tasks and compare them on the respective tasks. 
    \item Section-\ref{sense_lam_oes} shows the robustness of hyperparameter $\lambda$ when KL regularization is used. 
    \item Section-\ref{mri_recon} extends the OES framework for MRI reconstruction from undersampled k-space measurements. 
    \item Section-\ref{l1reg} shows the comparison and disadvantages of finding mask through $L_{1}$ regularization as done in \citet{sreenivasan2022rare}.
    \item Section-\ref{sense_weight_init} studies the sensitivity of masks obtained at different initialization distribution/initialization scale and when IMP masks are learned at early stop time. 
    \item Section-\ref{deepdec} shows the adverse effects of pruning an already underparameterized deep decoder. 
    \item Section-\ref{comp_class_rec} highlights the difference in neural network pruning for image classification and image reconstruction. To the best of our knowledge, our work shows the phenomenon of Stong Lottery Ticket Hypothesis in image reconstruction for the first time. 
\end{enumerate}

\section{Denoising Results}
\label{denoising}
\begin{figure*}[!h]
    \centering

    \begin{minipage}{0.35\textwidth}
        \centering
        \includegraphics[width=\textwidth]{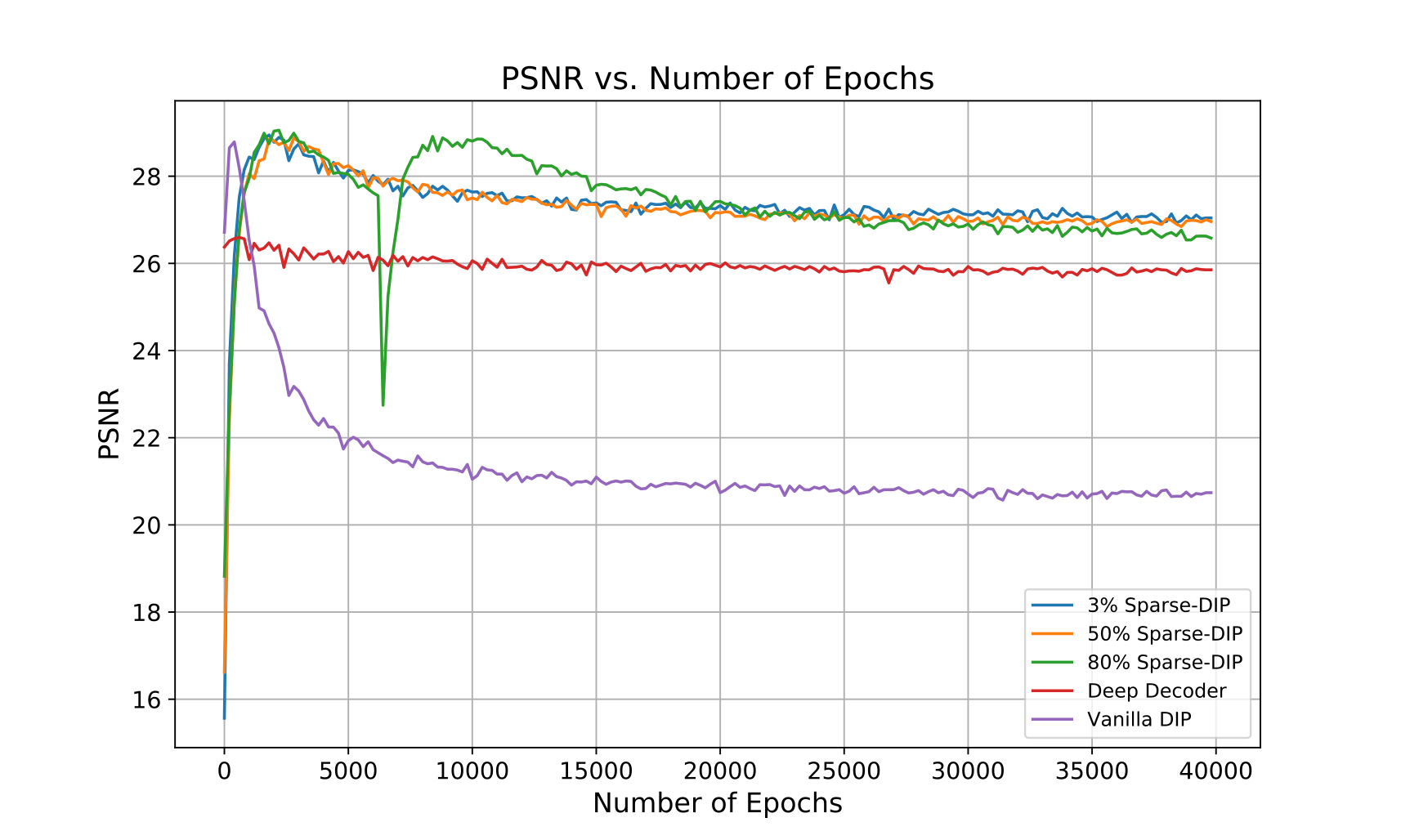}
        \subcaption{Pepper}
        \label{fig:new_figure_1}
    \end{minipage}\quad
    \begin{minipage}{0.35\textwidth}
        \centering
        \includegraphics[width=\textwidth]{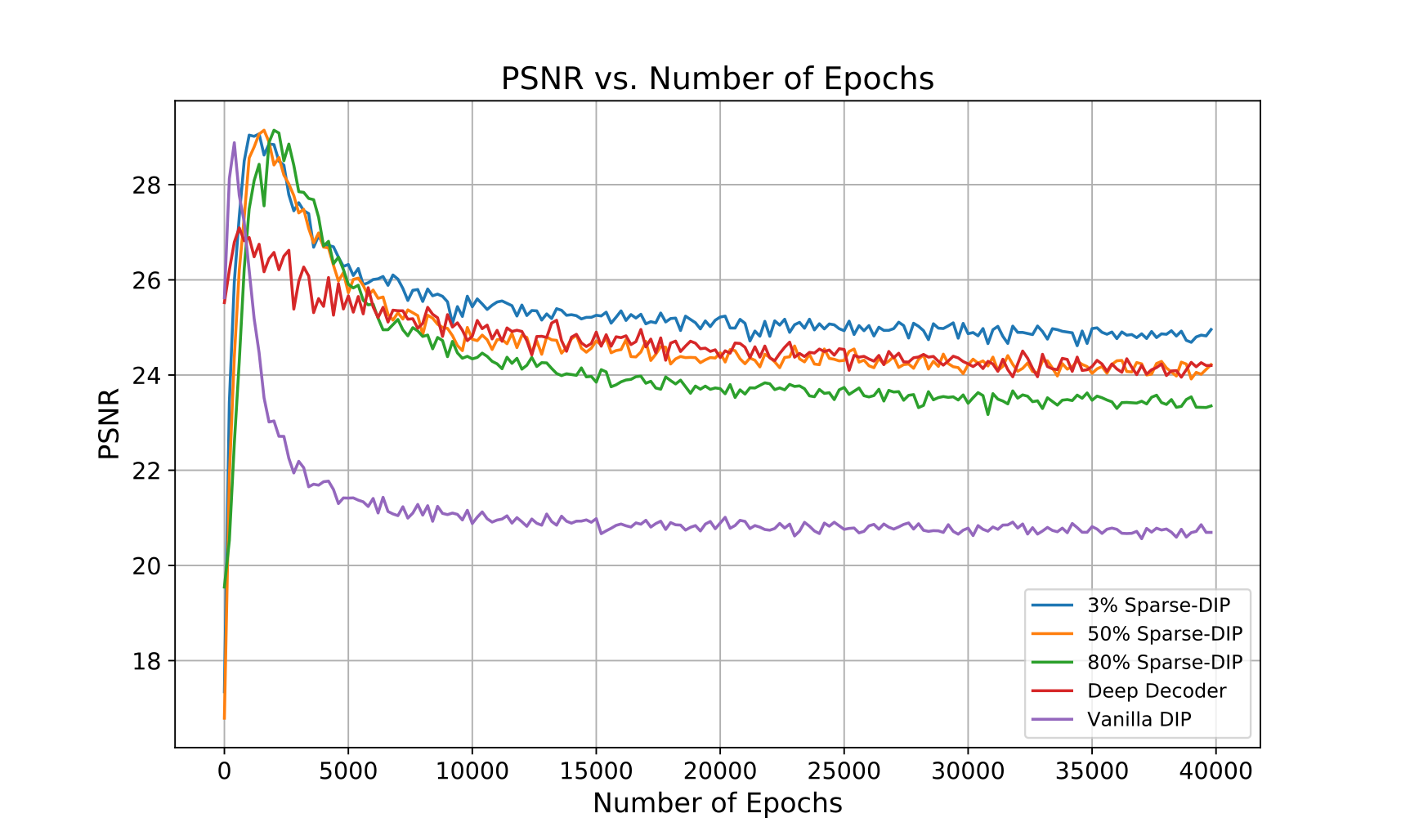}
        \subcaption{Foreman}
        \label{fig:new_figure_2}
    \end{minipage}\quad

    \begin{minipage}{0.35\textwidth}
        \centering
        \includegraphics[width=\textwidth]{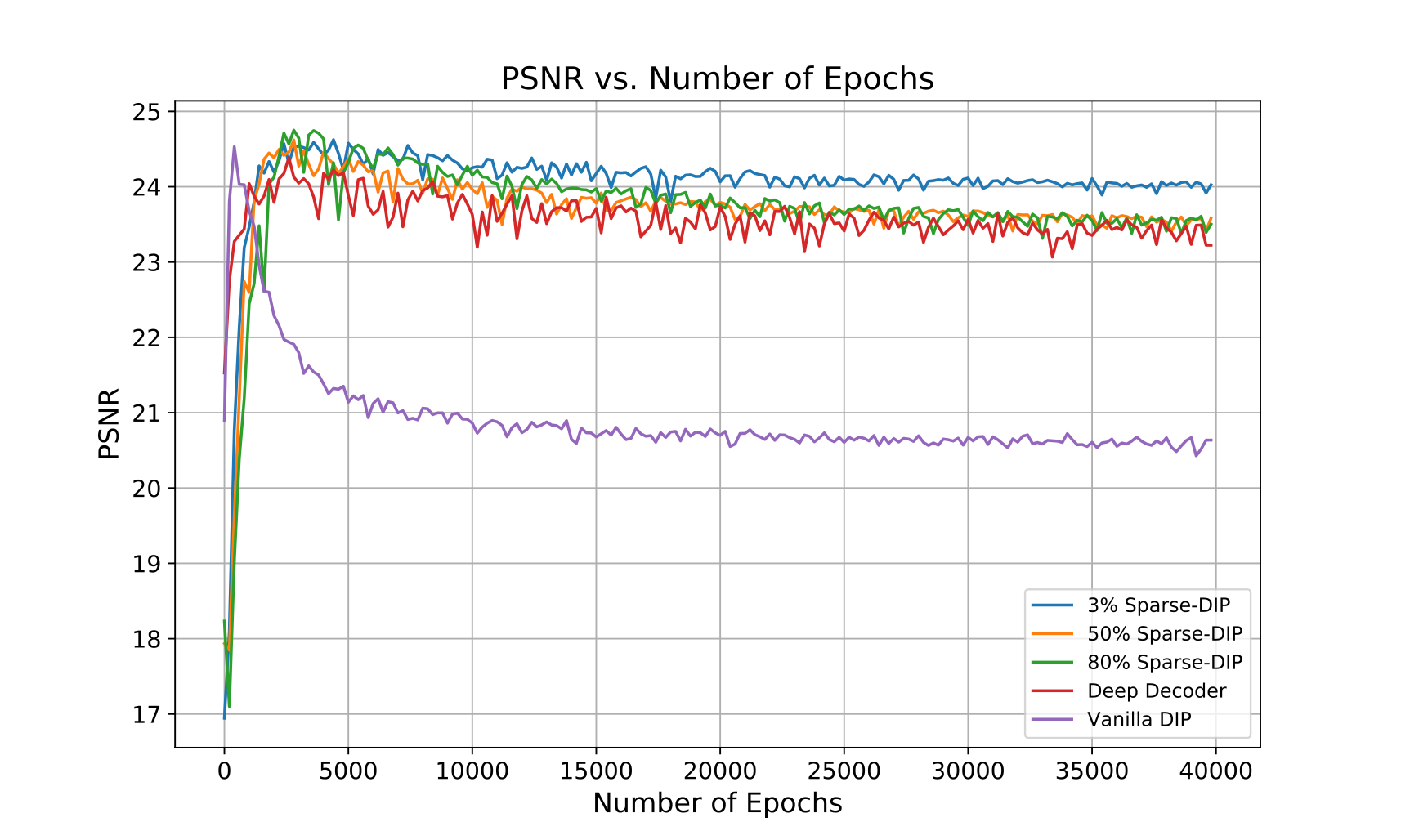}
        \subcaption{Comic}
        \label{fig:new_figure_2}
    \end{minipage}\quad
        \begin{minipage}{0.35\textwidth}
        \centering
        \includegraphics[width=\textwidth]{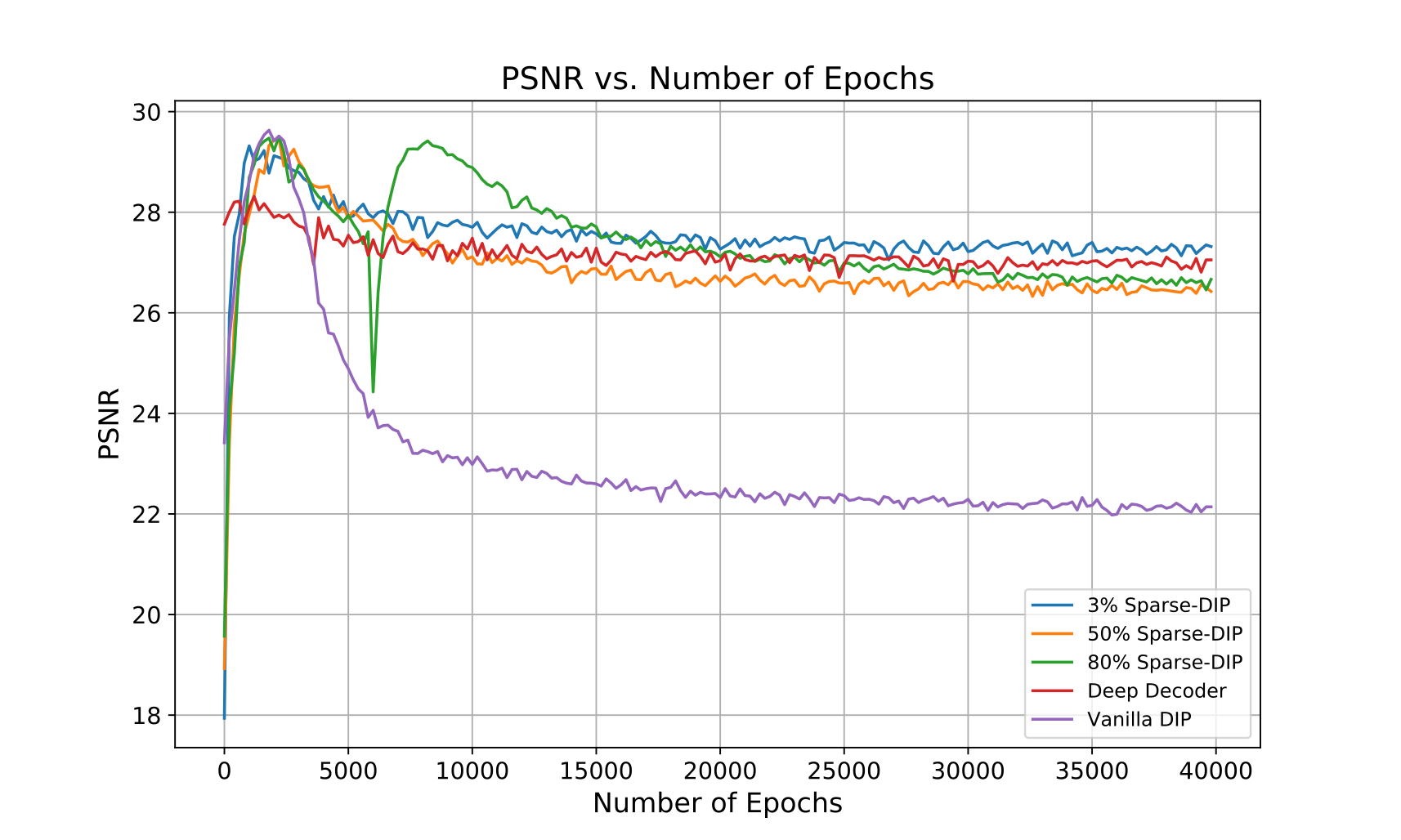}
        \subcaption{Lena}
        \label{fig:new_figure_2}
    \end{minipage}\quad
        \begin{minipage}{0.35\textwidth}
        \centering
        \includegraphics[width=\textwidth]{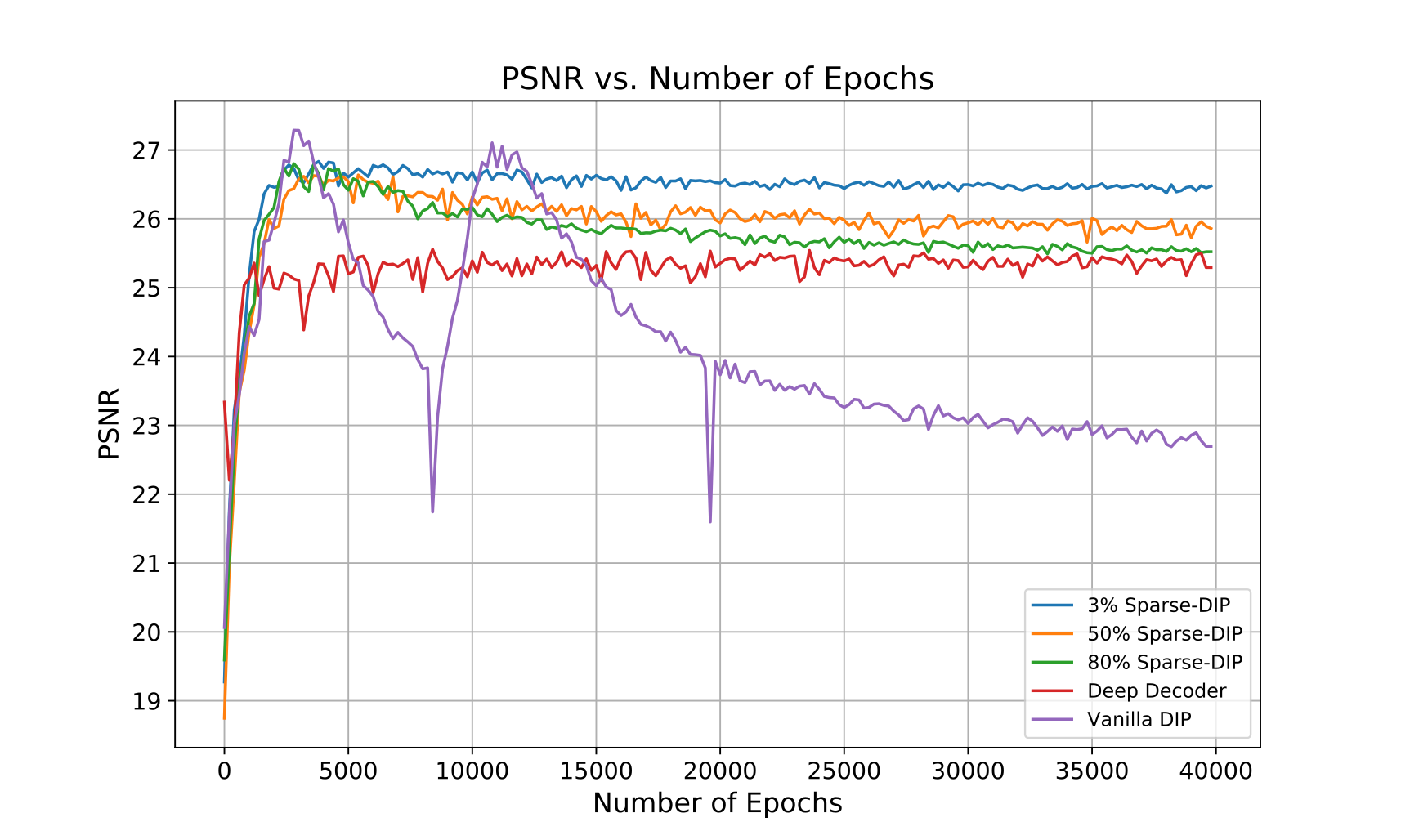}
        \subcaption{Barbara}
        \label{fig:new_figure_2}
    \end{minipage}\quad    
        \begin{minipage}{0.35\textwidth}
        \centering
        \includegraphics[width=\textwidth]{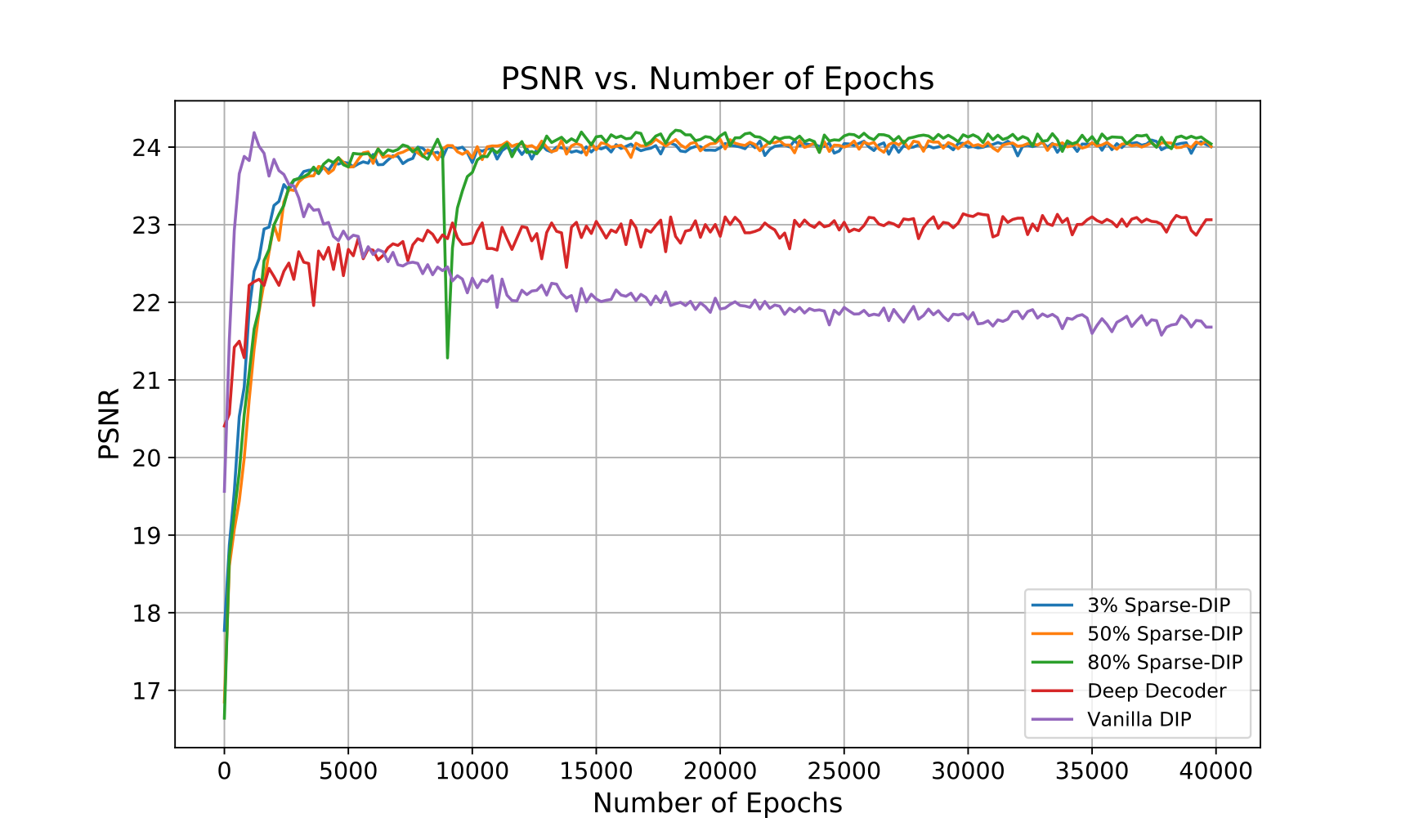}
        \subcaption{Baboon}
        \label{fig:new_figure_2}
    \end{minipage}\quad
        \begin{minipage}{0.35\textwidth}
        \centering
        \includegraphics[width=\textwidth]{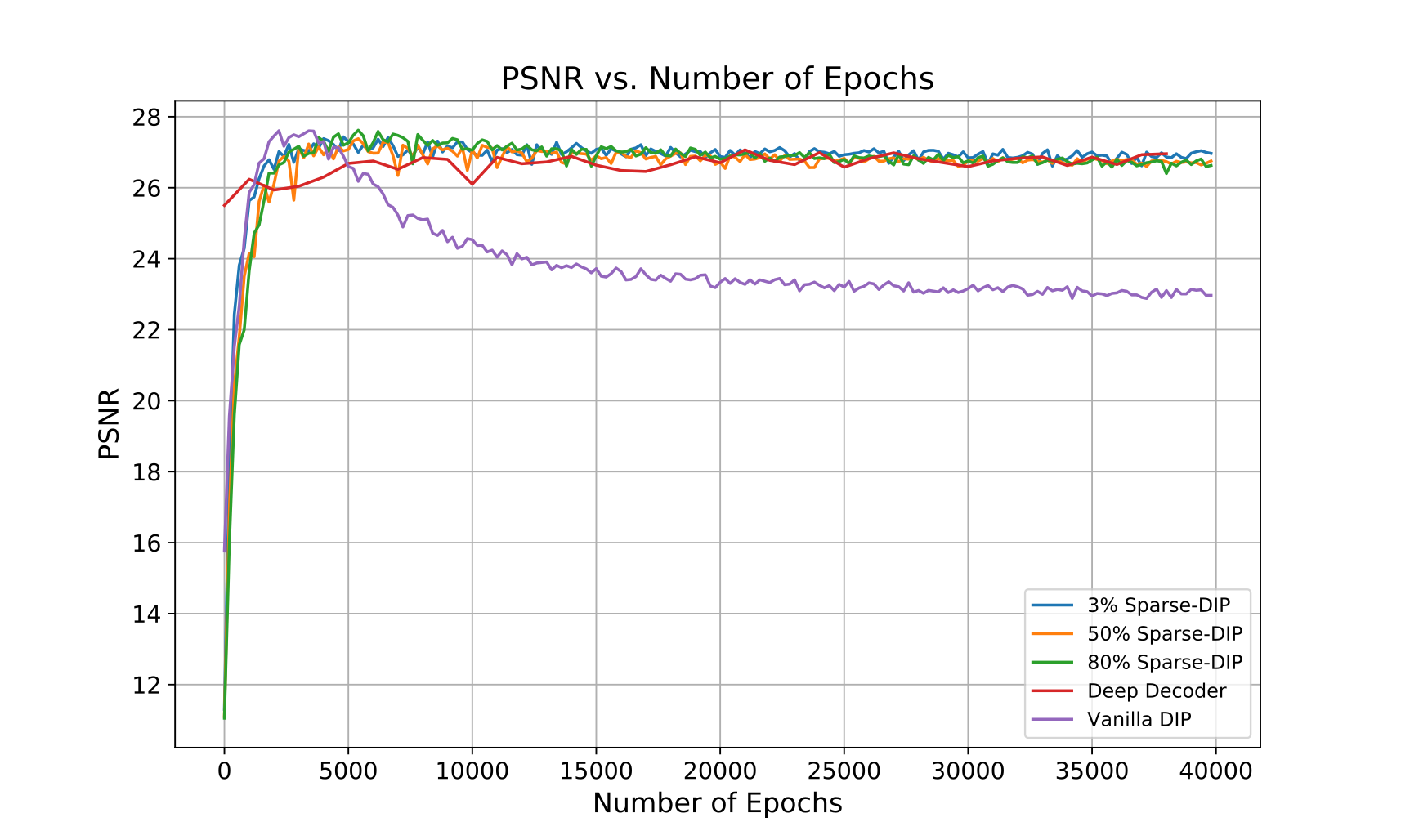}
        \subcaption{Ppt3}
        \label{fig:new_figure_2}
    \end{minipage}\quad
        \begin{minipage}{0.35\textwidth}
        \centering
        \includegraphics[width=\textwidth]{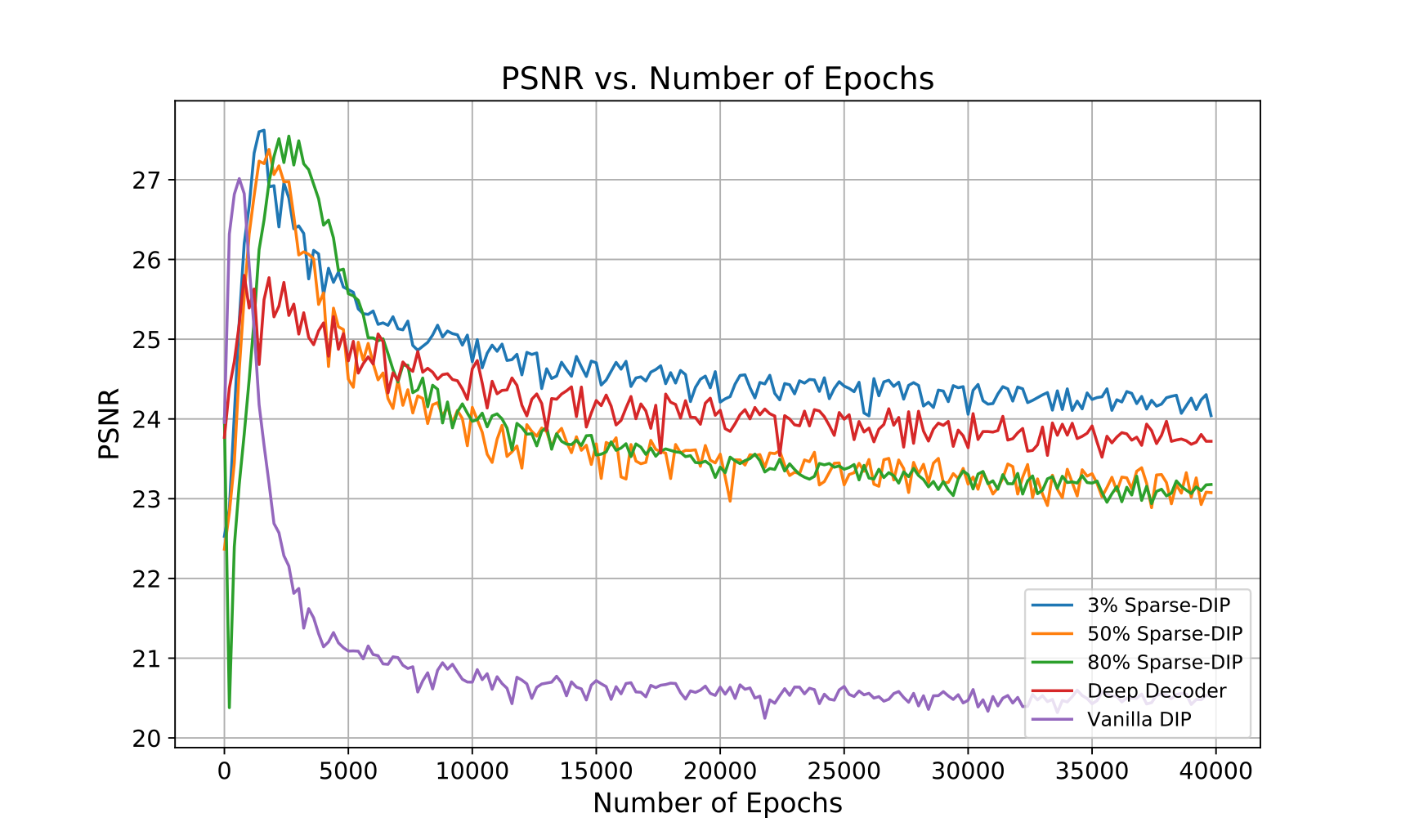}
        \subcaption{Coastguard}
        \label{fig:new_figure_2}
    \end{minipage}\quad
     \begin{minipage}{0.35\textwidth}
        \centering
        \includegraphics[width=\textwidth]{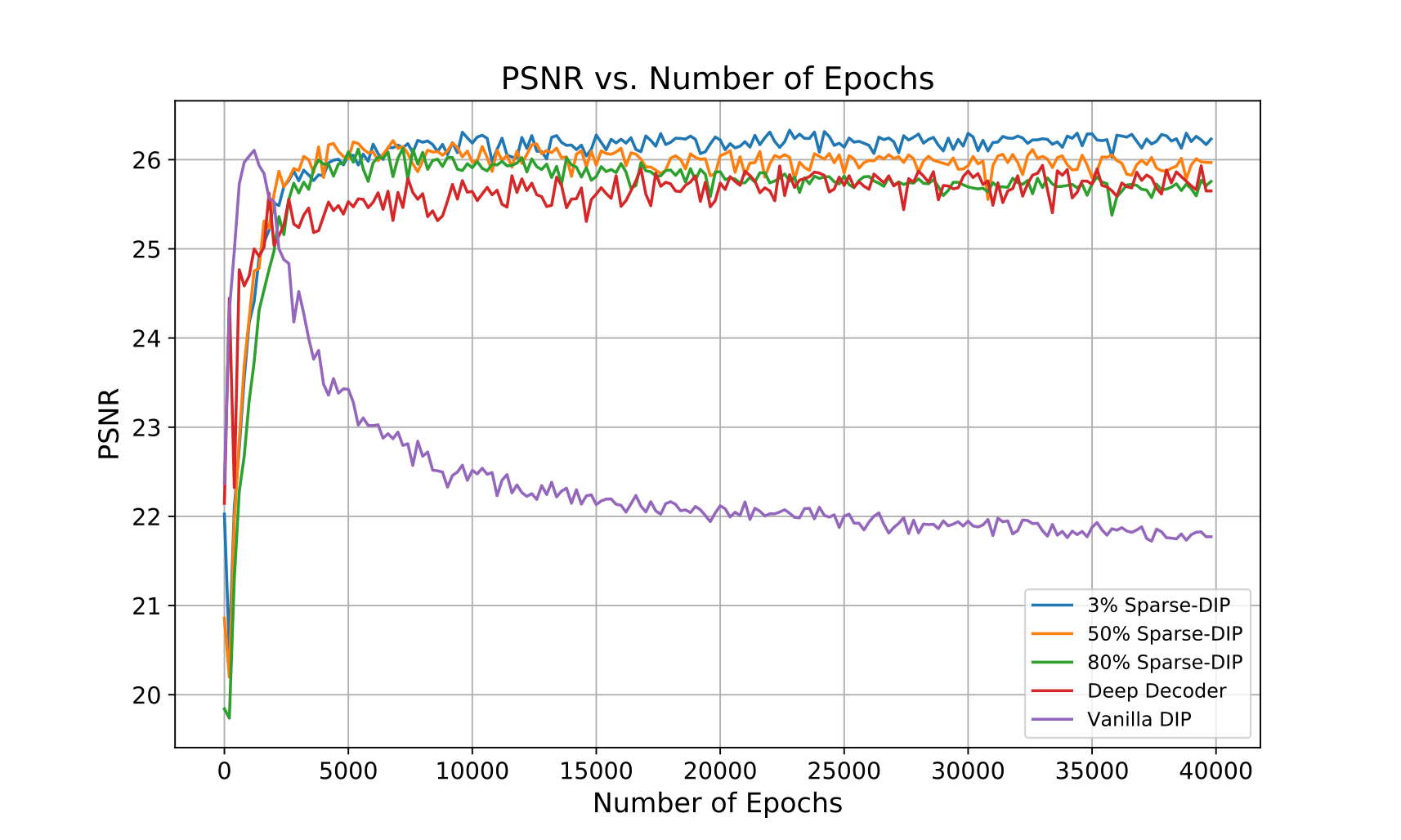}
        \subcaption{Bridge}
        \label{fig:new_figure_2}
    \end{minipage}\quad
        \begin{minipage}{0.35\textwidth}
        \centering
        \includegraphics[width=\textwidth]{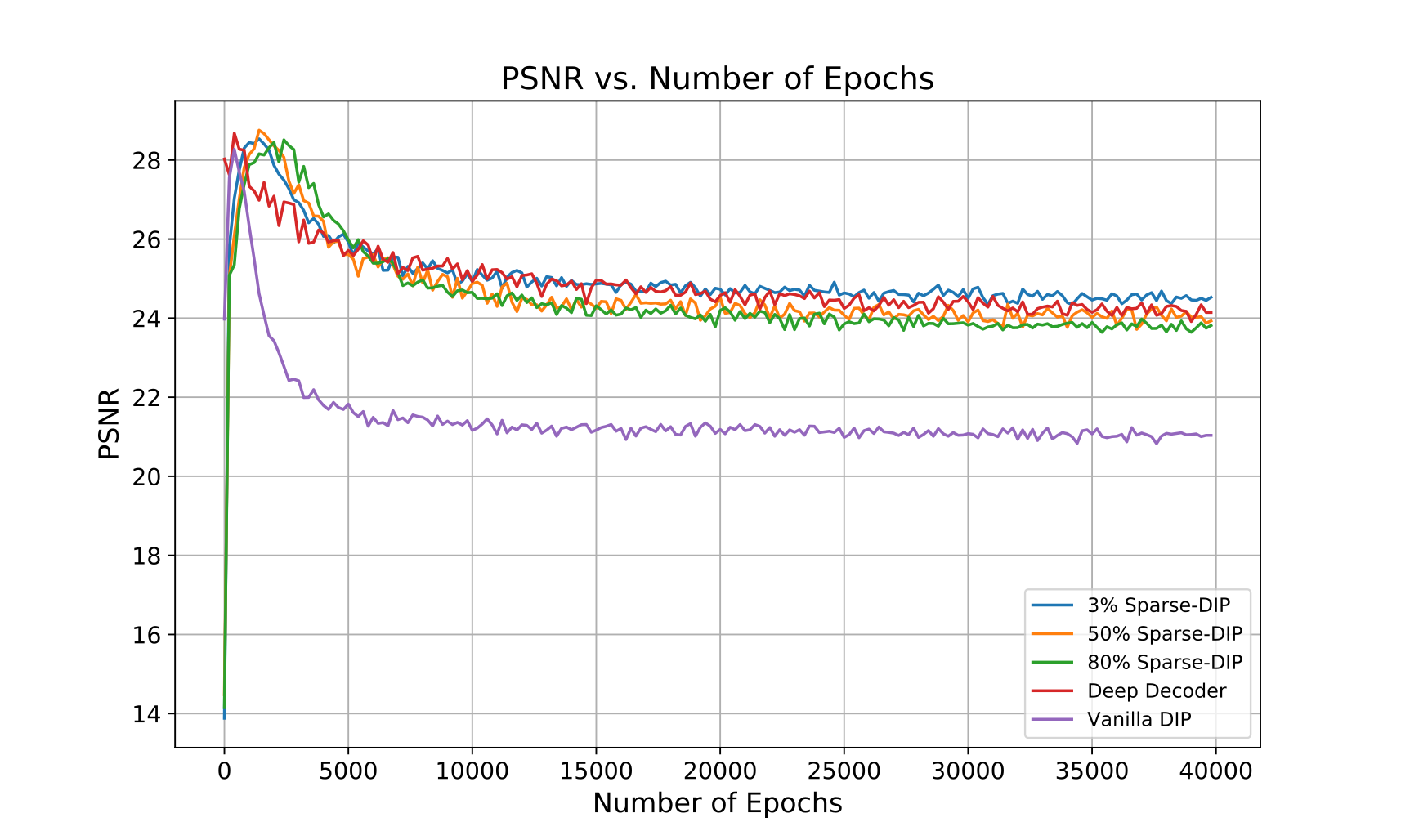}
        \subcaption{Face}
        \label{fig:new_figure_2}
    \end{minipage}

    \caption{Denoising performances ($\sigma=25dB$) of OES at 3 sparsity levels (3\%,50\%,80\%) and comparison to underparamterized deep decoder and overparameterized dense DIP. We observe that the peak performance of vanilla DIP is comparable with the final convegence of sparse-DIP.}
    \label{set14-den-figures}
\end{figure*}

In this section, we report the denoising performance for all the images in the 3 datasets.  In Table-\ref{tab:compact_table}, we report the number of parameters used in each network. In Table-\ref{main_table_denoise}, we report the PSNR at convergence for images across 3 datasets. We further plot the PSNR convergence curves of a subset of these images in Figure-\ref{set14-den-figures}. In these figures, we want to emphasize that dense DIPs overfit to noise at convergence. With Sparse-DIP's obtained at OES, the overfitting is reduced by a large extent. We also observe that $80\%$-sparse DIP is more prune to overfitting than $3\%$-sparse DIP. 

\begin{table}[h]
\centering
\caption{Number of parameters count of sparse and dense networks. Number of pixels in image is $512 \times 512 \times 3 = 786432$ (0.7M)}
\footnotesize
\setlength\tabcolsep{4pt}
\begin{tabular}{cccccc}
\toprule
\textbf{Model} & \textbf{Dense} \newline \textbf{DIP} & \textbf{Dense Decoder} \newline  & \textbf{Sparse-decoder }(50\%)  \newline & \textbf{Sparse-DIP} \newline ($3 \%$) & \textbf{Sparse-DIP} \newline ($4 \%$) \\
\midrule
\textbf{Number of parameters} & 3008867\newline (3M) & 100224 \newline (0.10M) & 50112 \newline (0.05M) & 90217 \newline (0.09M) & 120354 \newline (0.12M) \\
\bottomrule
\end{tabular}
\label{tab:compact_table}
\end{table}

\begin{table}[ht]
\footnotesize
\caption{Denoising capabilities comparison without early stopping on Set-14 dataset. The decoder has 100,224 parameters. Dense DIP has 3,008,867 parameters. Sparse networks are $3\%$ sparse and have 90217 parameters. The PSNR values are noted at the end of convergence of training after 40000 epochs. The average of three runs using different random seeds are noted. For each implementation, a random $\z$ and network initialization is used for evaluation. For $X_{y}$, $X$ denotes the average of three runs and $y$ denotes the standard deviation. }
\setlength\tabcolsep{1pt}
\centering
\small 
\begin{tabular}{|c|c|c|c|c||c|c|c|c||c|c|c|c|}
\hline
\multicolumn{5}{|c||}{$\sigma = 25dB$} & \multicolumn{4}{c||}{$\sigma = 12dB$} & \multicolumn{4}{c|}{$\sigma = 17dB$} \\ \hline
Image & Dense & GP & Deep & Sparse- & Dense & GP & Deep & Sparse- & Dense & GP & Deep & Sparse- \\ 
 & DIP & DIP & Decoder & DIP & DIP & DIP & Decoder & DIP & DIP & DIP & Decoder & DIP \\
 & &  & & & &  & & & &  & & \\ \hline
Pepper  & $21.21_{0.22}$ & $25.14_{0.29}$ & $27.01_{0.25}$ & \textbf{27.45}$_{0.22}$ & 27.34$_{0.18}$ & 29.17$_{0.26}$ & 28.81$_{0.18}$ & \textbf{30.41}$_{0.36}$ & 24.39$_{0.39}$ & 26.02$_{0.16}$ & 27.52$_{0.32}$ & \textbf{28.92}$_{0.22}$\\ \hline

Foreman & $20.69_{0.17}$ & 21.84$_{0.29}$ & 24.20$_{0.24}$ & \textbf{25.15}$_{0.13}$ & 26.59$_{0.07}$ & 28.33$_{0.05}$ & 29.81$_{0.08}$ & \textbf{30.55}$_{0.14}$ & 23.71$_{0.22}$ & 25.14$_{0.17}$ & 27.14$_{0.28}$ & \textbf{27.70}$_{0.16}$ \\ \hline

Flowers & 22.27$_{0.18}$ & 23.76$_{0.35}$ & 27.09$_{0.22}$ & \textbf{27.10}$_{0.18}$ & 28.55$_{0.17}$ & 30.78$_{0.14}$ & 30.33$_{0.09}$ & \textbf{31.07}$_{0.25}$ & 25.51$_{0.13}$ & 27.48$_{0.11}$ & 29.21$_{0.25}$ & \textbf{29.39}$_{0.22}$ \\ \hline
Comic  & 20.63$_{0.22}$ & 21.56$_{0.29}$ & 23.22$_{0.12}$ & \textbf{24.03}$_{0.21}$ & 26.45$_{0.36}$ & 28.27$_{0.36}$ & 28.01$_{0.06}$ & \textbf{28.57}$_{0.09}$ & 23.68$_{0.03}$ & 24.90$_{0.18}$ & 25.82$_{0.02}$ & \textbf{26.53}$_{0.19}$ \\ \hline 
Lena & 21.28$_{0.39}$ & 22.76$_{0.22}$ & \textbf{26.80}$_{0.35}$ & 26.40$_{0.29}$ & 27.50$_{0.28}$ & 29.48$_{0.19}$ & \textbf{30.96}$_{0.24}$ & 30.89$_{0.43}$ & 24.52$_{0.11}$ & 26.30$_{0.18}$ & \textbf{29.33}$_{31}$ & 29.03$_{13}$ \\ \hline
Barbara & $23.90_{0.10}$ & $23.49_{0.35}$ & $25.30_{0.09}$ & $\textbf{26.50}_{0.37}$ & $28.27_{0.20}$ & $\textbf{30.81}_{0.25}$ & $27.50_{0.31}$ & $27.81_{0.21}$ & $25.15_{0.13}$ & $27.55_{0.08}$ & $26.65_{0.04}$ & $\textbf{27.63}_{0.30}$\\ \hline
Monarch & $23.62_{0.36}$ & $23.35_{0.14}$ & $\textbf{27.87}_{0.33}$ & $27.62_{0.03}$ & $28.25_{0.05}$ & $31.17_{0.24}$ & $32.00_{0.22}$ & $\textbf{32.12}_{0.17}$ & $25.14_{0.14}$ & $27.20_{0.25}$ & $30.29_{0.18}$ & $\textbf{30.42}_{0.30}$ \\ \hline
Baboon  & $21.68_{0.14}$ & $23.04_{0.31}$ & $22.93_{0.18}$ & $\textbf{24.00}_{0.29}$ & $27.27_{0.39}$ & $27.03_{0.22}$ & $24.12_{0.21}$ & $\textbf{25.91}_{0.04}$ & $24.80_{0.11}$ & $25.57_{0.20}$ & $23.78_{0.27}$ & $\textbf{25.04}_{0.32}$ \\ \hline

Ppt3 & $24.07_{0.35}$ & $24.57_{0.39}$ & $26.81_{0.35}$ & $\textbf{26.90}_{0.22}$ & $28.88_{0.10}$ & $31.94_{0.13}$ & $31.73_{0.18}$ & $\textbf{32.41}_{0.17}$ & $25.85_{0.21}$ & $28.78_{0.20}$ & $29.49_{0.23}$ & $\textbf{29.96}_{0.22}$ \\ \hline
Coastguard & $20.53_{0.01}$ & $21.23_{0.21}$ & $23.71_{0.20}$ & $\textbf{24.19}_{0.06}$ & $26.50_{0.07}$ & $28.13_{0.10}$ & $29.43_{0.17}$ & $\textbf{30.60}_{0.11}$ & $23.54_{0.04}$ & $24.53_{0.14}$ & $26.36_{0.08}$ & $\textbf{27.09}_{0.35}$\\ \hline
Bridge & $21.77_{0.31}$ & $25.07_{0.02}$ & $25.55_{0.26}$ & $\textbf{26.12}_{0.30}$ & $28.58_{0.20}$ & $30.47_{0.10}$ & $28.10_{0.09}$ & $\textbf{29.23}_{0.31}$ & $25.31_{0.08}$ & $28.17_{0.42}$ & $27.04_{0.28}$ & $\textbf{28.08}_{0.38}$ \\ \hline

Zebra & $21.94_{0.08}$ & $23.46_{0.02}$ & $27.37_{0.19}$ & $\textbf{27.40}_{0.29}$ & $28.45_{0.17}$ & $30.93_{0.20}$ & $30.81_{0.12}$ & $\textbf{31.54}_{0.20}$ & $25.25_{0.38}$ & $27.39_{0.34}$ & $29.21_{0.29}$ & $\textbf{29.42}_{0.05}$ \\ \hline
Face & $21.03_{0.07}$ & $21.76_{0.30}$ & $24.32_{0.11}$ & $\textbf{24.53}_{0.22}$ & $26.90_{0.02}$ & $27.81_{0.37}$ & $29.93_{0.36}$ & $\textbf{29.93}_{0.02}$ & $24.10_{0.06}$ & $24.96_{0.12}$ & $27.01_{0.25}$ & $\textbf{27.23}_{0.38}$\\ \hline
Man & $21.98_{0.31}$ & $24.18_{0.10}$ & $26.27_{0.33}$ & $\textbf{26.59}_{0.39}$ & $28.45_{0.39}$ & $31.22_{0.19}$ & $29.84_{0.25}$ & $\textbf{30.94}_{0.31}$ & $25.12_{0.26}$ & $28.63_{0.29}$ & $28.77_{0.20}$ & $\textbf{29.11}_{0.13}$ \\ \hline

\end{tabular}
\label{main_table_denoise}
\end{table}

\begin{figure}[ht]
    \centering
    \begin{subfigure}[b]{0.4\textwidth} 
        \centering
        \includegraphics[width=\textwidth]{arxiv_figures/psnr_comb0.png}
        \caption{Pepper (Set-14 dataset)}
        \label{fig:sub1}
    \end{subfigure}
    \begin{subfigure}[b]{0.4\textwidth} 
        \centering
        \includegraphics[width=\textwidth]{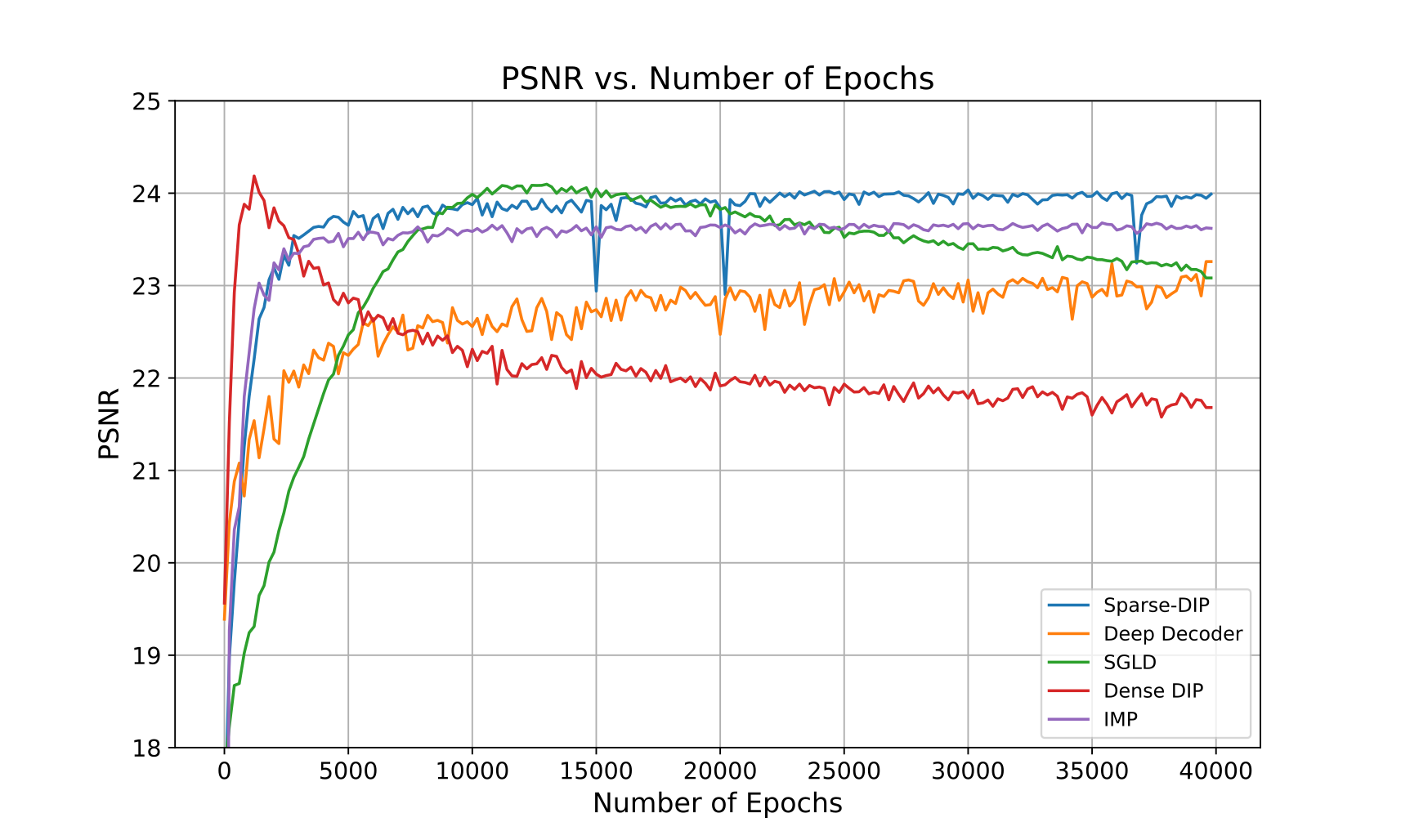}
        \caption{Baboon (Set-14 dataset)}
        \label{fig:sub2}
    \end{subfigure}
    \begin{subfigure}[b]{0.4\textwidth} 
        \centering
        \includegraphics[width=\textwidth]{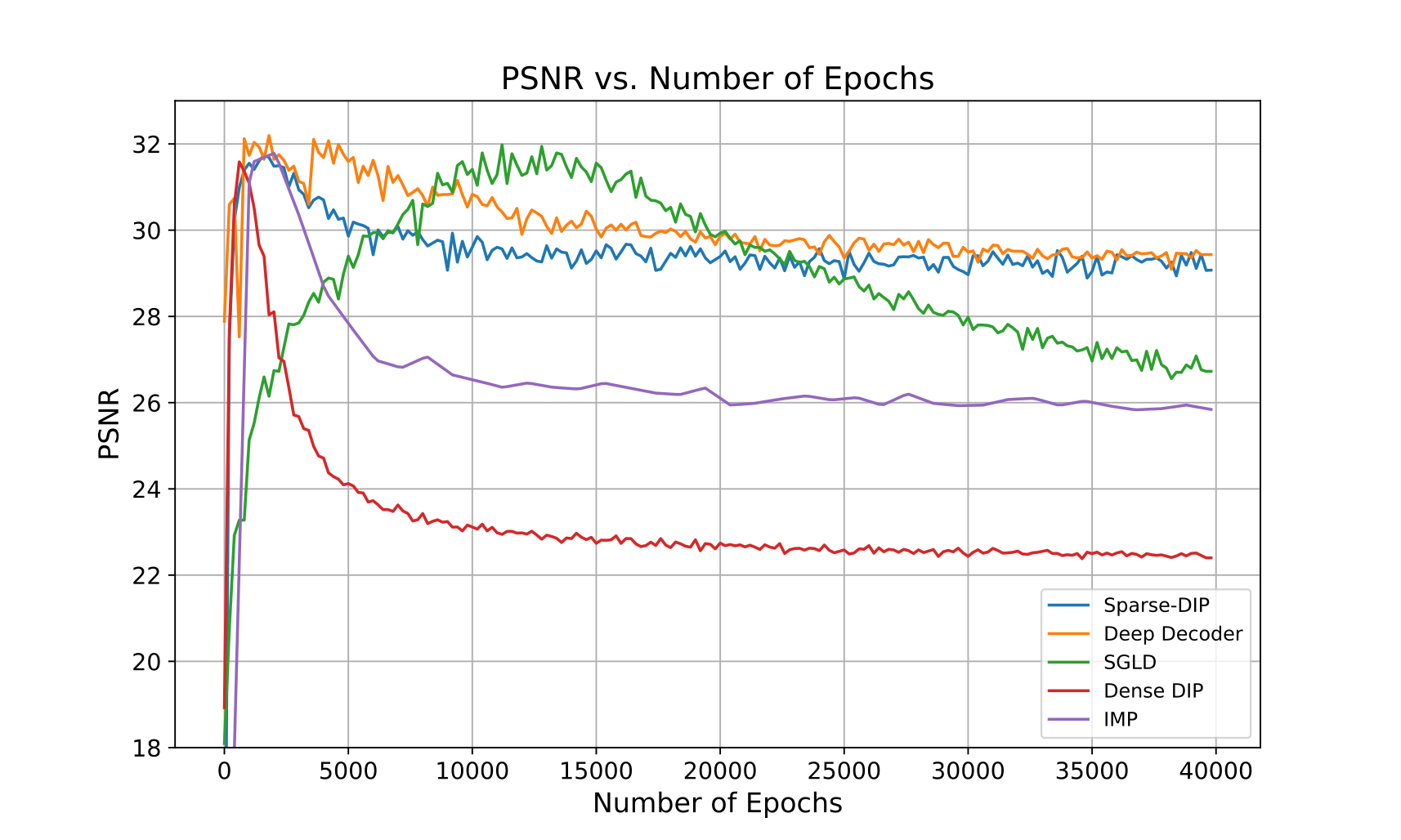}
        \caption{Face-2 (Face dataset)}
        \label{fig:sub3}
    \end{subfigure}
    \begin{subfigure}[b]{0.4\textwidth} 
        \centering
        \includegraphics[width=\textwidth]{arxiv_figures/psnr_1_comb_dataset.png}
        \caption{Door (Standard Dataset)}
        \label{fig:sub4}
    \end{subfigure}
    \caption{Denoising results of various methods on noisy images ($\sigma=25$ dB) across 3 popularly used datasets.}
    \label{fig:combined_denoise_appendix}
\end{figure}

\begin{table}[ht]
\centering
\small
\caption{Denoising capabilities comparison without early stopping for $\sigma=25$dB on Face Dataset and Standard dataset.}
\begin{tabular}{c}
\begin{minipage}{.5\linewidth}
\centering
\subcaption{Standard Dataset}
\begin{tabular}{|c|c|c|c|c|}
\hline
\multicolumn{5}{|c|}{$\sigma = 25dB$} \\ \hline
Image & Dense & GP & Deep & Sparse \\
 & DIP & DIP & Decoder & DIP \\ \hline
Flight & 20.49 & 22.02 & 23.99 & \textbf{24.02}\\  \hline
House & 21.88 & 23.30 & 28.35 & \textbf{29.27}\\ \hline
Building & 21.93 & 23.55 & 27.23 & \textbf{27.23}\\ \hline
Door & 21.85 & 23.31 & 27.02 & \textbf{28.18}\\  \hline
Hats & 21.76 & 24.12 & 24.86 & \textbf{26.07}\\ \hline
\end{tabular}
\end{minipage}

\begin{minipage}{.5\linewidth}
\centering
\subcaption{Face Dataset}
\begin{tabular}{|c|c|c|c|c|}
\hline
\multicolumn{5}{|c|}{$\sigma = 25dB$} \\ \hline
Image & Dense & GP & Deep & Sparse \\
 & DIP & DIP & Decoder & DIP \\ \hline
Face-1 & 22.40 & 26.72 & 28.90 & \textbf{29.07}\\  \hline
Face-2 & 22.02 & 26.02 & 29.50 & \textbf{29.58}\\ \hline
Face-3 & 21.96 & 25.88 & \textbf{28.27} & 27.91\\ \hline
Face-4 & 21.83 & 26.37 & \textbf{28.31} & 27.89\\  \hline
\end{tabular}
\end{minipage}
\end{tabular}
\end{table}

\section{Comparison with Standard Pruning Methods}
\label{prune_comp_tab}
In section-\ref{finding-2}, we briefly showed some results on comparison of OES with standard pruning methods that comprised of pruning at Initialization methods like Synflow, Grasp and magnitude/random based pruning and pruning after training methods like Iterative magnitude pruning. In Table-\ref{pruning_methods}, we show all the results for images in three different datasets: Set-14, Face, and Standard image. All the PSNR values were noted at convergence. Our observation suggests that OES outperforms the traditional pruning methods at initialization. We did not report the performance of SNIP as it resulted in layer collapse for Unet. We see that magnitude and random choice of parameters serve as a bad indication of importance score and most often than not leads to layer-collapse. We explored this part in the manuscript in Section \ref{finding-2}. Synflow, Grasp pruning at initialization leads to overfitting when run for longer iterations. Lastly, our comparison with IMP (Iterative magnitude pruning), shows that using the mask obtained from converged DIP training easily leads to overfitting of the masked subnetwork.  We implement IMP with two schedules: IMP-$(0.8)^{14}$ denotes pruning and training was run for 14 iterations and at each iteration $20\%$ of the remaining weights were pruned, IMP-$(0.2)^{3}$ denotes pruning and training was run for 3 iterations and at each iteration $80\%$ of the remaining weights were pruned. Having gradual pruning performed better when compared to aggressive pruning. This further shows that runnign IMP to get good masks can be costly since we need to run more iterations of pruning to reach a desired sparsity level. 

\begin{table}[ht]
\centering
\footnotesize 
\caption{Denoising capabilities ($\sigma=25dB$) comparison without early stopping for standard pruning methods for images from 3 different datasets. PAI refers to Pruning At Initialization. PAT refers to Pruning After Training. All networks have sparsity level of $5\%$. IMP refers to Iterative Magnitude Pruning (IMP) with weight rewinding. IMP-$(1-p)^n$ denotes at each pruning iteration $p \%$ of weights have been deleted and has been run for n number of pruning iterations.  None of the methods use clean image for training. }
\textbf{Set-14}
\begin{tabular}{|l||c|c|c|c||c|c||c|}
\hline
\multicolumn{1}{|c||}{Image} & \multicolumn{4}{c||}{PAI} & \multicolumn{2}{c||}{PAT} & \multicolumn{1}{c|}{Ours (PAI)} \\ \hline
 & GraSP & SynFlow & Magnitude & Random & IMP-$(0.8)^{14}$ & IMP-$(0.2)^{3}$ & OES \\ \hline
Pepper   & 22.22 & 22.07  & 12.42 & 10.80 & 25.52 & 25.66 &  \textbf{27.45}\\ \hline
Foreman   & 21.67 & 20.93 & 12.13  & 10.75 & 21.66 & 23.78  & \textbf{25.15}  \\ \hline
Flowers   & 23.02 & 23.07 & 12.22 & 10.61 & 26.11 & 26.43 & \textbf{27.10} \\ \hline
Comic   & 21.07 & 21.50 & 12.13 & 11.75 & 22.13  & 22.42 & \textbf{ 24.03} \\ \hline
Lena   & 13.39 & 22.19  & 14.37 & 13.33 & 25.73 & 25.75 &\textbf{ 26.40} \\ \hline
Barbara   & 23.45 & 23.51  & 13.56 & 13.03 & 26.20 & 26.05 & \textbf{26.50} \\ \hline
Monarch   &22.67  & 22.93 & 14.52 & 12.73 & 26.27 & 26.52  & \textbf{ 27.62} \\ \hline
Baboon   & 22.42 & 22.56 & 12.50 & 11.61 & 23.75 & 23.49 & \textbf{24.00 } \\ \hline
Ppt3  & 23.42 & 24.23  & 9.54  & 8.51 & 26.34 & 26.22 & \textbf{26.90} \\ \hline
Coastguard   & 20.78 & 20.73  & 13.31 & 13.16 & 21.38 & 21.90 & \textbf{24.19}  \\ \hline
Bridge   & 24.25 & 23.29  & 13.36 & 13.10 &\textbf{ 26.27} & 26.20  & 26.12 \\ \hline
Zebra   & 22.92 & 23.22 & 13.29 & 12.34 & 26.56  & 26.54 & \textbf{27.40} \\ \hline
Face   & 21.38 & 21.14 & 10.74 & 9.50  & 21.80 & 22.50 & \textbf{24.53} \\ \hline
Man   & 23.62 & 23.71 & 12.89 & 11.30 & \textbf{26.82} & 26.72 & 26.59 \\ \hline
\end{tabular}

\vspace{5mm}

\textbf{Face}
\begin{tabular}{|l||c|c|c|c||c|c||c|}
\hline
\multicolumn{1}{|c||}{Image} & \multicolumn{4}{c||}{PAI} & \multicolumn{2}{c||}{PAT} & \multicolumn{1}{c|}{Ours (PAI)} \\ \hline
 & GraSP & SynFlow & Magnitude & Random & IMP-$(0.8)^{14}$ & IMP-$(0.2)^{3}$ & OES \\ \hline
Face-1  & 22.88 & 22.97 & 12.64 & 8.41 & 26.89 & 26.62 & \textbf{29.07}\\ \hline
Face-2  & 22.64 & 22.90 & 12.35 & 10.40 & 26.74 & 26.19 & \textbf{29.58}  \\ \hline
Face-3  & 22.74 & 22.80 & 13.46 & 11.86 & 26.94 & 26.50 & \textbf{27.91} \\ \hline
Face-4  & 22.71 & 22.57 & 12.16 & 11.61 & 26.33 & 26.46 & \textbf{27.89} \\ \hline
\end{tabular}

\vspace{5mm}

\textbf{Standard}
\begin{tabular}{|l||c|c|c|c||c|c||c|}
\hline
\multicolumn{1}{|c||}{Image} & \multicolumn{4}{c||}{PAI} & \multicolumn{2}{c||}{PAT} & \multicolumn{1}{c|}{Ours (PAI)} \\ \hline
 & GraSP & SynFlow & Magnitude & Random & IMP-$(0.8)^{14}$ & IMP-$(0.2)^{3}$ & OES \\ \hline
House & 20.46 & 20.24 & 13.51 & 13.20 & 26.61 & 26.88 & \textbf{29.27}\\ \hline
Building & 22.72 & 22.52 & 15.32 & 13.23 & 26.30 & 26.02 & \textbf{27.23}  \\ \hline
Door & 21.87 & 21.80 & 12.32 & 10.49 & 26.80 & 26.46 & \textbf{28.18} \\ \hline
Hats & 22.43 & 21.60 & 11.20 & 12.45 & 25.97 & 25.92 & \textbf{26.07} \\ \hline
\end{tabular}

\vspace{5mm}
\label{pruning_methods}
\end{table}

\section{Details of the Gumbel Softmax Reparameterization Trick}
\label{gumb_softmax_det}
Let $s$ be the final number of non-zero elements we want to have in the subnetwork and $d$ is the total number of parameter. Then we fix the prior to be $\pr_{0} =\frac{s}{d} \times \mathbf{1}$, which means each parameter will have a prior probability $p_{0}$ for selecting the weight.  We solve the following optimization problem using the Gumbel softmax reparameterization trick, but first we explain the challenges of solving this optimization problem:

\begin{equation}
\begin{aligned}
\m^{*}(\y) &= C(\pr^{*})  \quad \text{such that} \quad \\
& \pr^{*} = \arg \min_{\pr} \underbrace{\mathbb{E}_{\m \sim Ber(\pr) }  \left[ || G(\p_{in} \circ \m,\z) - \y||_{2}^2 \right]}_{R(\pr)} \\
& + \lambda KL(Ber(\pr)||Ber(\pr_{0}))
\end{aligned}
\end{equation}

The standard way to minimize $R(\pr)$ is to obtain a direct Monte Carlo estimate of $\partial_{p_{i}} R(\pr)$ for every $i=1,2,..,d$ by several random realizations of the network. Let $Q := Ber(\pr)$ denote the posterior distribution. Then for every $i$, let $e_{i}(m'_{i}) = \mathbb{E}_{Q} \left[ || G(\p_{in} \circ \m,\z) - \y||_{2}^2 | m_{i} = m'_{i}\right] $, we have $R(\pr) = p_{i}e_{i}(1) + (1-p_{i})e_{i}(0) $, which yields  $\partial_{p_{i}} R(\pr) = e_{i}(1) -e_{i}(0)$. Finding the Monte Carlo estimate of $\partial_{p_{i}} R(\pr)$ is computationally infeasible because of computing the conditional expectation for every $i$. 
The loss $R(\pr)$ depends on $\pr$ in an implicit way and calculating the gradient $\partial_{\pr}R(\pr)$ using Monte Carlo samples is not straightforward.  

To make the relation of the loss $R(\pr)$ and variable $\pr$ explicit for gradient based methods, a classical approach called the \textit{reparameterization trick} is used to find the mapping that makes it explicit. For discrete Bernoulli distribution, the reparamterization trick is called the Gumbel-Max (GM) trick which is a method of sampling from discrete random variables using explicit dependence on the probabilities of each state. The GM trick allows straightforward simulation of discrete variables, but it is not practical for gradient computing because it involves differentiation through a max function. To overcome this disadvantage of GM trick,  \citet{maddison2016concrete} introduced the Gumbel-Softmax trick which relaxes the discrete distribution to CONCRETE distribution: CONtinuous relaxations of disCRETE random variables. 
Let $T \in \mathbb{R}^{+}$ denote the temperature which controls the degree of relaxation from the discrete distribution to the continuous distribution. The sampling from the Concrete distribution $Concrete(p_{i},1-p_{i})$ is as follows:
\begin{enumerate}
    \item fix $T$ and sample $G_{k},G_{l} \sim$ Gumbel i.i.d  ($-\log(-\log(U[0,1]))$). 
    \item set $\hat{m_{i}}(p_{i}) = \frac{ \exp{(\frac{(\log(p_{i})+ G_{l})}{T})}}{ \exp{(\frac{(\log(p_{i})+ G_{l})}{T})} + \exp{(\frac{(\log(1-p_{i})+ G_{k})}{T})}}  \sim Concrete(p_{i},1-p_{i})$  for $i=1,2,..,d$. 

\end{enumerate}
 Here, $\hat{\m}$ obeys a \textit{continuous} Concrete distribution denoted as Concrete($\pr$,$\mathbf{1}-\pr$) instead of discrete Bernoulli distribution $Ber(\pr)$. This continuous approximation of discrete distribution is controlled by the temparature variable $T$. As $T$ tends to 0, Concrete($\pr$,$\mathbf{1}-\pr$)  distribution converges to the $Ber(\pr)$ distribution, however, for small $T$, there are numerical instability issues in estimating $\hat{m}_{i}$. In  our experiments, $T$ is fixed to 0.2 for all the experiments, as it gives a good approximation to the discrete distribution without suffering from the numerical instability issue. Note that unlike in Bernoulli distribution $\m \sim Ber(\pr)$, where the dependence of $R(\pr)$ on $\pr$ was implicit, making \eqref{ber_mask} challenging to optimize, for Concrete distribution the dependence on $\pr$ is explicit, making it amenable to solve by gradient based optimizers. So, given random network initialization, $\p_{in}$, and the noisy corrupted image $\y$, the steps to learn mask $\m$ to the model parameters are as follows.
     
\begin{algorithm} 
\caption{Optimal Eye Surgeon (Learning Mask at initialization)}
\begin{algorithmic}[1]
\STATE {\bfseries Input:} $\p_{in}$, $\pr_{0}$, $\y$, $G(.,\z),C(.), $ number of samples $K$
\STATE {\bfseries Output:} Final mask $\m^*(\y)$
\STATE Initialize $\pr = 0.5 \times \mathbf{1}$, set $T = 0.2, \lambda = 1e-9$
\FOR{each iteration}
      
    \FOR{$k = 1$ to $K$}
        \STATE $\hat\m^k(\pr) \gets \text{Concrete}(\pr, 1-\pr)$
        \STATE $L^k(\pr) \gets \| G(\p_{in} \circ \hat\m^k(\pr), z) - \y \|_2^2$
    \ENDFOR
    \STATE $L_{C}(\pr) \gets \frac{1}{K} \sum_{k=1}^K L^k(\pr) + \lambda \text{KL}\left(\text{Ber}(\pr)||\text{Ber}(\pr_{0})\right)$
    \STATE Compute $\nabla_{\pr}L_{C}(\pr)$, do GD : $ \pr \gets \pr - \eta \nabla_{\pr}L_{C}(\pr)$
\ENDFOR
\STATE $\m^*(\y) \gets C(\pr^*)$, where $\pr^*$ is the converged probability mean.
\end{algorithmic}
\label{algo:mask}
\end{algorithm}

While optimizing \eqref{ber_mask} by Algorithm-\ref{algo:mask}, we reparameterize the optimization variable $\pr$ through a sigmoid function $\pr = sigmoid(\vr)$, which maps the domain of the variable $\pr$ from $[0,1]$ to the optimization variable $\vr$ :$[-\infty,\infty]$. So our initialization, which ensures unbiased selection of weights is at $\vr = sigmoid^{-1}(\pr) = sigmoid^{-1}(0.5)= 0$. The prior probability which controls sparsity, is also related as  $\vr_{0} = sigmoid^{-1}(\pr_{0})$ where $\pr_{0}$ is the prior probability vector. This reparameterization of the optimization variable ensures that the variable domain is not restrictive. 

Once $\pr^{*}$ is obtained by gradient descent, mask $\m^*(\y) \gets C(\pr^*)$ is obtained by ranking the elements of $\pr^{*}$ and setting the indices of $\m^*(\y)$ corresponding to the top $s \%$ values of $\pr^{*}$ to be 1, and 0 otherwise. This way the sparsity of the mask is set to be the desired sparsity $s$ and is accomplished by the $C(.)$ function. $C(.)$ is a ranking function, which ranks the values of $\pr$ and then thresholds the weight indices corresponding to $s \%$  highest values of $\pr$ to achieve the desired sparsity. 
We chose the initialization $\pr = 0.5 \times \mathbf{1}$, so that there is no bias towards any weight selection and all weights have equal probability of selection/pruning at initialization. Although with prior knowledge, for certain layers $\pr$ can be initialized to higher probability values, but in our preliminary experiments, we do not introduce bias towards any weights in any layers.

\section{Differences with Related Work}
\label{related_works}
While we provide an interesting insight on the image generation capability of hour-glass Unet architecture, we acknowledge the existence of previous works which further substantiate our current findings. In the following points, we highlight the difference of our work with the following and also mention the similarity of the findings:

\subsection{Comparison to NAS-DIP \citep{chen2020dip} and ISNAS-DIP \citep{arican2022isnas}}

NAS-dip proposes to apply the NAS (Neural architecture search) algorithm on DIP framework. They build a searching space for upsampling cells in the decoder and the skip connections between encoder and decoder. Then they leverage reinforcement learning (RL)
 with RNN controller and use \textit{PSNR wrt clean image} as reward to guide the architecture. After the network search, they transfer the best-performing architecture and optimize the model the same way as DIP. We highlight the points of difference:

 \begin{itemize}
     \item \textit{Architecture search vs. pruning}: \citet{chen2020dip} and \citet{arican2022isnas} search for the best architecture. The final architecture found by NAS-DIP is a dense architecture. Instead, we start with a dense deep Unet architecture and then make it sparse. Instead of searching for the best architecture combination, we focus on each weight parameter and evaluate it's importance in the context of image generation. Infact, a NAS-DIP model found by \citet{chen2020dip} can be further pruned by OES. 
     \item \textit{Limited search space}: NAS-DIP searches over only the upsampling and residual connections. For OES, a 6 layer encoder-decoder network is the base architecture and each parameter gets it's individual importance metric through learning $\pr$. We believe that although upsampling layers play a crucial role, the encoder layers can't be entirely discarded. 
     \item \textit{Using clean image to find architecture}: We want to emphasize this is the main point of difference between the previous works like \citet{chen2020dip,arican2022isnas,wu2023chasing} and ours. \textit{We do not need to use the clean image for pruning the network. Masking at initialization induces image prior even when trained against a corrupted image.} We discuss this phenomenon in detail in Finding-1. 
 \end{itemize}

\subsection{Comparison to The Devil is in the Upsampling \citep{liu2023devil}}

\citet{liu2023devil} proposed a heursitic strategy for designing appropriate architecture by analyzing the frequency response of architecture parts of DIP. Their observation was that the bilinear upsampling layers are the most important parts for image generation. Followed by the convolutional layers as they observed that only when the decoder part is used, convolutional decoders performed better than non-convolutional or MLP decoders. Furthermore, they suggest whether to increase/decrease depth or width or whether to keep skip connections (or not) based on signal processing intuition and sanity check based experiments. 
Our Alorithm OES relies on the mask learning algorithm to convey the similar information obtained in \citet{liu2023devil} and both these works agree on three findings.
\begin{enumerate}
    \item \textit{Importance of decoder}: In Figure-\ref{fig:subhist2}, we also find that given a hour-glass Unet architecture, the decoder part seems to be more important while the encoder part is more compressible. This is the main finding in \citet{liu2023devil} based on the frequency response of the upsampling layer. However, in OES, the final converged value of $\pr$ conveys this information. 
    \item \textit{Reduced depth in Unet}: For hour-glass architecture, the authors observe that increased depth can lead to oversmoothing of final image. Hence, for decoder architectures, the authors advocate reduced upsampling operations and for Unet architecture, they advocate decreasing the depth of the network. In Figure-\ref{fig:subhist2}, we see that the converged and thresholded value of $\pr$ conveys the same finding. For 6 layer Unet architecture, the middle layer of the encoder-decoder architecture seems to get pruned the most showing a 'W' shape in encoder-decoder hour glass architecture. This denotes that we can do with reduced depth. 
    \item \textit{Not using skip connections}: The authors notice that the skip connections ameliorate the oversmoothing issue when the network has large depth. Hence, they may lower the effective upsampling rate, making deep networks perform similarly to shallower ones. Thus in our base architecture, we use the simple hour-glass Unet architecture. Trying to understand and analyze OES with skip architecture can be more complicated and we leave it as future work. 
\end{enumerate}
We also want to highlight one point of difference in the findings between these two works. We observe that using an irregular pruned Hour-glass architecture outperforms deep decoder based architecture. Hence, although \textit{devil is in the upsampling layers}, we observe that the encoder-decoder junction also plays a crucial role. 

\subsection{Comparison to Lottery Image Prior \citep{wu2023chasing}}
\label{lottery_image}
The main message of our paper was to advocate learning the mask at initialization instead of learning the mask based on magnitude obtained post-training. However, we acknowledge that \citet{wu2023chasing} is the first work to apply unstructured pruning for image reconstruction based problems. However, they use the early stopping time to obtain the mask through Lottery Ticket Hypothesis. Generally speaking, identifying the early stopping time in image reconstruction tasks is itself a challenging task.  It is hard to come up with an estimate of an early-stopping time based on observations from different images and corruption levels. For example in Figure-\ref{fig:early_var}, we see that for two different images with two different corruption levels, the early stopping time can vastly vary. 
We also observe that in their python script in their \href{https://github.com/VITA-Group/Chasing-Better-DIPs}{github repo} they use the clean image to train the mask for a single image. 
In our experiments, we show the adverse effects of LTH based masks obtained at convergence. But further we also compare our method when LTH masks are obtained at early-stopping time or using clean images. We observe that LTH based masks obtained at early stop time perform well when the image used for training the mask is also used for denoising in Figure-\ref{fig:imp_early_oes}-a. But when a different image is used for denoising, the transferability of OES masks seems to be better (Figure-\ref{fig:imp_early_oes}-b). In Table-\ref{noise_lena}, we compare the transferability of OES masks and IMP based masks. Here, OES masks are obtained at initialization and LTH based masks are obtained by training the network to convergence but with a clean image. We study the pruning pattern of Unet architecture in details and compare OES and IMP methods, something that was not studied comprehensively \citep{wu2023chasing}. 

\section{Comparing IMP-based Denoising}
\label{imp-den}

\begin{figure}[!h]
    \centering
    \begin{minipage}{0.5\textwidth}
        \centering
        \includegraphics[width=\textwidth]{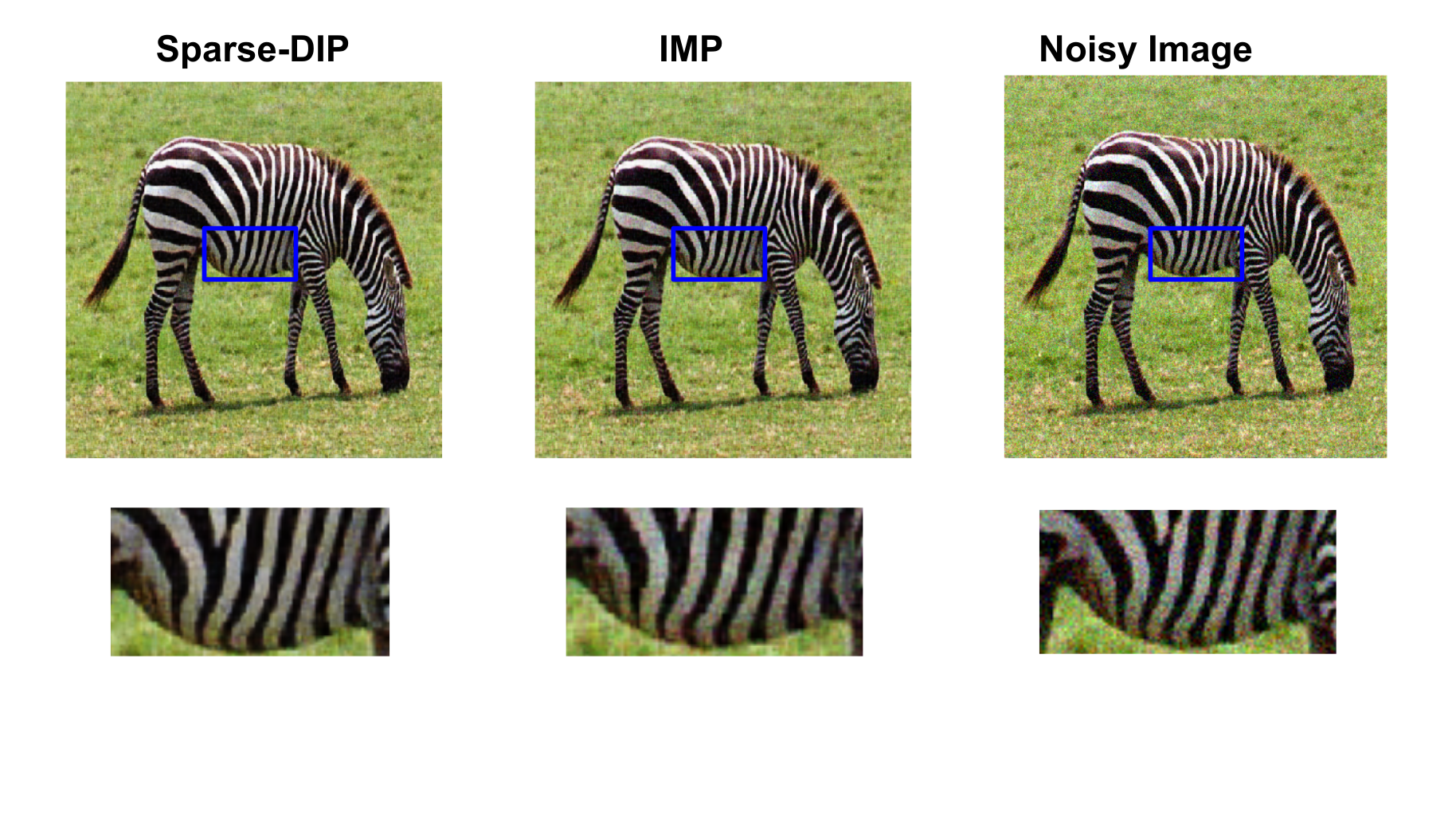}
        \subcaption{Denoising Zebra ($\y_{target}$ )}
        \label{fig:bridge}
    \end{minipage}\hfill
    \begin{minipage}{0.5\textwidth}
        \centering
        \includegraphics[width=\textwidth]{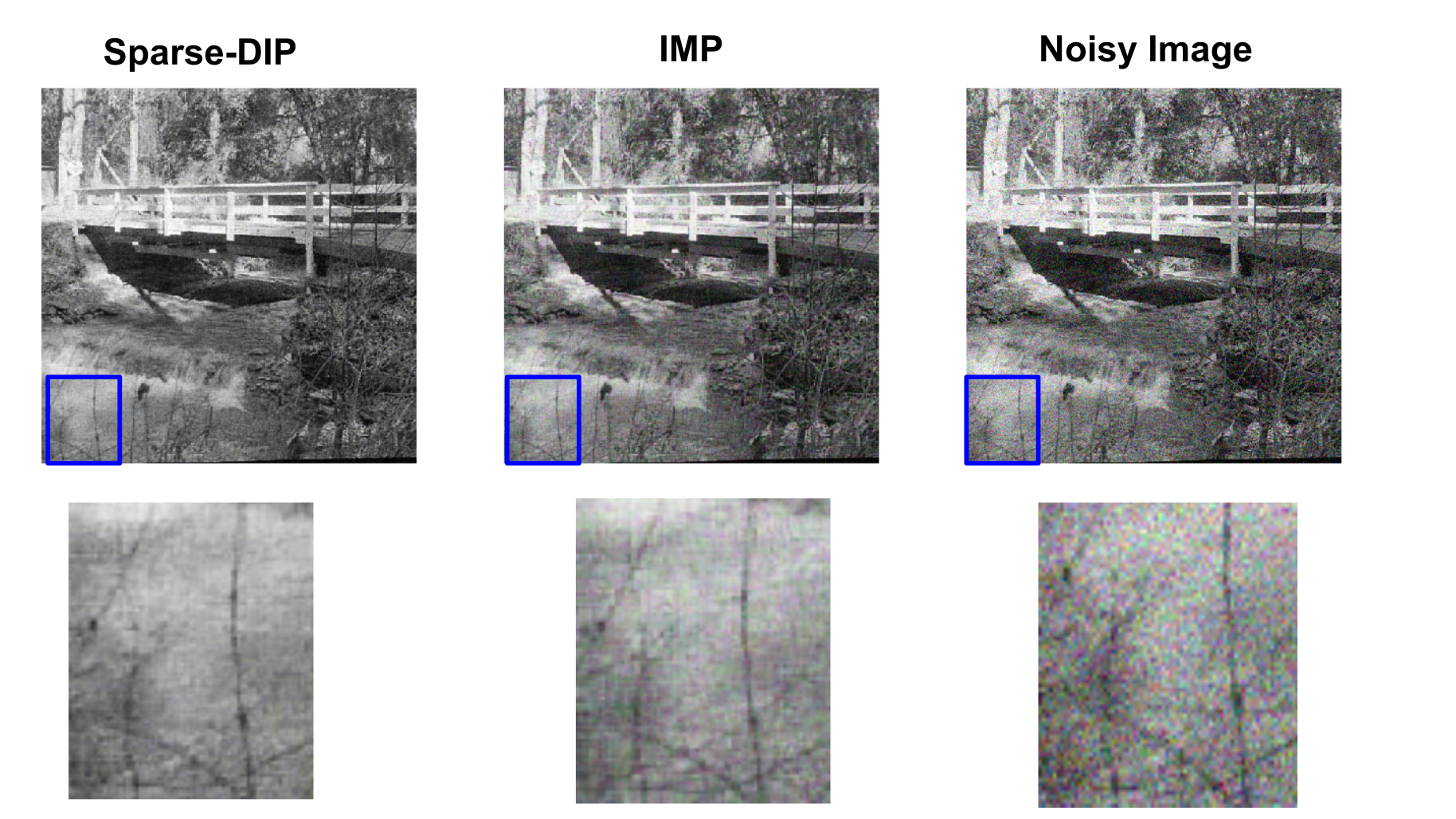}
        \subcaption{Denoising Bridge ($\y_{target}$ )}
        \label{fig:bridge}
    \end{minipage}\hfill
    \begin{minipage}{0.5\textwidth}
        \centering
        \includegraphics[width=\textwidth]{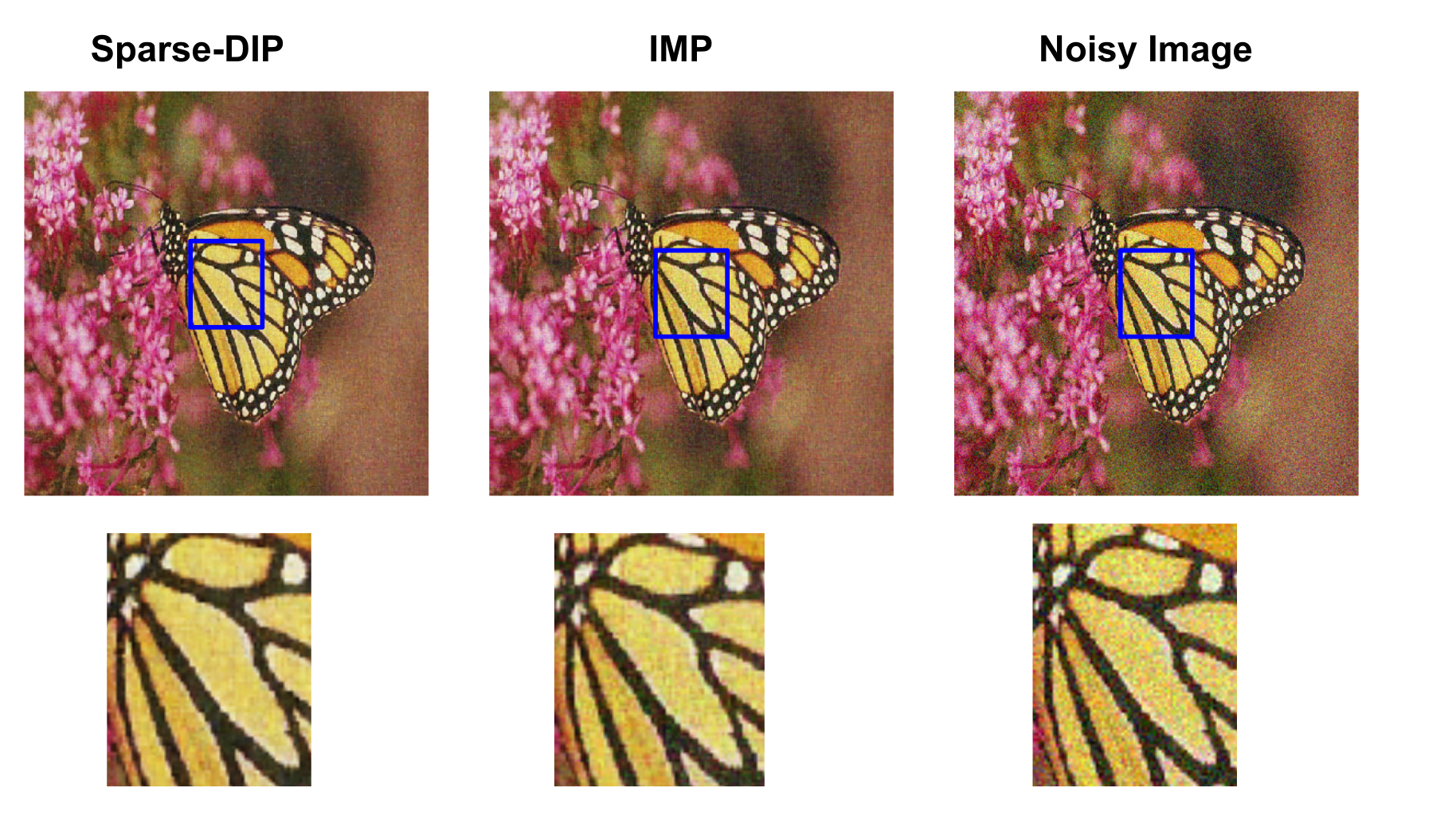}
        \subcaption{Denoising Monarch ($\y_{target}$ )}
        \label{fig:monarch}
    \end{minipage}\hfill
    \begin{minipage}{0.5\textwidth}
        \centering
        \includegraphics[width=\textwidth]{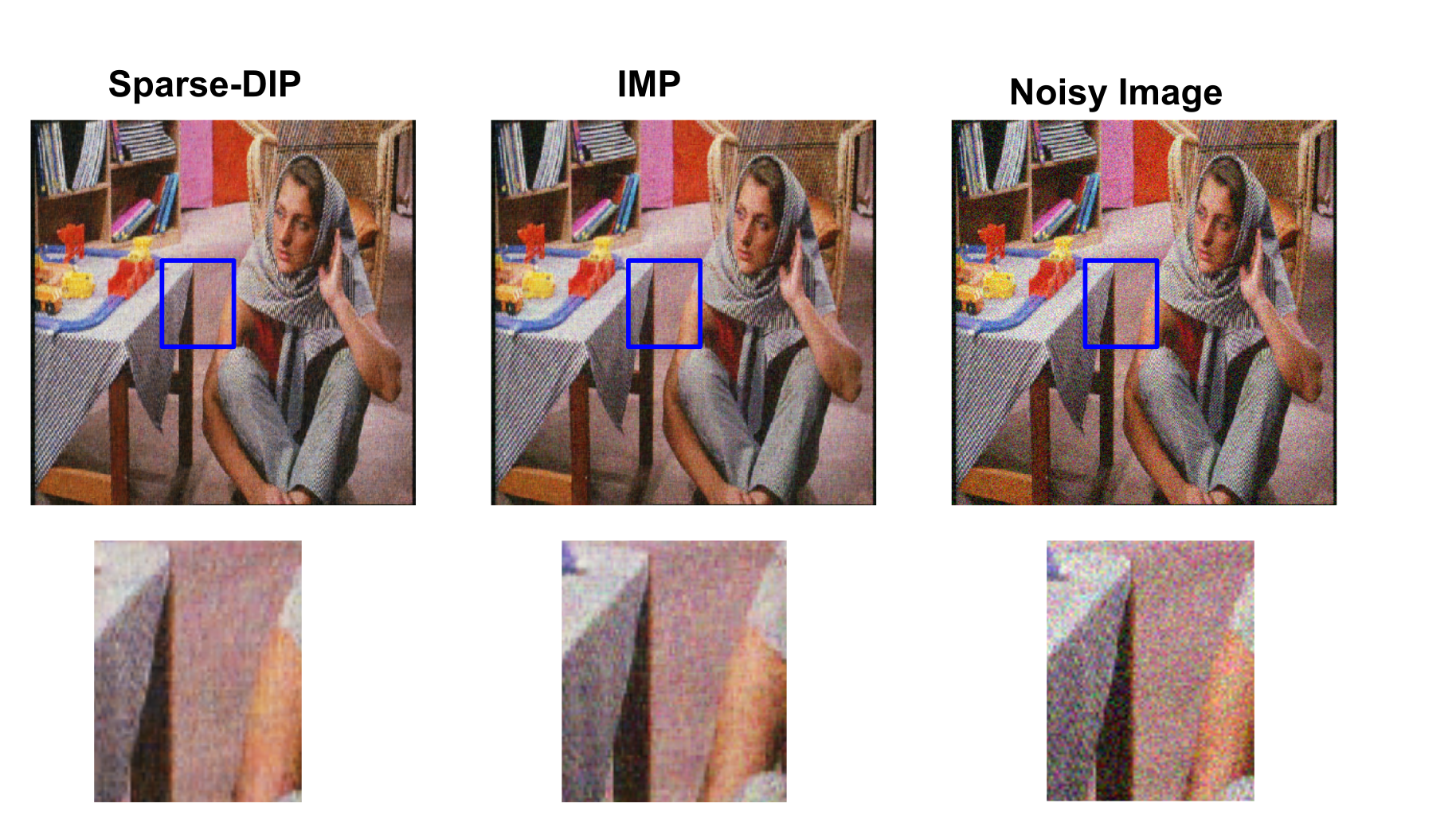}
        \subcaption{ Denoising Barbara ($\y_{target}$ )}
        \label{fig:barbara}
    \end{minipage}
    \caption{Comparing the denoising performance of transferred subnetworks found by OES vs subnetworks found by IMP in Set-14 dataset. Here $\y_{source}$ is the Lena image. Both masks are at sparsity level of $5\%$. IMP based subnetworks overfit to noise as shown in the zoomed version. All noisy images are corrupted with $\sigma=25dB$. The PSNR values are found in Table-\ref{noise_lena}. }
    \label{impvsoes-lena}
\end{figure}

In Finding-4, in the manuscript, we discussed the transferability of OES masks and compared how these masks transferred with the same image, within images of the same dataset and within images of varying datasets. Here, in this section, we report additional performance where we use the mask learned on Lena image (clean and noisy) at $5\%$ sparsity. In Table-\ref{noise_lena}, we compare the performance of OES masks with IMP masks at convergence for several noise levels. Here in Table-\ref{noise_lena}, the IMP masks were learned on the noisy Lena images. We demonstrate the corresponding figures in Figure-\ref{impvsoes-lena}. We run the denoising algorithm till 40k iterations. We see that in Figure-\ref{impvsoes-lena}, the IMP based masks overfit to noise, whereas OES-masks learned at initialization do not overfit. For this particular experiment, we do not use any knowledge of early-stopping time, so at convergence the parameters overfit to noisy Lena image. The IMP mask in Table-\ref{noise_lena} is obtained based on the magnitude of these parameters.   

\begin{table}[ht]
\centering
\caption{Comparison of denoising capabilities (for various noise levels) of transferred masks for OES vs IMP based pruning. $\y_{source}$ used is the \textbf{noisy Lena image}. All masks are $5\%$ sparse.}
\small 
\begin{tabular}{|l||c|c||c|c||c|c|}
\hline
\multicolumn{1}{|c||}{Image ($y_{target}$)} & \multicolumn{2}{c||}{$\sigma = 25dB$} & \multicolumn{2}{c||}{$\sigma = 12dB$} & \multicolumn{2}{c|}{$\sigma = 17dB$} \\ \hline
& $m(IMP)$ & $m(OES)$ & $m(IMP)$ & $m(OES)$ & $m(IMP)$ & $m(OES)$ \\ \hline
Pepper & 26.57 & \textbf{27.05} & 29.66  & \textbf{30.37} & 27.92 &  \textbf{28.55} \\ \hline
Flowers & 26.17 & \textbf{27.10} & 30.03 & \textbf{31.02} & 28.54 & \textbf{29.31} \\ \hline
Lena (self) & 25.85 & \textbf{26.35} & 29.36 & \textbf{30.95} & 28.45 & \textbf{28.89} \\ \hline
Barbara & 25.31 & \textbf{26.34} & 28.60 & \textbf{30.36} & 27.30 & \textbf{28.43} \\ \hline
Monarch & 26.45 & \textbf{27.38} & 31.01 & \textbf{32.84} & 29.14 & \textbf{30.23} \\ \hline
Baboon & 23.26 & \textbf{23.91} & 24.87 & \textbf{25.25} & 24.24 & \textbf{24.89} \\ \hline
Ppt3 & 26.11 & \textbf{26.96} & 30.92 & \textbf{32.32} & 29.05 & \textbf{29.57} \\ \hline
Bridge & 25.09 & \textbf{26.17} & 28.06 & \textbf{28.74} & 26.93 & \textbf{27.54} \\ \hline
Zebra & 26.34 & \textbf{27.20} & 30.54 & \textbf{31.45} & 28.80 & \textbf{29.87} \\ \hline
Man & 25.83 & \textbf{26.92} & 29.67 & \textbf{30.22} & 27.99 & \textbf{29.13} \\ \hline
\end{tabular}
\label{noise_lena}
\end{table}

\begin{figure*}[!h]
    \centering
    \begin{subfigure}[b]{0.48\textwidth} 
        \centering
        \includegraphics[width=\textwidth]{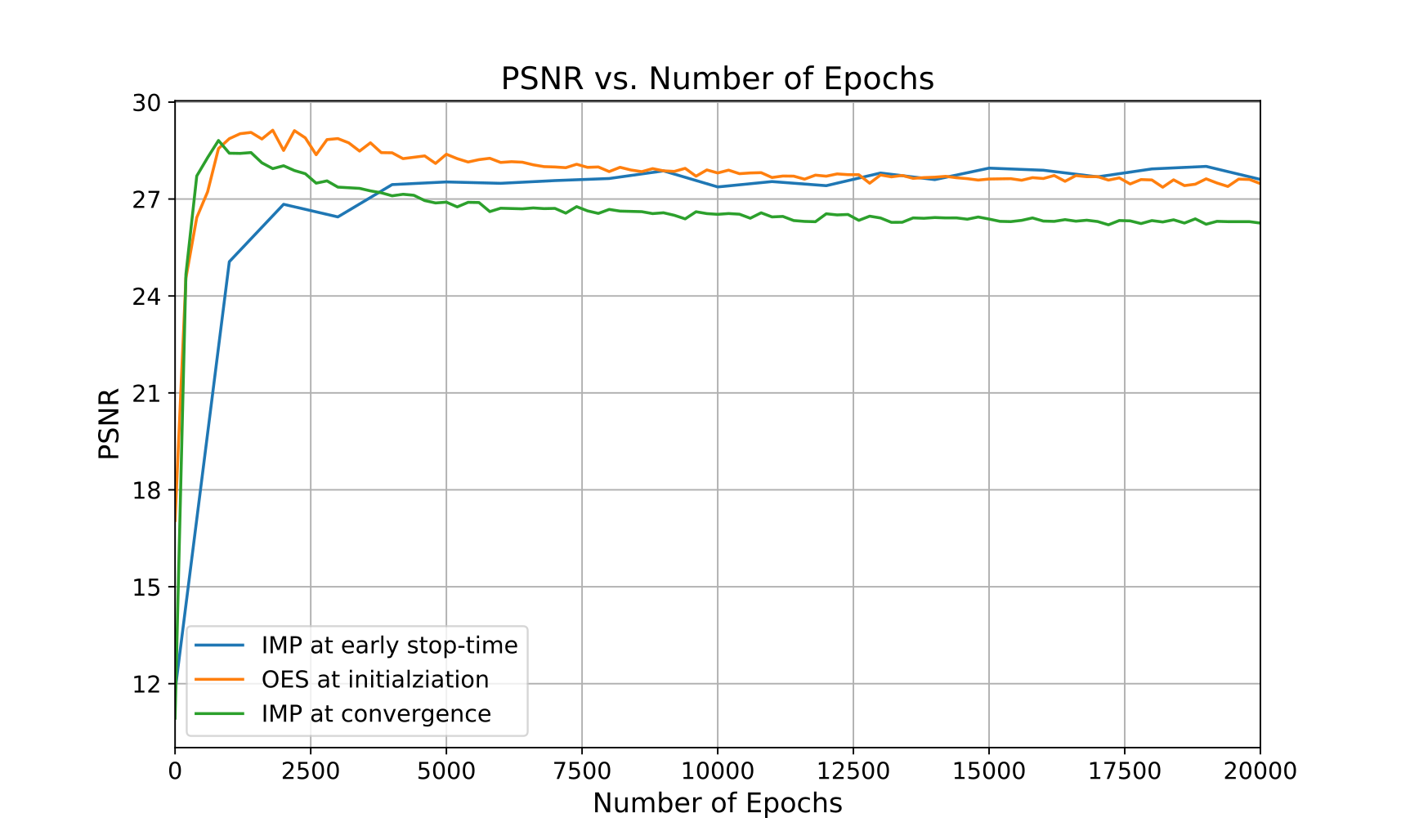}
        \caption{Comparison of IMP masks at early-stop time with IMP mask at convergence and OES at initialization on the same image (Pepper).}
        \label{fig:sub1}
    \end{subfigure}\hfill 
    \begin{subfigure}[b]{0.48\textwidth} 
        \centering
        \includegraphics[width=\textwidth]{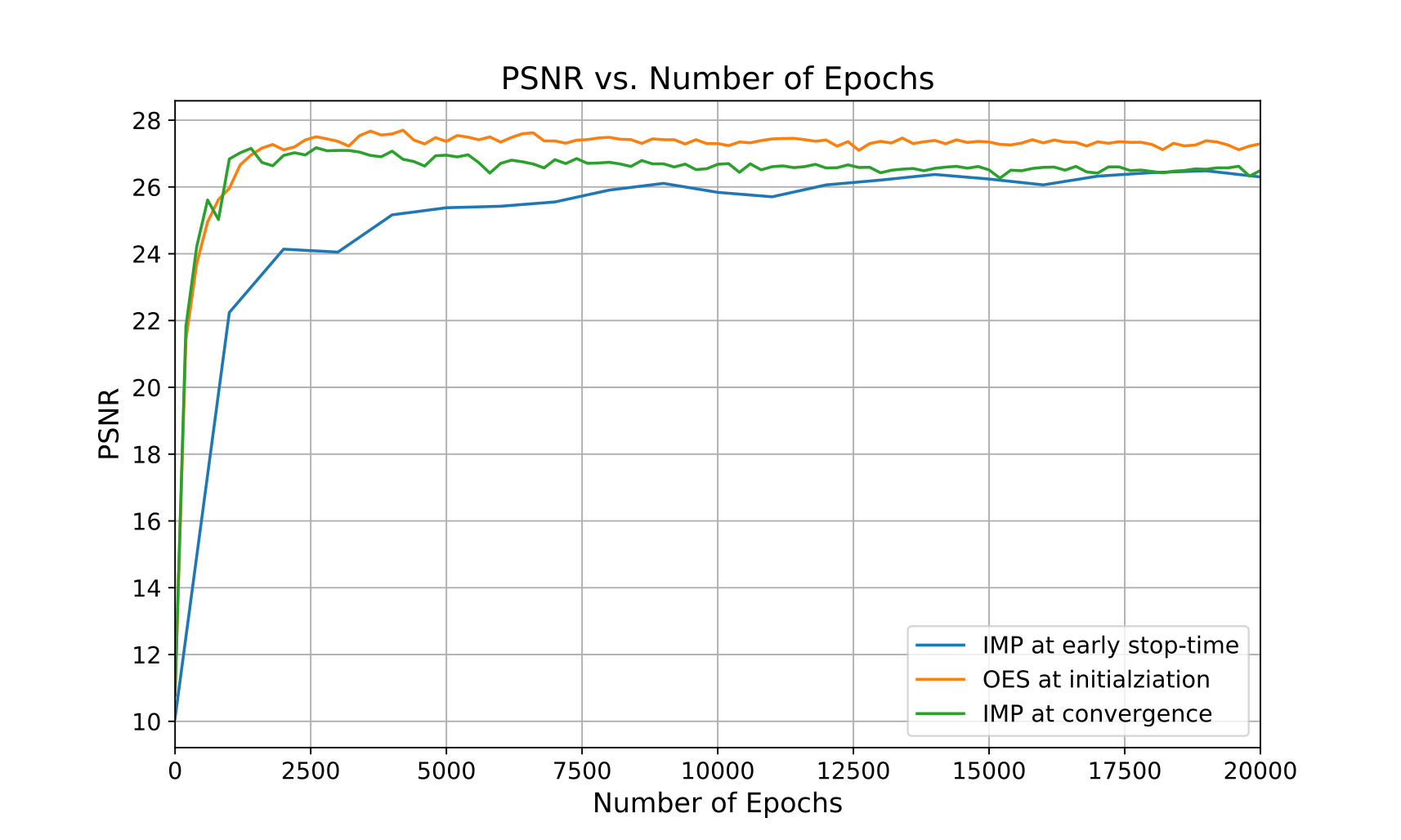}
        \caption{Comparison of IMP masks at early-stop time with IMP mask at convergence and OES at initialization on different images. Mask learned with Lena, used to denoise Flowers.}
        \label{fig:sub2}
    \end{subfigure}
    \caption{IMP masks learned at early-stopping time performs comparatively well. But when used on transfer tasks performs worse than OES masks at initialization. All the masks are $5\%$ sparse.}
    \label{fig:imp_early_oes}
\end{figure*}

To further make an apple-to-apple comparison with \citet{wu2023chasing}, we compare our method when the clean Lena and Pepper images were used to learn the mask. We observe that when IMP uses clean Lena and Pepper image for learning the mask, the denoising performance is improved as compared to when IMP only used the corrupted image. Like in Table-\ref{noise_lena}, the final PSNR achieved when IMP used the noisy image was 25.85dB (Lena-self in Table-\ref{noise_lena}) whereas when IMP used the clean image to learn the mask, the denoising performance improved to 26.65 dB (Lena-self in Table-\ref{clean_lena}). But the improvement, for denoising other images does not increase when compared to OES. For example, \textit{the PSNR of IMP masks using the clean image (26.65 dB (Lena-self in Table-\ref{clean_lena}) is still less than when OES used the corrupted image (27.05 dB in Table-\ref{noise_lena}).} When the target image was different, say for Barbara image, the mask learned on clean Lena image using IMP gives a PSNR of 25.66 dB (Table-\ref{clean_lena}) but using OES mask with a corrupted image gives PSNR of 26.34 dB (Table-\ref{noise_lena}, Barbara). We further explore this phenomenon of transferability in Figure-\ref{fig:imp_early_oes} where IMP masks learned on clean image performed well, when it was used for denoising on the same image but performed worse than OES when it was used for a different image. 
 
\begin{table}[ht]
\centering
\small 
\caption{Comparison of Denoising Capabilities of Transferred Masks Obtained from Sparse-DIP Pruning at Initialization vs IMP/OES Based Pruning. Here, both the OES and IMP masks were learned on \textit{clean Lena image.}}
\begin{minipage}{.45\linewidth}
\centering
\subcaption{Masks learned on \textbf{clean Pepper image}.}
\begin{tabular}{|l||c|c|}
\hline
\multicolumn{1}{|c||}{Image} & \multicolumn{2}{c|}{$\sigma = 25dB$} \\ \hline
& $m(IMP)$ & $m(OES)$ \\ \hline
Pepper (self) & 26.89 & \textbf{27.68} \\ \hline
Flowers & 26.48 & \textbf{26.80} \\ \hline
Lena  & 25.96 & \textbf{26.38} \\ \hline
Barbara & 25.42 & \textbf{26.32} \\ \hline
Monarch & 26.73 & \textbf{27.40} \\ \hline
Baboon & 23.49 & \textbf{23.89} \\ \hline
Ppt3 & 26.36 & \textbf{26.84} \\ \hline
Bridge & 25.75 & \textbf{26.03} \\ \hline
Zebra & 26.58 & \textbf{27.20} \\ \hline
Man & 26.19 & \textbf{26.94} \\ \hline
\end{tabular}
\end{minipage}%
\quad
\begin{minipage}{.45\linewidth}
\centering
\subcaption{Masks learned on \textbf{clean Lena image}.}
\begin{tabular}{|l||c|c|}
\hline
\multicolumn{1}{|c||}{Image} & \multicolumn{2}{c|}{$\sigma = 25dB$} \\ \hline
& $m(IMP)$ & $m(OES)$ \\ \hline
Foreman & 26.39 & \textbf{26.67} \\ \hline
Lena(self) & 26.65 & \textbf{26.83} \\ \hline
Barbara  & 25.66 & \textbf{26.46} \\ \hline
Monarch & 26.61 & \textbf{27.35} \\ \hline
Baboon & 23.65 & \textbf{24.02} \\ \hline
Ppt3 & 23.19 & \textbf{26.85} \\ \hline
Bridge & 25.81 & \textbf{26.21} \\ \hline
Man & 26.31 & \textbf{26.88} \\ \hline
\end{tabular}
\label{clean_lena}
\end{minipage}
\end{table}

We observe a similar phenomenon in transferability in Figure-\ref{fig:imp_early_oes} when IMP masks were obtained at an early stopping time. 

\section{Transfer to Different Task: Inpainting}
\label{inpaint_task}
In the manuscript, we performed on learning mask $\m(\y)$ from noisy images $\y$, where $\y$ is corrupted by additive Gaussian noise with standard deviation $\sigma$. In this section, we show the efficiency of OES masks, when $\y$ is a masked image with probability of masking $p=0.5$, i.e, on average $50\%$ of the image pixels are missing. We compare the masks learned when $\y$ had missing pixels (referred to as Sparse-DIP-in) and when $\y$ was corrupted with Gaussian noise (referred to as Sparse-DIP-den). We evaluate the relative comparison of both these masks against deep decoder and dense DIP at convergence. We report the results of these masks in the inpainting task in Table-\ref{inpaint}. Furthermore, we use this set of masks for denoising in two different noise levels $\sigma=25dB$ and $\sigma=12dB$ and report in Table-\ref{inpaint_denoise}. Based on this observation in Table-\ref{inpaint}, we see that sparse-DIPs (mask learned from missing pixel $\y$ or noisy $\y$) seems to perform comparably with Deep-decoder and Vanilla DIPs. This is because for inpainting tasks, the effect of overfitting is not as pronounced as compared to denoising tasks. Both the masks learned from denoising task and the inpainting task seem to perform comparably in Table-\ref{inpaint_denoise}.

\begin{table}[ht]
\centering
\small 
\caption{Comparison of inpainting capabilities of transferred masks (denoising training) obtained from Sparse-DIP pruning at initialization vs IMP based pruning.  $p=0.5$  }
\begin{tabular}{|c|c|c|c|c|}
\hline
& Dense DIP & Deep Decoder & Sparse-DIP (den) & Sparse-DIP (in)\\ \hline
Ppt3 & \textbf{28.62} & 28.40 & 28.43 & 28.33\\ \hline
Baboon & 21.36 & 22.13 & \textbf{22.60} & 22.38\\ \hline
Coastguard & \textbf{27.80} & 27.27 & 27.45 & 27.45 \\ \hline
Man & 25.47 & \textbf{26.63} & 26.13 & 26.16 \\ \hline
Zebra & 31.20 & 29.52 & 31.62 & \textbf{31.27}\\ \hline
Pepper & 28.40 & 28.45 & 30.76 & \textbf{30.81}\\ \hline
Face & 28.64 & \textbf{31.27} & 31.12 & 29.10 \\ \hline
Comic & 22.36 & \textbf{24.53} & 22.55 & 22.54 \\ \hline
Flowers & 30.73 & 29.61 & \textbf{31.10} & 30.85\\ \hline
Bridge & 24.85 & \textbf{25.16} & 25.01 & 24.78 \\ \hline
Foreman & 31.57  & 33.60  & 30.75 & 31.53\\ \hline
Monarch & 30.54 & 31.08 & 31.44 & \textbf{31.70}\\ \hline
Barbara & \textbf{27.62} & 25.71 & 27.23 & 27.28\\ \hline
Lena & 28.85 & \textbf{31.23} & 31.17 & 31.17 \\ \hline

\end{tabular}
\label{inpaint}
\end{table}

\begin{table}[ht]
\centering
\caption{Denoising capabilities comparison without early stopping on Set-14 dataset. The Sparse-DIP masks have been generated with two procedures. "denoise" denotes the masks generated from the denoising operation. "inpaint" denotes the masks generated from the inpainting operation.}
\small 
\begin{tabular}{|c|c|c|c|c|c|c|c|c|c|c|c|c|}
\hline
\multicolumn{1}{|c|}{} & \multicolumn{4}{c||}{$\sigma = 25dB$} & \multicolumn{4}{c|}{$\sigma = 12dB$} \\ \hline
Image & Dense & Deep & \multicolumn{2}{c||}{Sparse-DIP} & Dense & Deep & \multicolumn{2}{c|}{Sparse-DIP} \\ 
 & DIP & Decoder & denoise & inpaint & DIP & Decoder & denoise & inpaint \\ \hline
Pepper & 21.21 & 27.08 & 27.45 & \textbf{27.46} & 27.34 & 28.81 & 29.89 & \textbf{30.40}   \\ \hline
Foreman & 20.69 & 24.20 & \textbf{25.15} & 24.36 & 26.59 & 29.81 & 30.12 & \textbf{30.37} \\ \hline
Flowers & 22.27 & 27.03 & \textbf{27.10} & 27.10 & 28.55 & 30.33 & \textbf{31.07} & 31.07 \\ \hline
Comic & 20.63 & 23.35 & \textbf{24.03} & 23.96 & 26.45 & 28.01 & 28.57 & \textbf{28.81} \\ \hline
Lena & 21.28 & \textbf{26.85} & 26.40 & 26.23 & 27.50 & \textbf{30.96} & 30.89 & 30.85 \\ \hline
Barbara & 23.90 & 25.30 & \textbf{26.50} & 26.25  & 28.27 & 27.50 & 29.85  &  \textbf{30.09} \\ \hline
Monarch & 23.62 & 27.87 & \textbf{27.87} & 27.82 & 28.25 & 32.00 & 32.12 & \textbf{32.67} \\ \hline
Baboon & 21.68 & 22.93 & 24.00 & \textbf{24.10} & 27.27 & 24.12 & \textbf{25.91} & 25.90\\ \hline
Ppt3 & 24.07 & 26.81 & \textbf{27.20} & 26.64 & 28.88 & 31.73 & 32.41 & \textbf{32.50}\\ \hline
Coastguard & 20.53 & 23.71 & \textbf{24.19} & 24.16 & 26.50 & 29.43 & \textbf{30.60} & 29.80 \\ \hline
Bridge & 21.77 & 25.19 & \textbf{26.12} & 26.10 & 28.58 & 28.10 & \textbf{29.23} & 28.98 \\ \hline
Zebra & 21.94 & \textbf{27.40} & 27.32 & 27.29 & 28.45 & 30.81 & 31.54 & \textbf{31.62} \\ \hline
Face & 21.03 & 24.14 & \textbf{24.18} & 24.05 & 26.90 & 29.93 & \textbf{29.93} & 29.86\\ \hline
Man & 21.98 & 26.32 & \textbf{26.59} & 26.55 & 28.45 & 29.84 & \textbf{30.94} & 30.56 \\ \hline
\end{tabular}
\label{inpaint_denoise}
\end{table}

However, for denoising task in Table-\ref{inpaint_denoise}, we see that Sparse-DIP (learned through denoising and inpainting loss) outperforms both Deep decoder and Dense-DIP due to severe overfitting. This is something we already explored in Table-\ref{main_table_denoise}.

\begin{figure}[h]
    \centering
    \includegraphics[width=0.6\textwidth]{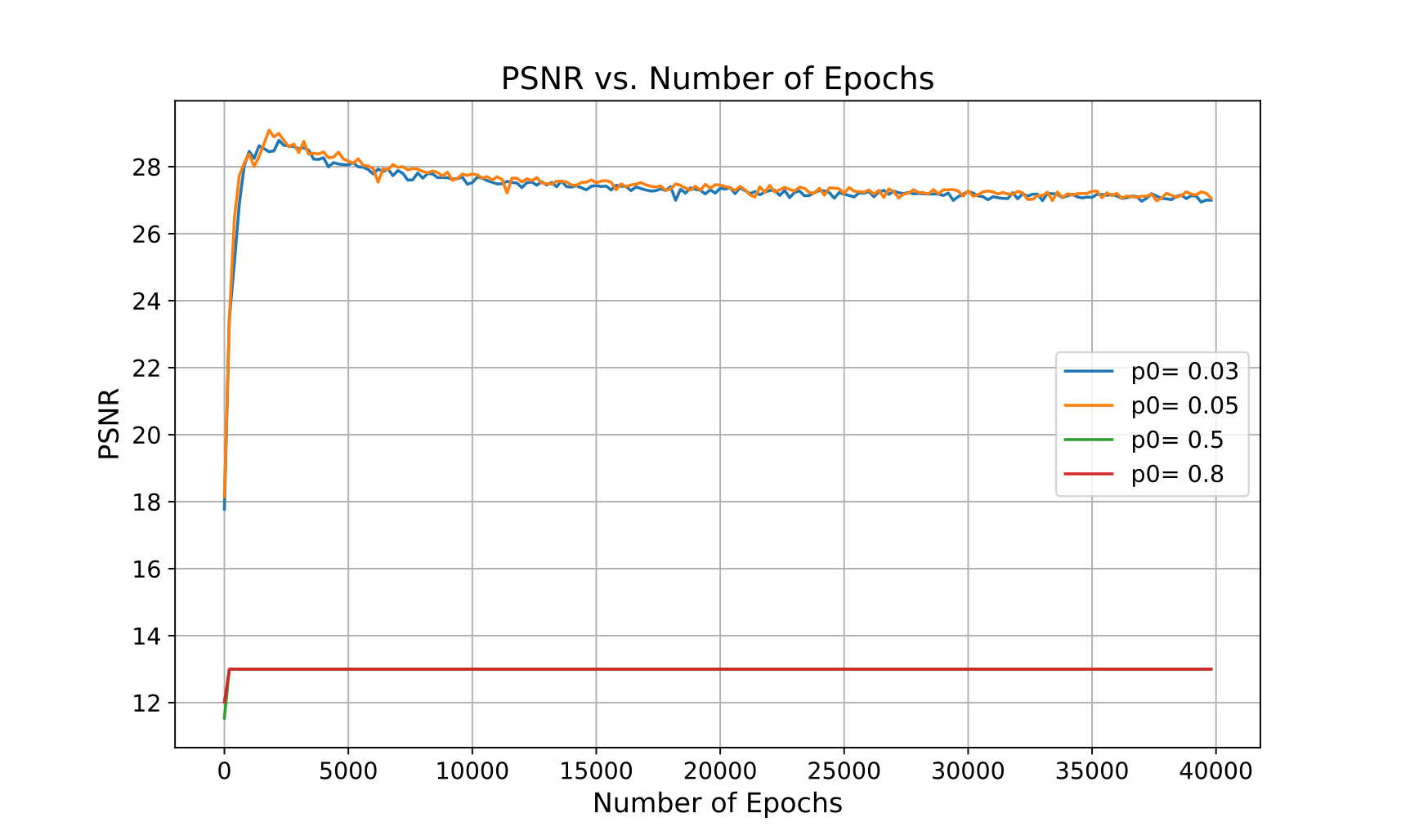}
    \caption{Performance of subnetworks trained with different prior $p_{0}$'s in equation and then pruned $95\%$ by ranking. This shows that the importance ranking of $p$'s after training is dependent on prior $p_{0}$. Good results are expected when prior $p_{0}$ used in optimization, matches the pruning percentage. }
    \label{fig:diff_p0}
\end{figure}

\begin{figure}[h]
    \centering
    \includegraphics[width=0.8\textwidth]{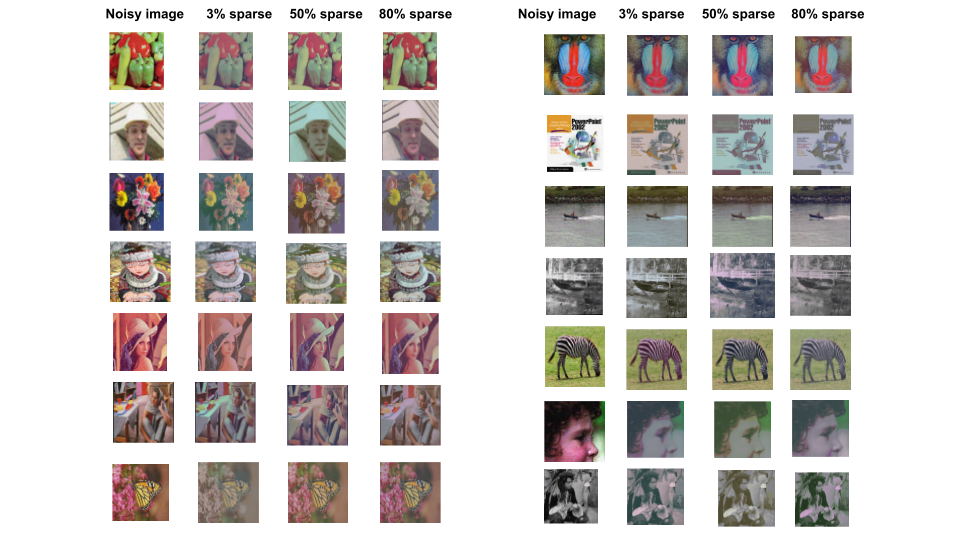}
     \caption{ \centering $G(\p_{in} \circ \m^{*}(\y),\z)$ for Set-14 dataset. 
    Images generated by the randomly initialized network found after applying OES mask.}
    \label{fig:set-14_masked}
\end{figure}

\begin{figure*}[ht]
  \centering
  \begin{minipage}[b]{0.49\textwidth}
    \centering
    \includegraphics[width=\linewidth]{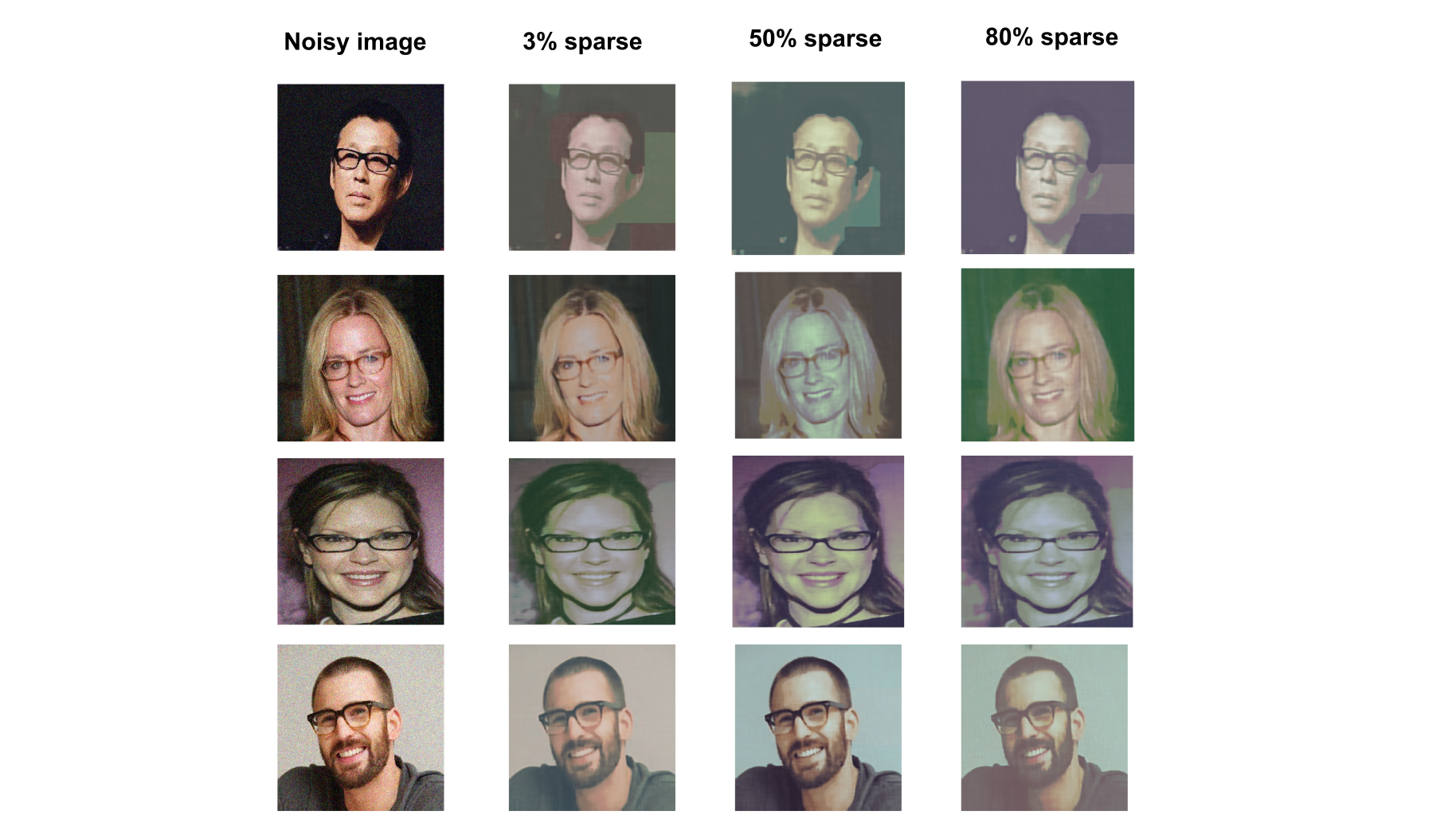}
    \subcaption{ \centering $G(\p_{in} \circ \m^{*}(\y),\z)$ for Face dataset.} 
    \label{fig:left-image}
  \end{minipage}
  \hfill
  \begin{minipage}[b]{0.49\textwidth}
    \centering
    \includegraphics[width=\linewidth]{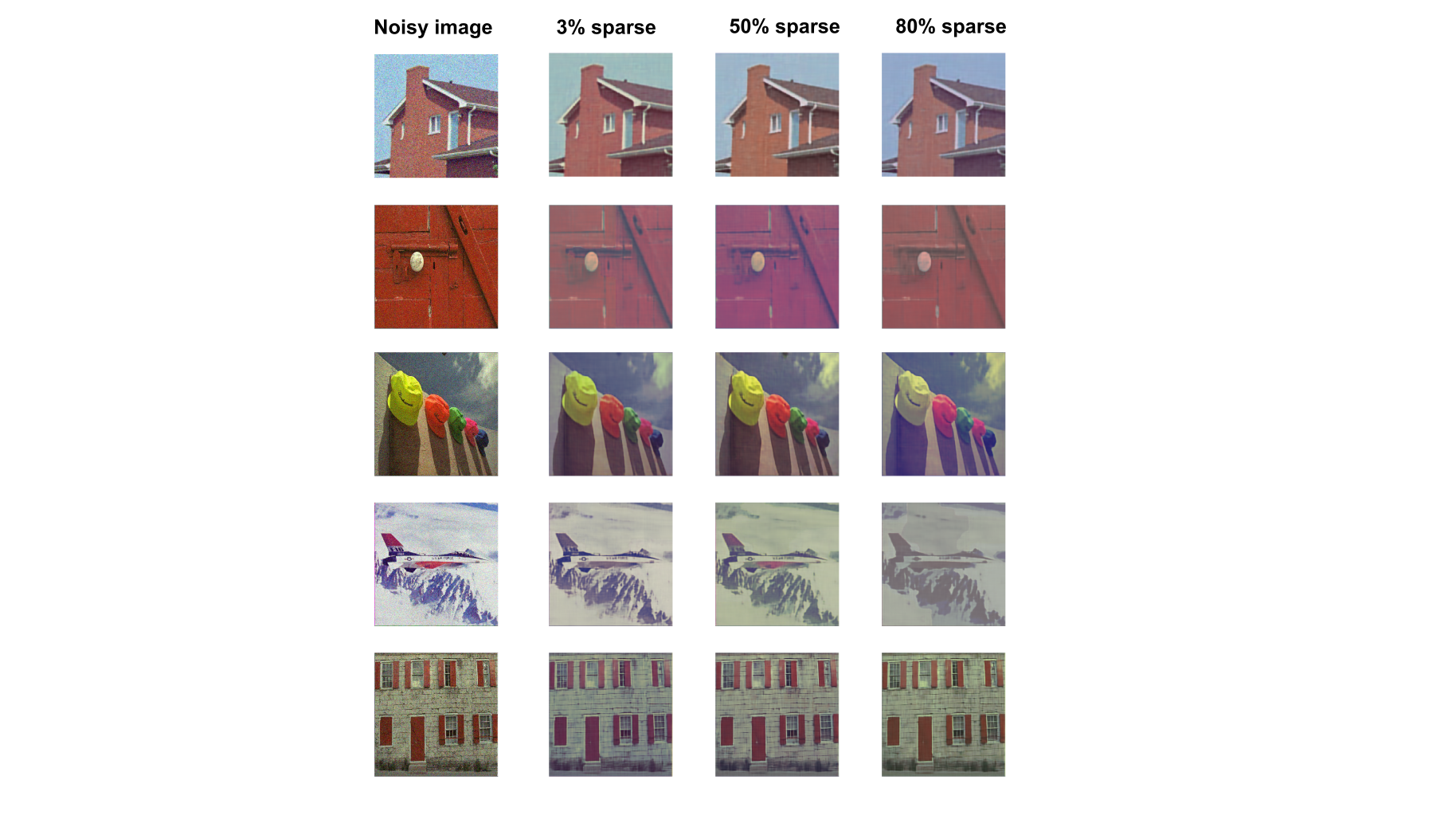}
    \subcaption{\centering $G(\p_{in} \circ \m^{*}(\y),\z)$ for standard dataset.} 
    \label{fig:right-image}
  \end{minipage}
   \caption{ \centering Masking at initialization can induce image prior. 
   Figures shows the images after masking image generator at initialization $G(\p_{in} \circ \m^{*}(\y),\z)$. 
  The mask $\m^{*}$ was learned using OES algorithm. Images corresponding to several sparsity levels are shown. } 
  \label{fig:both-images}
\end{figure*}


\begin{figure}[h]
    \centering
    \includegraphics[width=0.9\textwidth]{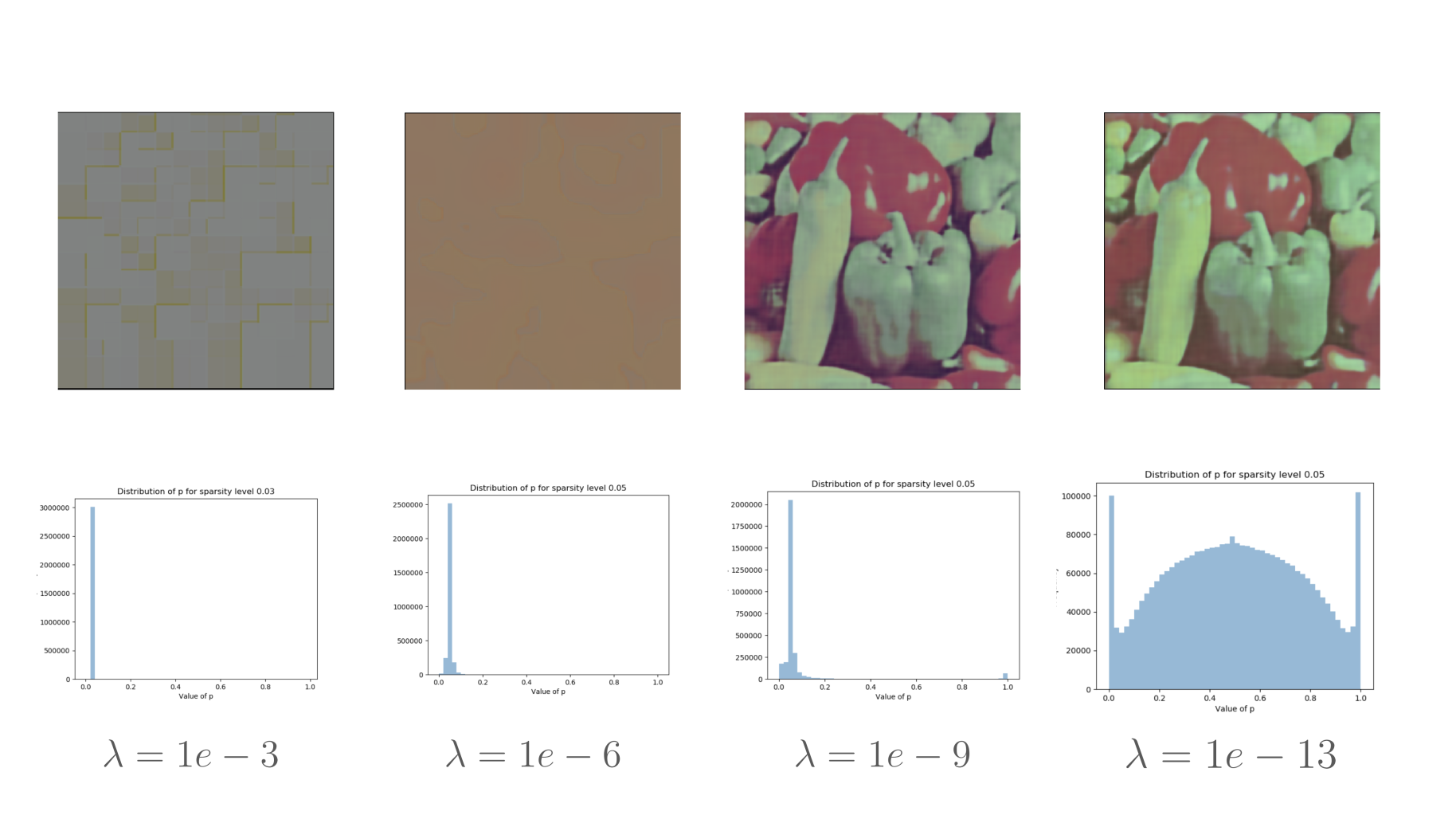}
    \caption{Distribution of logits ($\pr$) for various $\lambda$ in front of KL term and its effect of the output image ($G(\p_{in} \circ \m^{*}(\y),\z)$) after thresholding the logits $\pr$ to reach the desired sparsity level. Here the prior $\pr_{0}=0.05 \times \mathbf{1}$ and the desired threshold level is also $5\%$ sparsity. Different strength of KL term $\lambda$ leads to the distribution of logits $\pr$ centered around the desired prior $\pr_{0}$. From eq-, we observe that higher $\lambda =1e-3$ or  $\lambda =1e-6$ gives more importance to the KL term and less importance to the image data-fidelity term (no image formation). $\lambda =1e-9$ gives the best balance of regularization and data-fidelity. For $\lambda =1e-9$, although the centre of distribution is at $\pr_{0}$, there is some concentration near $\pr=1$, ensuring that there is a clear distinction between the important and non-important parameters. OES subnetwork is $5\%$.  }
    \label{fig:kl}
\end{figure}

\begin{figure*}[h]
    \centering
    \includegraphics[width=0.8\textwidth]{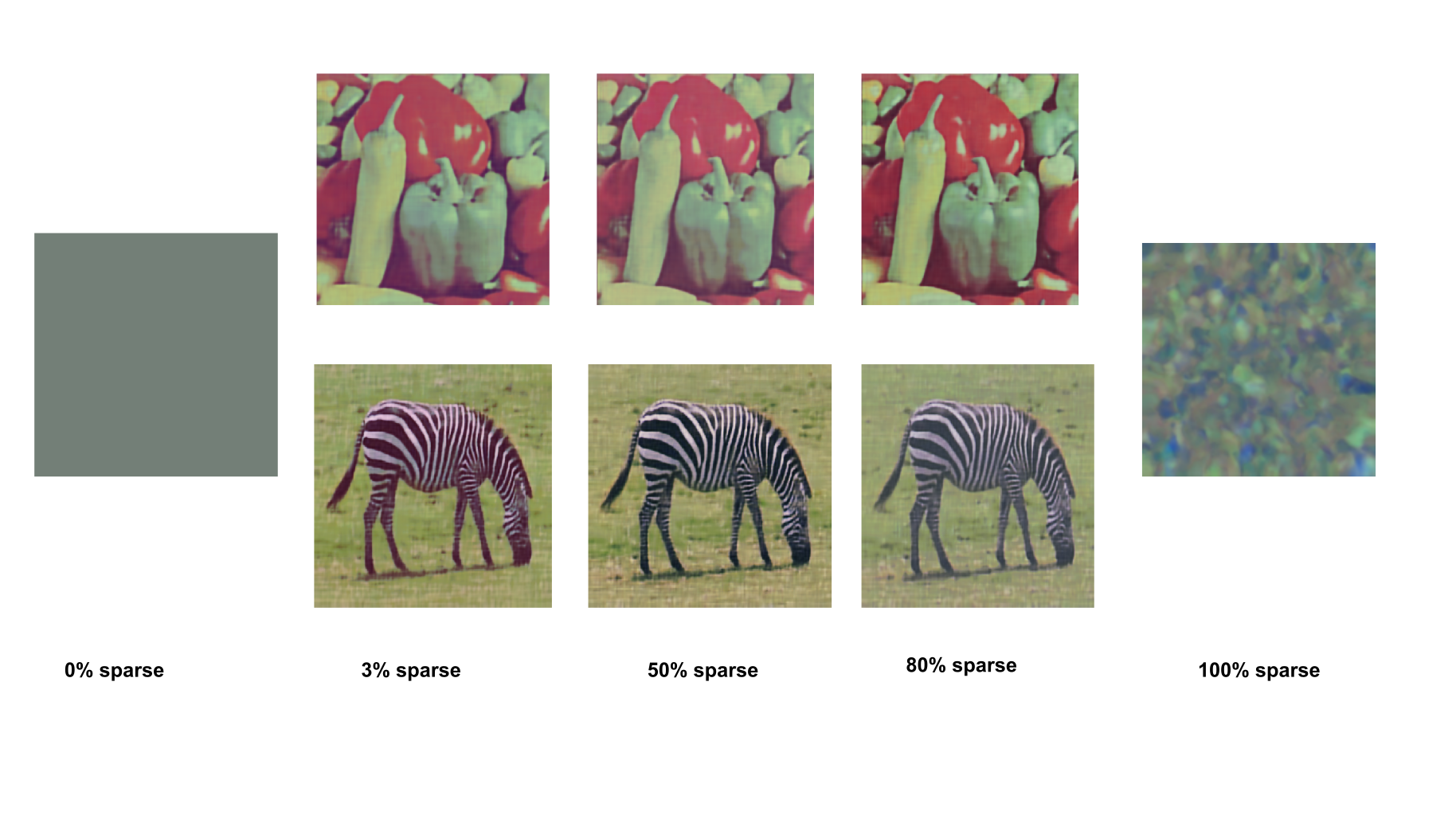}
    \caption{ $G(\p_{in} \circ \m^{*}(\y),\z)$: capability of image representation by just masking network parameters. When $\m = \mathbf{1}$, images correspond to stochastic processes producing spatial structures with self-similarity as noticed in \citet{ulyanov2018deep}. For $\m = \mathbf{0}$, it produces a constant image (assuming no bias terms). However, in the middle ground, different images (even at a fixed sparsity level) can be represented by the combination of chosing to select a weight parameter or delete it. }
    \label{fig:mask_rep}
\end{figure*}

\begin{figure}[h]
    \centering
    \includegraphics[width=1\textwidth]{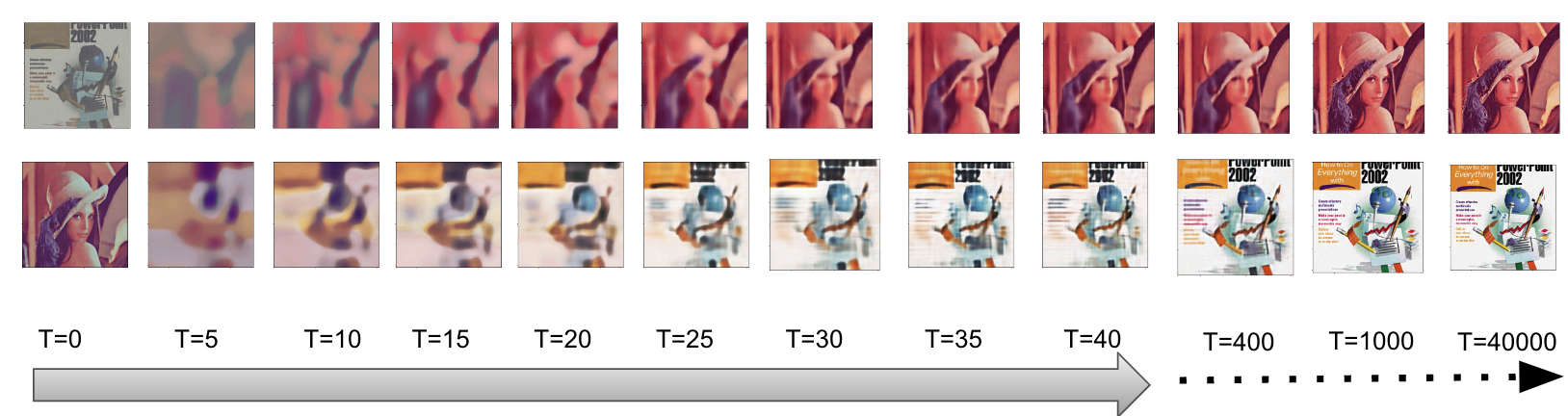}
     \caption{Transferability of OES subnetworks. OES masks trained on $\y_{1}$, denoted as  $G(\p_{in} \circ \m^{*}(\y_{1}),\z)$ can be used for denoising image $\y_{2}$. Here interchanging $\y_{1}$ and $\y_{2}$ in the opposite way also ensures the operation of OES. At epoch $T=0$, just the application of mask on random network initialization (on which mask was learned), produces an image. Epoch $T=40000$ denotes the final recovered image that does not suffer from overfitting. Underparameterization by OES subnetwork ensures that the output lies in the manifold of natural image prior. }
    \label{fig:transfer_flow}
\end{figure}

\begin{figure}[h]
    \centering
    \includegraphics[width=0.4\textwidth]{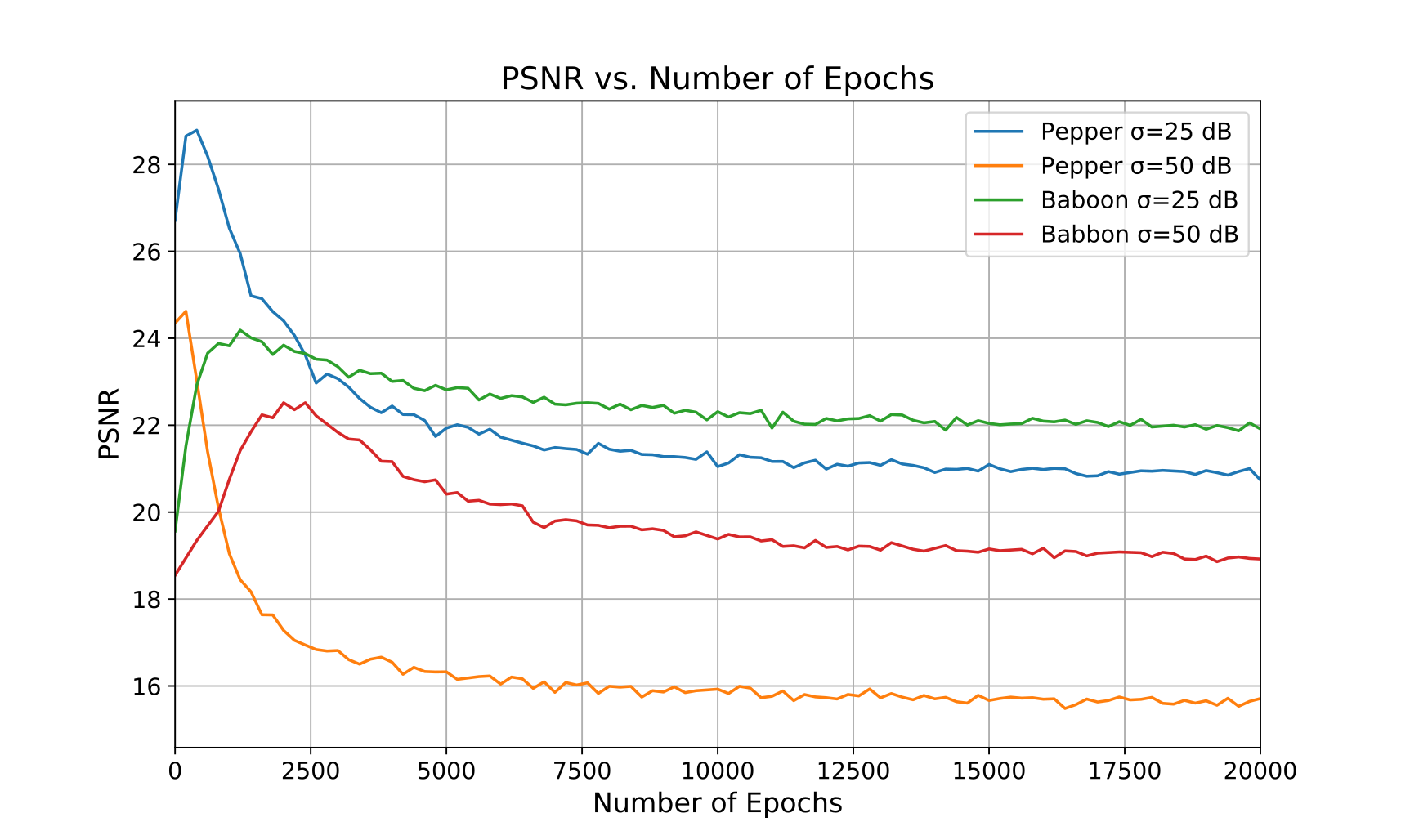}
    \caption{Early stopping time window can vary for different images and also various noise levels. Estimating this early-stopping time from an image distribution or a particular noise level can be difficult. Here we see that there can be a window as large as 2500 iterations between early stopping times of two images with different corruption levels. }
    \label{fig:early_var}
\end{figure}

\section{Sensitivity of $\lambda$ in OES Mask Learning and Selectivity of $\pr_{0}$}
\label{sense_lam_oes}
\subsection{Sensitivity of $\lambda$}

 We found empirically that OES algorithm is robust to the choice of $\lambda$, given a network architecture with fixed number of parameters (Unet in this case). We fix $\lambda=1e-9$ for all our experiments. For this particular experiment, we take $\pr_{0}=0.05 \times \mathbf{1}$ and threshold $95\%$ of the weights by ranking $\pr$. The initialization value was taken to be at  $\pr=0.5 \times \mathbf{1}$, where all the weights have equal chance of selection or deletion. $\lambda$ controls the regularization balance on fitting the image (first part of the loss) or by making the distribution $Ber(\pr)$ close to $Ber(\pr_{0})$ (second part of the loss). Note that $\pr_{0}$ is the pre-specified prior probability that is same for all the parameters of the network. As $\lambda \rightarrow \infty$, then $\pr \rightarrow \pr_{0}$, at this limit a) there is no image generation at initialization, as the first part of the loss is not minimized and b) there is no separation among the converged values $\pr$ and the probabilities for all the elements will collapse to $\pr_{0}$. So the mask can't be formed by ranking and thresholding at the desired sparsity level. 

\begin{align*}
\m^{*} &= C(\pr^{*})  \quad \text{such that} \quad \\
& \pr^{*} = \arg \min_{\pr} \underbrace{\mathbb{E}_{\m \sim Ber(\pr) }  \left[ || G(\p_{in} \circ \m,\z) - \y||_{2}^2 \right]}_{R(Q)}  + \lambda KL(Ber(\pr)||Ber(\pr_{0}(s))) \label{ber}
\end{align*}

$\lambda =1e-3$ correspond to this observation in Figure-\ref{fig:kl}. Increasing it to $\lambda=1e-6$, we observe that $\pr$'s for different weights start to vary and are not entirely localized at $\pr_{0}$. However, even in this case, ranking the values of $\pr$, leads to layer collapse. Layer collapse happens in this phenomenon because important weights are thresholded. For smaller $\lambda=1e-13$, we observe that the distribution $\pr$, is uniform around the initialization $\pr=0.5 \times \mathbf{1}$. Although the image is formed by masking in this case, the distribution remains uniform. We see that at $\lambda=1e-9$, the distribution of $\pr$ seems to have two modes. We see a clear distinction where some of the $\pr$'s are localized at 1 and other is centered around $\pr_{0}$. This leads to better separation while thresholding and pruning the weights based on $\pr$. However, we note that the value of the KL would depend on the size of the network, for the current Unet architecture we are using, which has 3 million parameters, we found that $1e-9$ works the best among all the other values in logarithmic scale.

\section{OES pruning for MRI reconstruction}
\label{mri_recon}
We extend the OES pruning and sub-network training framework to the setting of multi-coil magnetic resonance image (MRI) reconstruction from undersampled k-space measurements. In previous literature [4], dense networks based DIP was used for MRI reconstruction as folows:

\begin{equation}
\hat{\p} = \arg \min_{\p} \sum_{c=1}^{N_{c}} \| \A^{(c)} G(\p,\z) - \y^{(c)} \|_{2}^2 \tag{P1:Vanilla DIP}
\end{equation}

For multi-coil MRI, let there be $N_{c}$ number of coil sensitivity maps denoted as $\Se_{c} \in \mathbb{C}^{q \times q}$,  $c =1,2..,N_{c}$. The corresponding   
$\A^{c}$ denotes the undersampled forward linear operator $\A^{c}(\M) = \M \F \Se_{c}$. \(\left\{\mathbf{M} \in \{0,1\}^{q \times q}\right\}\)
 is the sampling mask in k-space, $\F \in \mathbb{C}^{q \times q}$ denotes the Fourier Transform operator and $\y^{(c)} \in \mathbb{C}^{q}$ denotes the undersampled k-space measurements. $G(\p,\z) $ is an overparameterized Unet with two channels that processes the real and complex channel separately and with trainable parameters $\p$ and fixed input $\z$. For our experiments, we use multi-coil fastMRI knee and brain datasets [1,2] which are available publicly. The coil sensitivity maps were obtained using the BART toolbox [3].
When the dense network [4] is trained with generic optimizer like ADAM, the above suffers from overfitting (Figure-\ref{fig:self-1}). In the OES framework, we first learn the mask for the subnetwork, denoted as $\m^{*}(\A,\y)$ (not to be confused with the k-space mask $\M$), where $\A(\M) =[\A^{c}(\M)]_{c=1}^{Nc}$ and $\y =[\y^{c}]_{c=1}^{Nc}$. For the sake of notation, we will omit the coil dependency $c$ as the loss can be combined across coils and written in terms of one forward operator $\A$ and measurements $\y$. 

\begin{equation}
\begin{aligned}
\m^{*}(\y,\A) &= C(\pr^{*})  \quad \text{such that} \quad \\
& \pr^{*} = \arg \min_{\pr} \mathbb{E}_{\m \sim Ber(\pr) }  \left[   \| \A G(\p_{in} \circ \m,\z) - \y \|_{2}^2  \right] \\
& + \lambda KL(Ber(\pr)||Ber(\pr_{0})).
\end{aligned}
\label{ber_mask}
\end{equation}


In Figure-\ref{fig:dataset}, we show the 4 MRI scans that are used in the following experiment. $\x$ denotes the ground truth MRI image (obtained from a full set of k-space measurements), $\M_{4\times}$ and $\M_{8\times}$ denote the $4\times$ and $8\times$ undersampling masks for k-space or Fourier space (white lines are sampled), respectively. $\A^{H}(\M_{4\times})\y$ and $\A^{H}(\M_{8\times})\y$ denote the conventional zero-filling MRI reconstructions that produce aliasing artifacts. 
We will denote the set of the forward operator and measurement pair as $(\A_{i}(\M_{4\times}), \y_{i})$ for data index $i= 1,2,3,4$ for $4\times$ undersampling rate. For $8\times$ undersampling rate, we denote the pair as $(\A_{i}(\M_{8\times}), \y_{i})$. 
In our experiments, we train the OES mask using the pair $(\A_{1}(\M_{4\times}), \y_{1})$, and then use the mask subnetwork to reconstruct MRI in four different scenarios across various network sparsity levels:

\begin{enumerate}
    \item \textbf{Self + same undersampling}: The target reconstruction pair is $(\A_{1}(\M_{4\times}), \y_{1})$. We denote this experiment as $P(\A_{1}(\M_{4\times}), \y_{1})$.
    \item \textbf{Self+higher undersampling}: The target reconstruction pair is $(\A_{1}(\M_{8\times}), \y_{1})$. We denote this experiment as $P(\A_{1}(\M_{8\times}), \y_{1})$
    \item \textbf{Cross + same undersampling}: The target reconstruction pair is $(\A_{i}(\M_{4\times}), \y_{i})$ for $i=2,3$ and $4$.  We denote this experiment as $P(\A_{i}(\M_{4\times}), \y_{i})$
    \item \textbf{Cross + higher undersampling}: The target reconstruction pair is $(\A_{i}(\M_{8\times}), \y_{i})$ for $i=2,3$ and $4$.  We denote this experiment as $P(\A_{i}(\M_{8\times}), \y_{i})$.
\end{enumerate}

Note that transfer to a higher undersampling rate demonstrates the capability of transferring to a different level of degradation. 
Once the mask $\m^{*}(\A_{1},\y_{1})$ is obtained, the subnetwork at initialization is further trained to convergence with the following optimization. Similar notations extend to $8\times$ undersampling rate. 

\begin{equation}
    \min_{\p} \| \A_{i}(\M_{4\times}) G(\p \circ \m^{*}(\y_{1},\A_{1}),\z) - \y_{i} \|_{2}^{2}    \tag{P$(\A_{i}(\M_{4\times}),\y_{i})$: Sparse-DIP}
\end{equation}
We make the following observations from the PSNR curves in Figure-\ref{mri-all-figs}.
\begin{itemize}
    \item  \textit{Sparse-DIP reduces overfitting:} Vanilla Dense DIP produces artifact-affected images in all the cases. This is due to the nullspace of the forward operator that does not offer any control over nonsampled frequencies.
    Sparse DIP has very less overfitting. 
    \item \textit{Sparse-DIP is robust to higher undersampling rate:} For higher undersampling factor, i.e, $8\times$ undersampling, vanilla dense DIP overfits much more. Sparse DIP at higher sparsities (above $90\%$) seems to be robust to overfitting even at $8\times$ undersampling. 
    \item \textit{Moderate overfitting at moderate sparsity}: With moderate sparsity level ($50\%,80\%$), subnetwork overfits artifacts when cross transfer tasks take place (different image's measurements) or when the undersampling rate is $8\times$. However, overfitting (at moderate sparsity levels) takes place to much less extent when self transfer takes place with the same undersampling rate $4 \times$. 
    \item \textit{Limited representation capability at very high sparsity:} For higher sparsity levels ($90\%$ or higher), overfitting rarely happens in any of the scenarios (cross-transfer or higher undersampling rate). For very high sparsity level $97\%$, the PSNR curve fails to rise very high, denoting that the network has already reached its representation capability. 
    
\end{itemize}

\begin{figure}[h]
    \centering
    \includegraphics[width=0.8\textwidth]{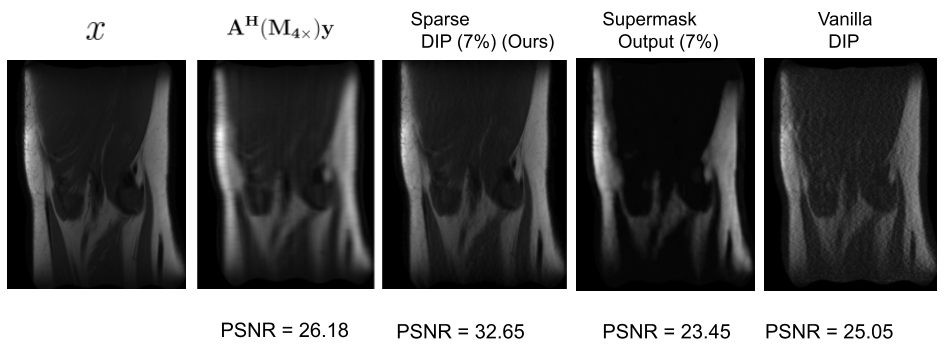}
    \caption{MRI reconstruction comparison with Sparse-DIP and Vanilla Dense DIP without early stopping. Sparse-DIP removes aliasing artifacts and preserves the important details of the images when compared to the ground-truth $\x$. Vanilla dense DIP overfits to artifacts (due to nullspace) and requires careful early stopping (See Figure-\ref{fig:self-1}). Supermasked output at network initialization still manages to capture some important image details. }
    \label{fig:main_result}
\end{figure}
 
\begin{figure}[h]
    \centering
    \includegraphics[width=\textwidth]{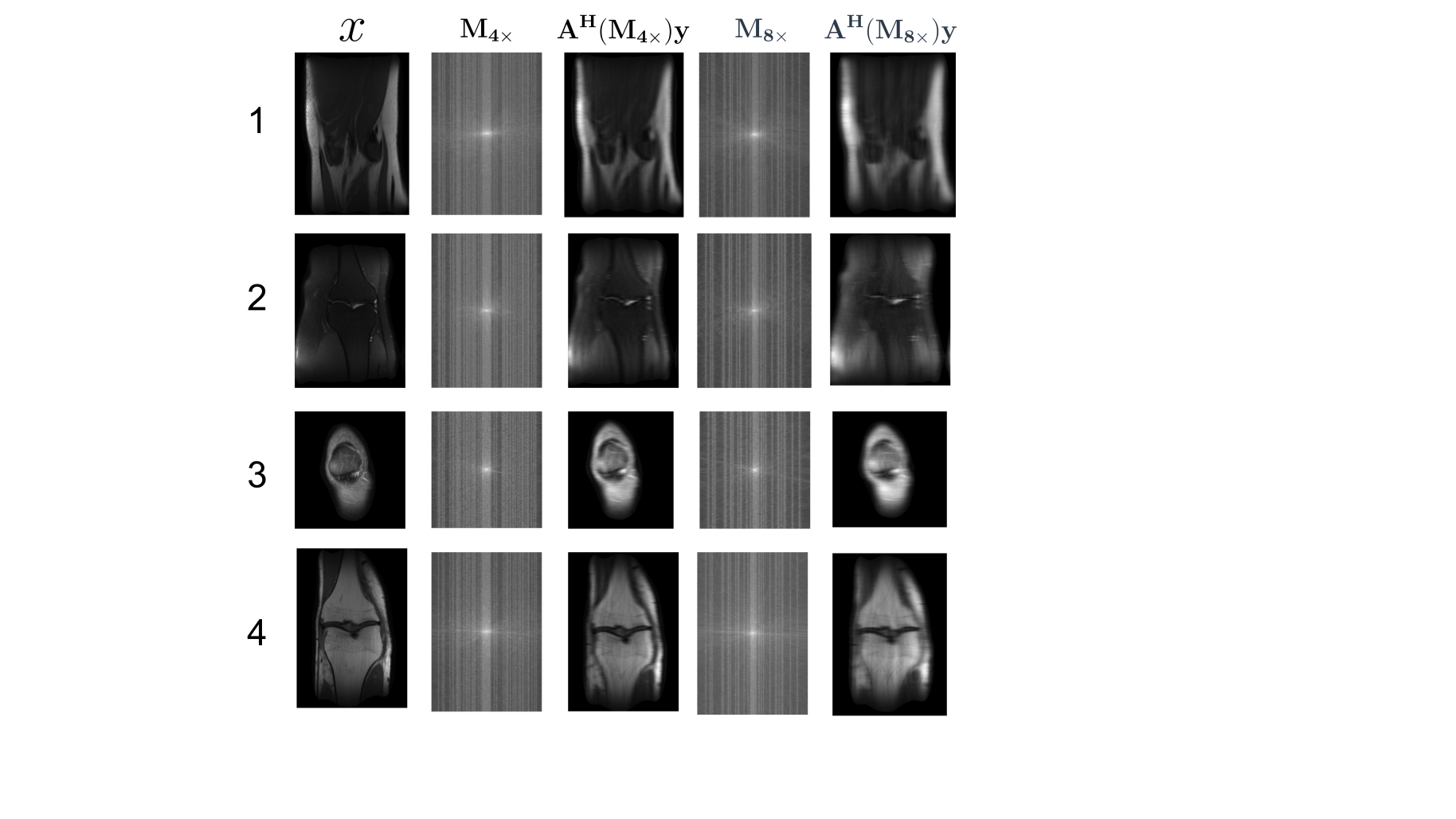}
    \caption{The 4 MRI ground-truth and measurements used in this experiment. $\x$ denotes the ground-truth image or full-kspace reconstruction. $\M_{4\times}$ and $\M_{8\times}$ denote the k-space undersampling masks. $\A^{H}(\M_{4\times})\y$ and $\A^{H}(\M_{8\times})\y$ denote the zero-filling reconstructions that produce aliasing artifacts. }
    \label{fig:dataset}
\end{figure}

\begin{figure*}[!h]

    \begin{minipage}{0.45\textwidth}
        \centering
        \includegraphics[width=\textwidth]{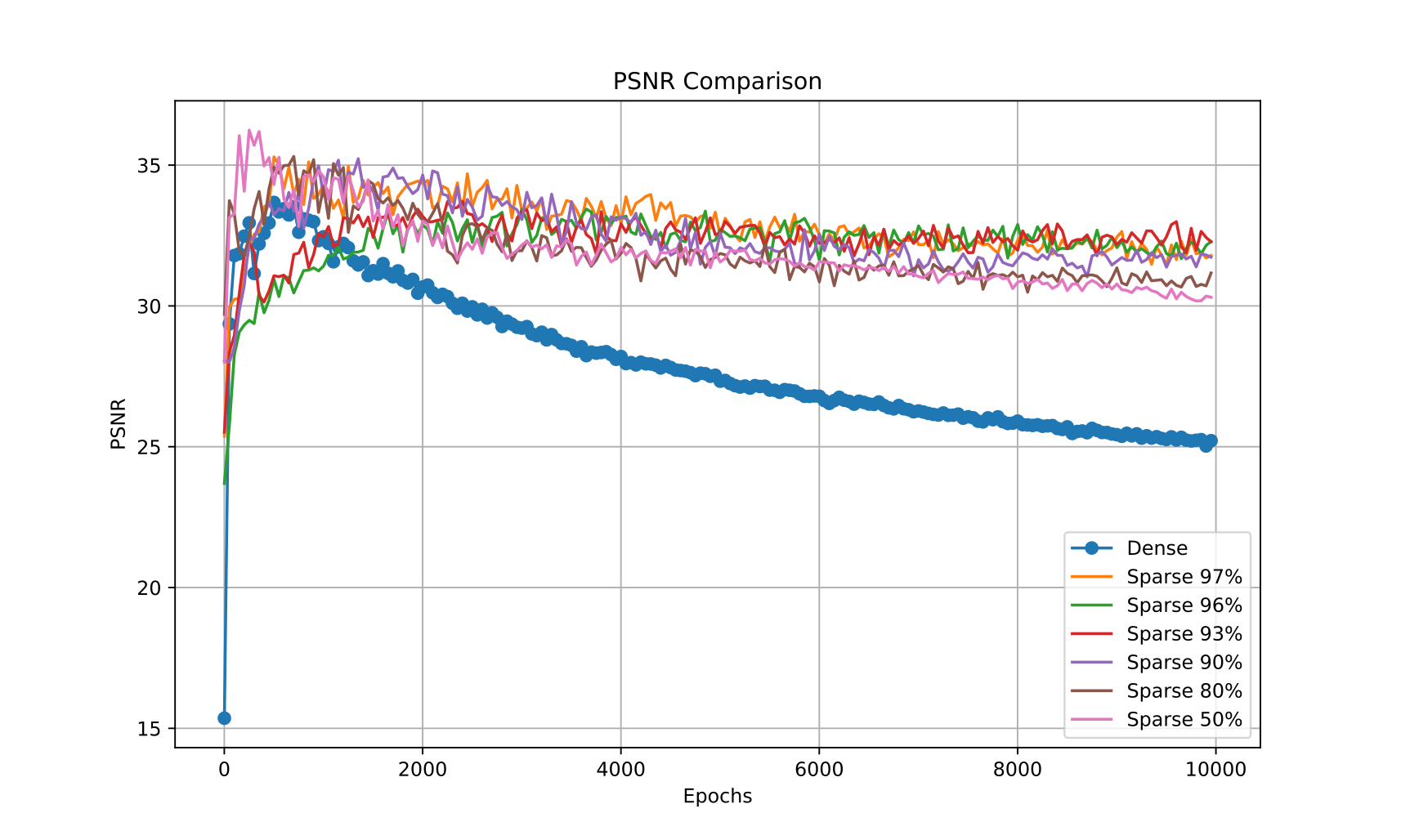}
        \subcaption{Self + same undersampling: $P(\A_{1}(M_{4 \times}),\y_{1})$}
        \label{fig:self-1}
    \end{minipage}\quad
    \begin{minipage}{0.45\textwidth}
        \centering
        \includegraphics[width=\textwidth]{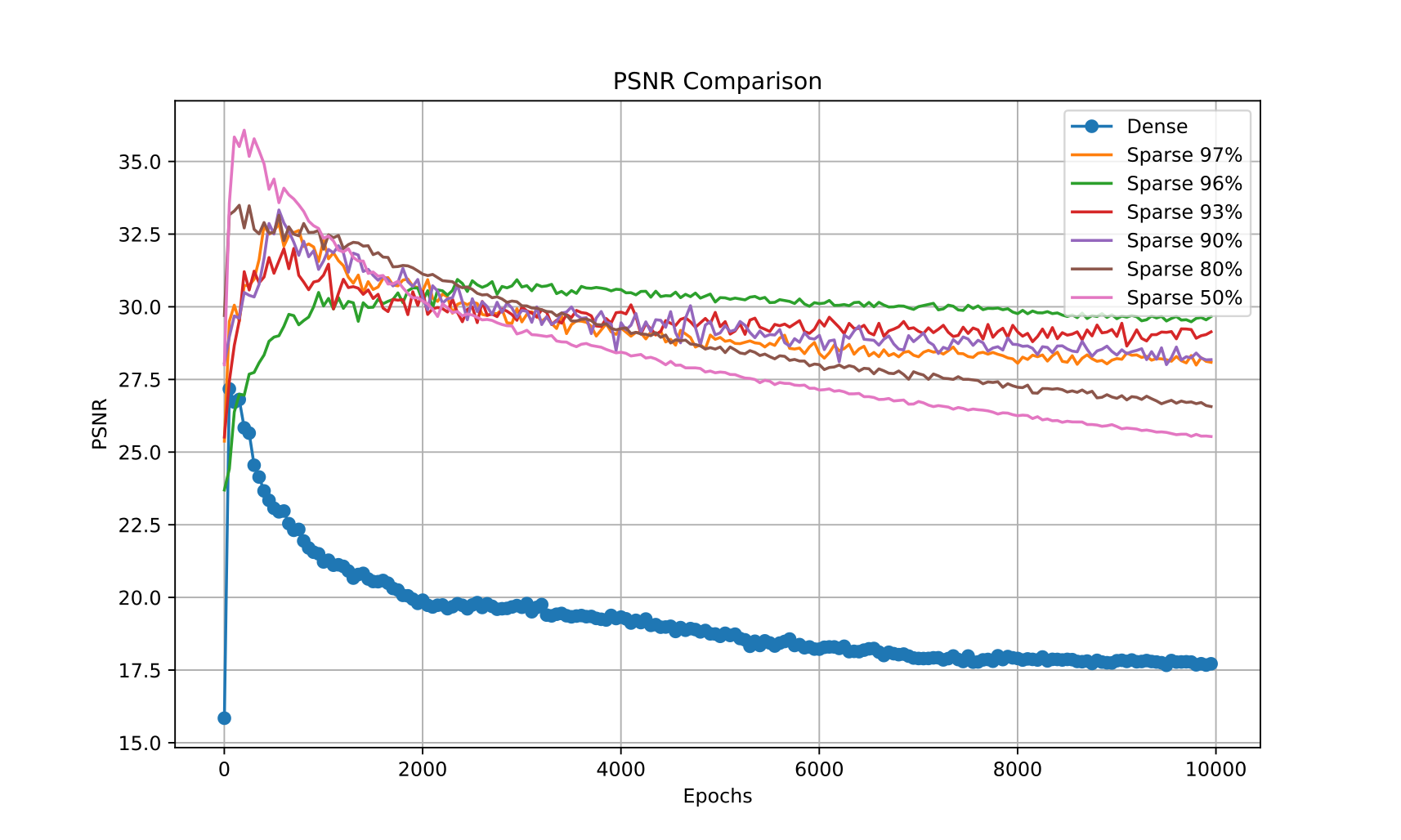}
        \subcaption{Self + higher undersampling: $P(\A_{1}(M_{8 \times}),\y_{1})$}
        \label{fig:self-cross-1}
    \end{minipage}\quad
        \begin{minipage}{0.45\textwidth}
        \centering
        \includegraphics[width=\textwidth]{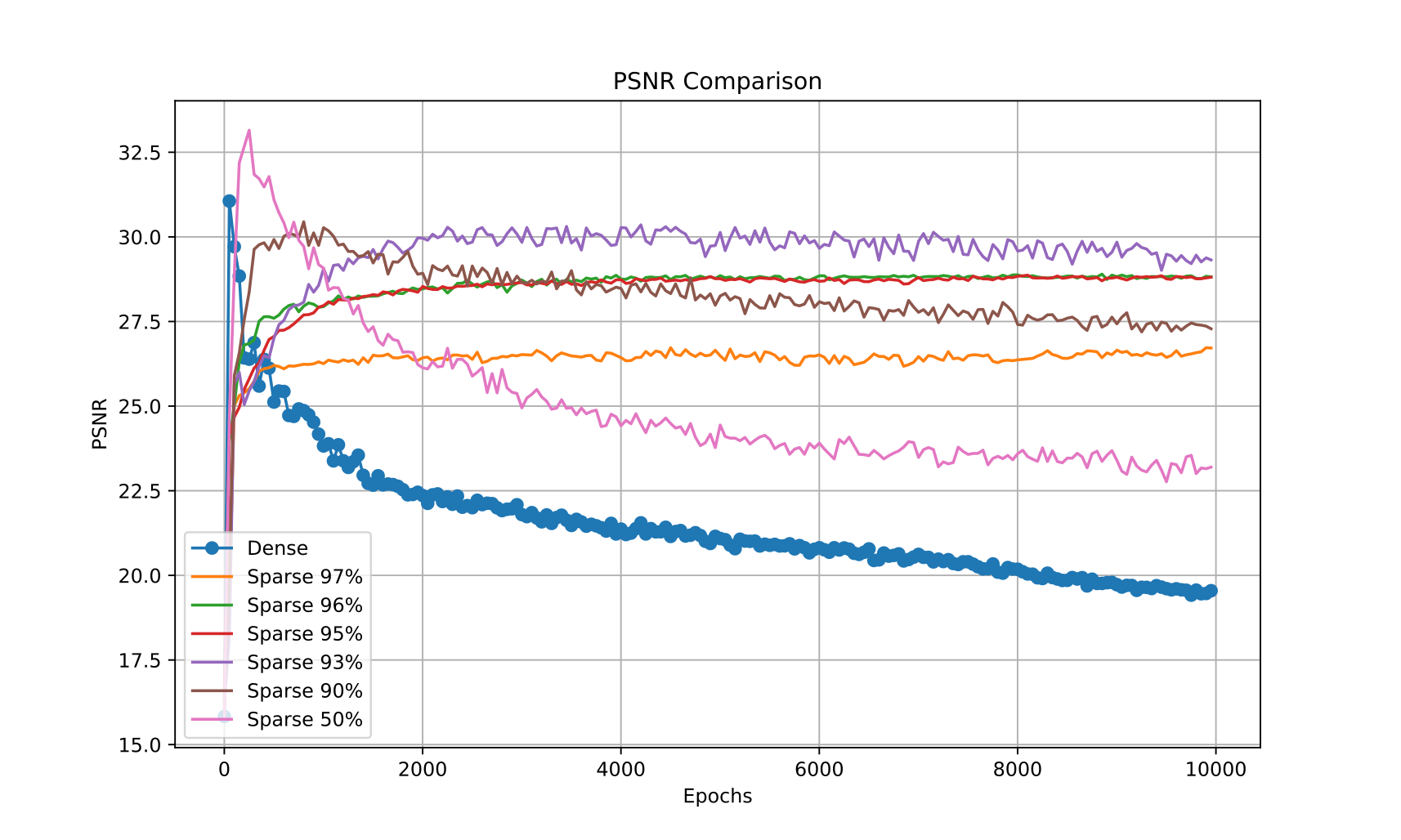}
        \subcaption{Cross + same undersampling: $P(\A_{2}(M_{4 \times}),\y_{2})$}
        \label{fig:self-2}
    \end{minipage}\quad
        \begin{minipage}{0.45\textwidth}
        \centering
        \includegraphics[width=\textwidth]{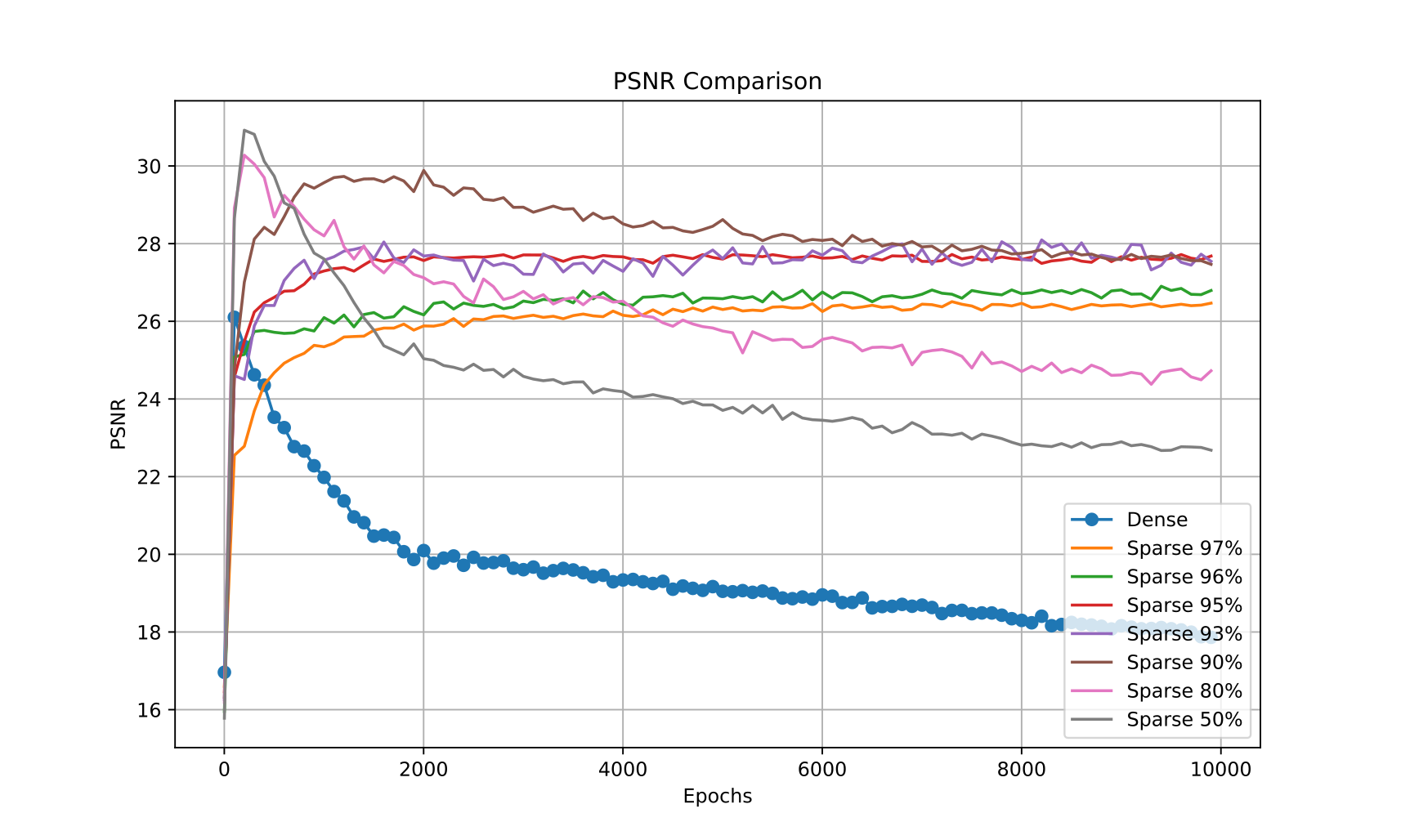}
        \subcaption{Cross + higher undersampling: $P(\A_{2}(M_{8 \times}),\y_{2})$}
        \label{fig:self-cross-2}
    \end{minipage}\quad    
        \begin{minipage}{0.45\textwidth}
        \centering
        \includegraphics[width=\textwidth]{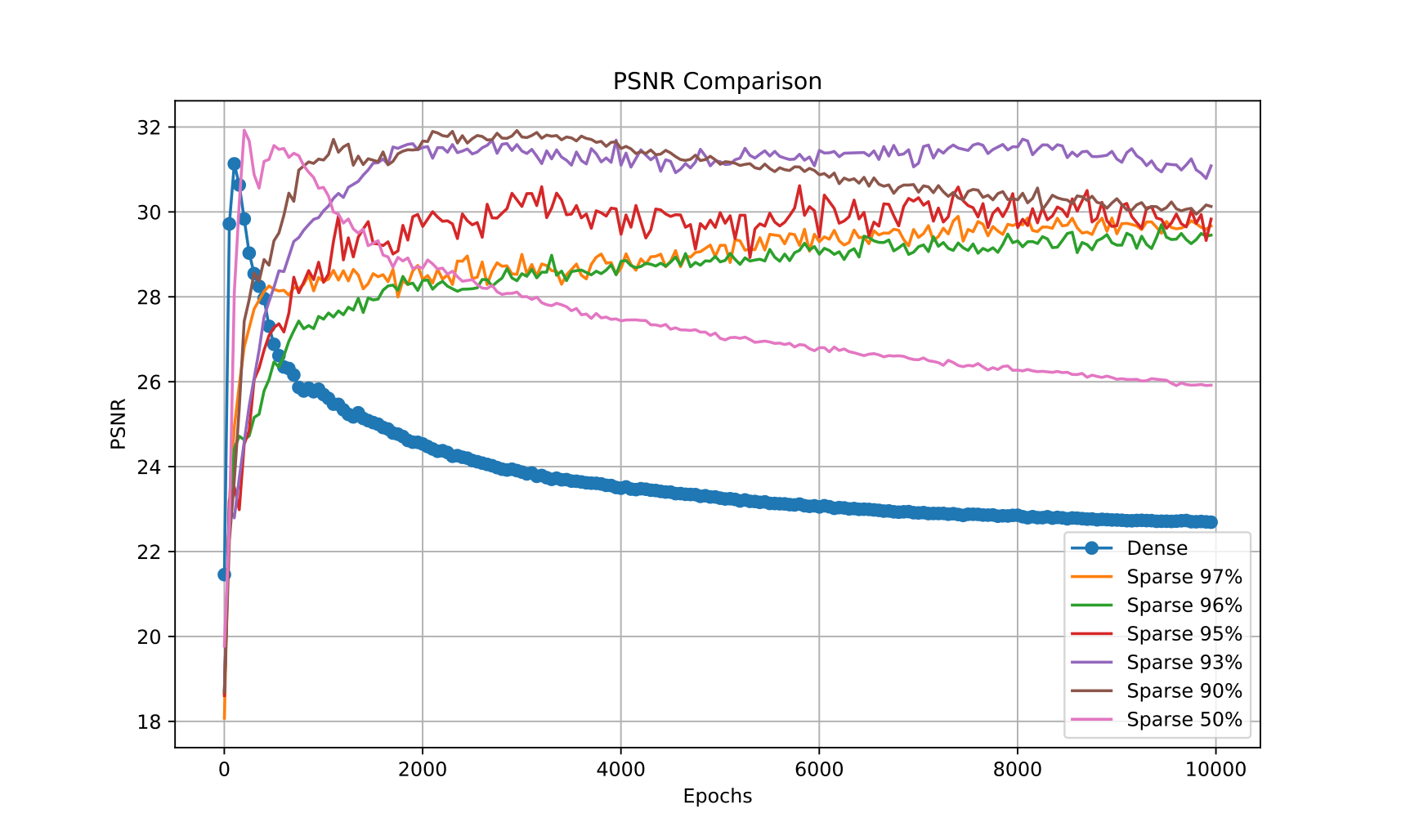}
        \subcaption{Cross + same undersampling: $P(\A_{3}(M_{4 \times}),\y_{3})$}
        \label{fig:self-3}
    \end{minipage}\quad
        \begin{minipage}{0.45\textwidth}
        \centering
        \includegraphics[width=\textwidth]{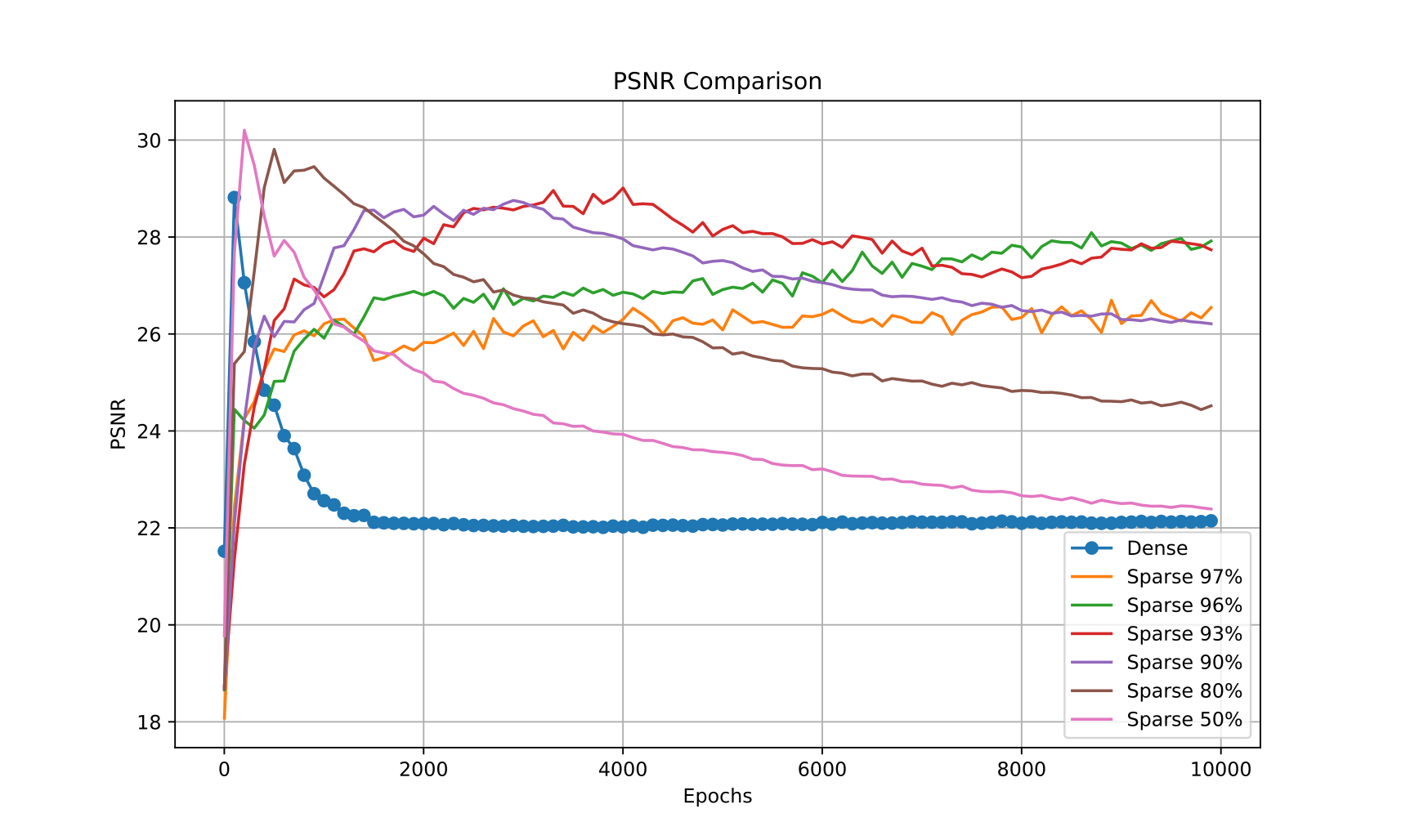}
        \subcaption{Cross + higher undersampling: $P(\A_{3}(M_{8 \times}),\y_{3})$}
        \label{fig:self-cross-3}
    \end{minipage}\quad
        \begin{minipage}{0.45\textwidth}
        \centering
        \includegraphics[width=\textwidth]{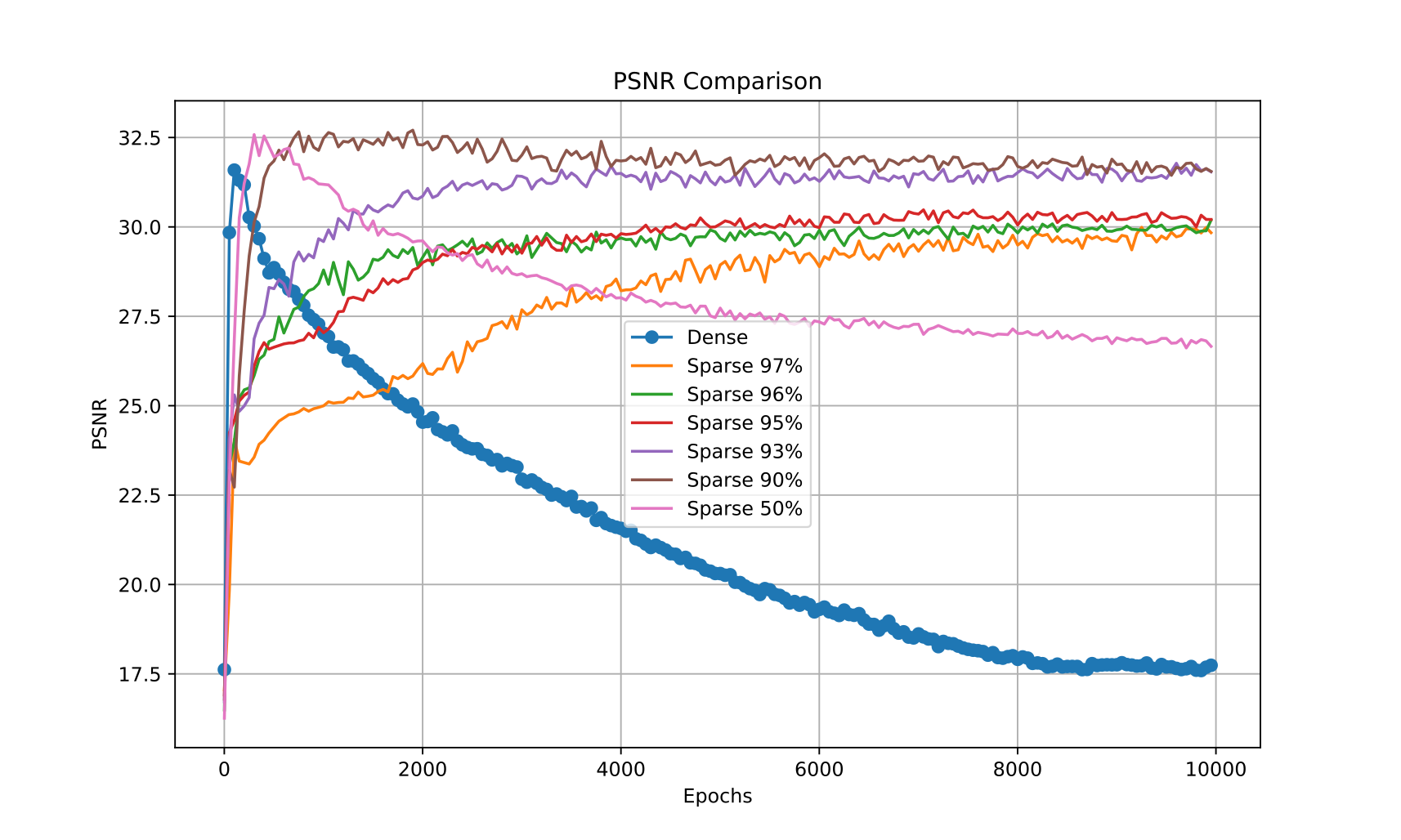}
        \subcaption{Cross + same undersampling: $P(\A_{4}(M_{4 \times}),\y_{4})$}
        \label{fig:self-4}
    \end{minipage}\quad
     \begin{minipage}{0.45\textwidth}
        \centering
        \includegraphics[width=\textwidth]{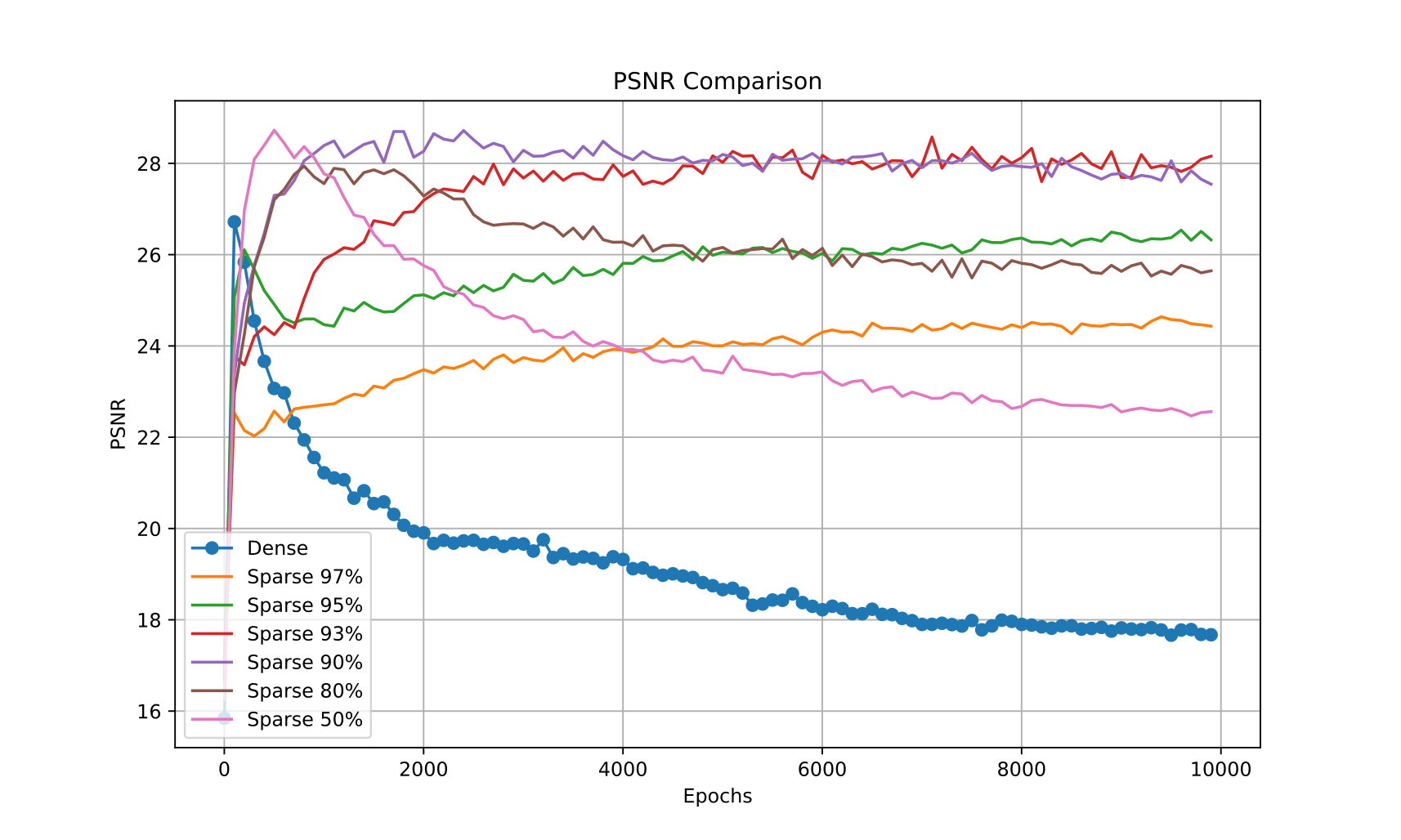}
        \subcaption{Cross + higher undersampling: $P(\A_{4}(M_{8 \times}),\y_{4})$}
        \label{fig:self-cross-4}
    \end{minipage}
 
        \caption{Performance of OES subnetworks for MRI reconstruction from $4\times$ (left column) and $8\times$ (right column) undersampled k-space measurements. In all the experiments, the OES network mask $m^{*}$ was learned from pair $(\A_{1}(M_{4 \times}),\y_{1})$. In Figure-a (self+ same undersampling), the subnetwork mask was used to reconstruct image from $(\A_{1}(M_{4 \times}),\y_{1})$. In Figure-b (self+ higher undersampling), mask was used to reconstruct from $(\A_{1}(M_{8 \times}),\y_{1})$ which has a higher undersampling. For Figures (c-h) (cross), the operator-measurement pair for image reconstruction were different from which the mask was learned $(\A_{1}(M_{4 \times}),\y_{1})$.}
    \label{mri-all-figs}
    
\end{figure*}

\subsection{Selectivity of Prior $\pr_{0}$ and Thresholding}

In our experiments, after the final convergence of our algorithm, we rank and threshold the value of $\pr$ to reach the desired sparsity level. An avid reader may ask the question that since the sparsity level is achieved through ranking and thresholding of the $\pr$ values, so is the selection of the prior $\pr_{0}$ important in getting a good ranking? Ideally speaking, the ranking should be based on the importance of the parameter in contributing to the loss. That means, a parameter $w_{1}$ is considered more important than parameter $w_{2}$ in the following case: if the objective when we fix $w_{1}=0$ (say $T(w_{1})$) would be more than the objective when we fix $w_{2}=0$ (say $T(w_{2})$). So in this case, if $T(w_{1}) \geq T(w_{2})$ then a proper ranking would imply $p(w_{1}) \geq p(w_{2})$. We observe that for $\lambda=1e-9$ choosing the prior $\pr_{0}$ to be the same as the desired sparsity level provides a good ranking that separates the important parameters from the non-important ones. Like in Figure-\ref{fig:kl} with $\lambda=1e-9$, when the prior $\pr_{0}$ is chosen to be the same as the desired sparsity level ($5\%$), we observe that most of the distribution is centered around $\pr_{0}$ with some values $\pr$ at 1. In the previous subsection, we discussed how the choice of $\lambda$ affects this distribution. In this section, we empirically show that the choice of $\pr_{0}$ is crucial in getting OES masks that are suitable for denoising. Fixing $\lambda=1e-9$, we perform a denoising experiment with different ranges of values where $\pr_{0}$ is as high as $0.5,0.8$ or as low as $0.05,0.03$. After the convergence of the loss we rank and threshold $95\%$ of the weights based on the value of $\pr$. We see that choosing a high value of $\pr_{0}$ outputs a mask that suffers from layer collapse and hence when further trained to denoise, completely breaks down. This is because, when $\pr_{0}$ is set to be high as $0.5$ or $0.8$, the distribution of $\pr$ across the network is centered at $0.5$ or $0.8$ respectively. Now when $95\%$ of the weights are thresholded after ranking, w.h.p all the weights in one layer are getting pruned because of improper ranking of $\pr$. This phenomenon of layer collapse seems to be avoided when the value of $\pr_{0}$ is chosen to be close to the pruning level. $\pr_{0}=0.03$ or  $\pr_{0}=0.05$ seem to give the same denoising performance when $95\%$ of the weights are pruned.

\section{Comparison with $L_1$ Regularization}
\label{l1reg}
In the image classification literature, supermasks have been used for obtaining a subnetwork by Bernoulli masking for example in 
 \citet{zhou2019deconstructing} without sparsity control. \citet{sreenivasan2022rare} used the $\ell_{1}$ regularization to control the sparsity of the mask and used iterative freezing at every epoch to reach the desired sparsity level. However, we observed that using $\ell_{1}$ regularization can't give a good ranking of $\pr$ based on the importance score. The optimal $\ell_{1}$ regularization coefficient can vary for different images but with KL regularization it is the same for all the images. 

Through extensive study, we find that in our experiments:
\begin{enumerate}
    \item That masking based on the ranking of the $\pr$ is sensitive to the choice of $\lambda$ when used in $L_{1}$ regularization like in \citet{sreenivasan2022rare}, i.e, the optimal $\lambda_{L_1}$ is not the same for two different images (Figure-\ref{fig:l1}). The best $\lambda_{L1}$ for the Pepper image can lead to a layer-collapse when used for the Flowers image for a desired sparsity level.
    \item Controlling the sparsity level through KL regularization leads to a better ranking in $\pr$ that can clearly separate out the important weights (Figure-\ref{fig:kl}). Using no (or extremely small) KL regularization, does not lead to a proper ranking of $\pr$ based on importance. The best ranking is obtained when the desired sparsity level is the same as the prior probability used in the KL. This alleviates the need to tune the prior $\pr_{0}$ and can be fixed to the target sparsity we want to achieve. We also demonstrate that using a severely different $\pr_{0}$ than the target sparsity can lead to improper ranking which leads to layer-collapse (Figure-\ref{fig:diff_p0}).
    \item The regularization strength $\lambda_{KL}$ is robust when KL regularization is used. We find $\lambda_{KL} =1e-9$ works for all the images unlike for $L_1$ regularization. We show this in Figure-\ref{fig:kl}. 
\end{enumerate}

Lastly, we want to emphasize that although learning masks by optimizing the Bernoulli probability $\pr$ has already been used in several works before, we show that using KL-based regularization gives us robust control over the sparsity we want to achieve. 
We compare our mask learning method with that of L1 regularization on $\pr$, which is known to promote sparsity in $\pr$. Sparsity in $\pr$ would ensure that the corresponding mask will be 0 with a very high probability. Although unlike in our formulation where we controlled the distribution $Ber(\pr)$ to be close to some prior distribution $Ber(\pr_{0})$, in $L_{1}$ regularization, we can only make $\pr$ sparse. 

\begin{align*}
\m^{*} &= C(\pr^{*})  \quad \text{such that} \quad \\
& \pr^{*} = \arg \min_{\pr} \mathbb{E}_{\m \sim Ber(\pr) }  \left[ || G(\p_{in} \circ \m,\z) - \y||_{2}^2 \right]  + \lambda \|  \pr \|_{1} 
\end{align*}

We solve the above optimization using the same algorithm as in Algorithm-1, but change the regularization term to $\|  \pr \|_{1}$ (scaled by $\lambda$) instead of the KL term. Here $\lambda$ would control the sparsity level of $\pr$, a higher $\lambda$ would ensure more $\pr$ is towards zero. Just like in OES, we also rank the $\pr$ values and threshold them at the desired level of sparsity. We observe through experiments that obtaining a reasonable network mask at initialization is sensitive for $L_{1}$ regularization. In Figure-\ref{fig:l1}, we see that the optimum $\lambda$ for the Pepper image and the Flowers image are different. For $\lambda=1e-9$, the mask produced by the Pepper image gives the best image representation, while for Flowers image, the best $\lambda=1e-8$. In fact for $\lambda=1e-9$, the Flower image seems to suffer from layer-collapse resulting in a constant image. This is unlike in the loss used for KL regularization, where $\lambda=1e-9$ performed consistently for all images. 

\subsection{Comparing KL, $\ell_1$ and Centered Mean Regularizaion}

We further investigate the use of $\ell_{1}$ regularization (given as $\|p \|_{1}$) and the centered mean regularizer (given as $|mean(p) - \frac{s}{d} |$), where we take $\frac{s}{d}=0.05$. 

When we minimize the objective with the centered mean regularizer and monitor the value of $mean(p)$, we see that starting from $p=0.5$ the loss can decrease to $p=0.05$ but not more, where it becomes stationary and does not change over 10 thousands of iterations (Figure-\ref{fig:centered_l1}). During this phase, this penalty has the same gradient as the $\ell_{1}$ norm regularizer. However, after $mean(p)$ reaches $0.05$, the mean $p$ becomes stationary and the loss seems to get stuck, although the penalty might behave differently than $\ell_{1}$. So, the overall effect of $\ell_{1}$ and the centered mean regularizer are similar.

Now, comparing $\ell_{1}$ regularizer to the KL regularizer, we notice across various experiments that $\ell_1$ regularization is less stable to the choice of the regularization strength. This is because $\ell_1$ regularizer encourages sparser solutions (for centered mean, (p-0.05) is sparse) than KL regularizer. This enforces a bulk of $p$ values to collapse on the same point. Hence the relative ranking gets lost due to this effect. 

For example, when the logits corresponding to the three regularizers are plotted in Figure-\ref{fig:histo_centered}, the logits in KL regularization seems to be more well spread than the $\ell_{1}$ and centered mean regularizer. When we look at the corresponding layerwise architecture in Figure-\ref{architecture}, we see that the middle layers are severely pruned by $\ell_{1}$ and centered mean regularization which may lead to layer collapse. We intentionally plot the sparsity percentage on the log scale to show the severity of this effect. 

So, based on this empirical observation, we think that sparser solutions may not be ideal for bringing the data misfit loss down (since loss of rank importance may lead to layer collapse). Furthermore, enforcing sparsity shrinks the search space of gradient descent, so it may be more likely to get stuck in local minima.

\subsection{On using Pointwise Regularization}

\begin{figure}[htbp]
    \centering
    \begin{minipage}[b]{0.45\textwidth}
        \centering
        \includegraphics[width=\textwidth]{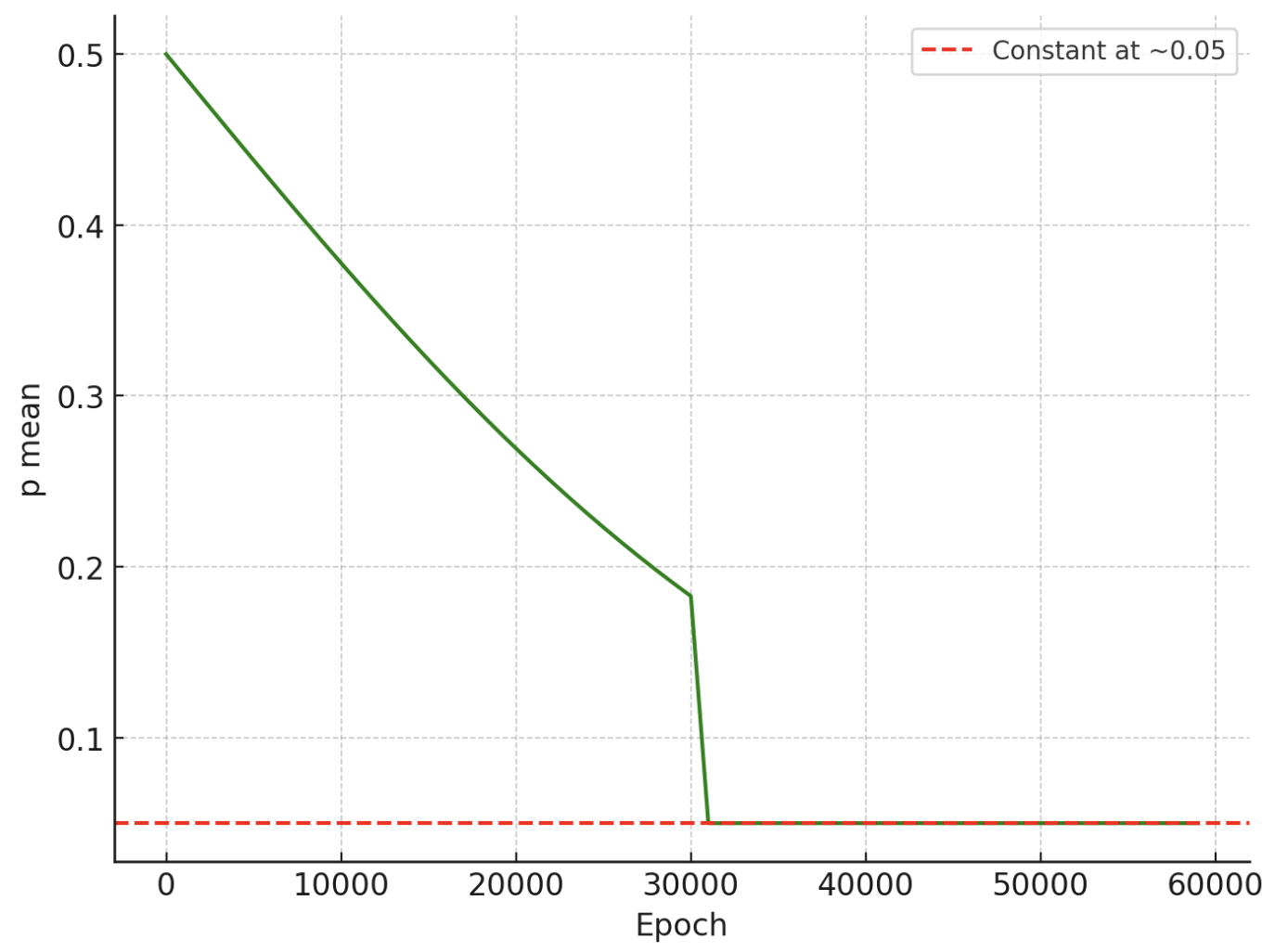}
        \caption{Mean of $p$ across various epochs when the regularization used is $|mean(p) - \frac{s}{d} |$. In this particular experiment, $\frac{s}{d} =0.05$.}
        \label{fig:centered_l1}
    \end{minipage}
    \hfill
    \begin{minipage}[b]{0.45\textwidth}
        \centering
        \includegraphics[width=\textwidth]{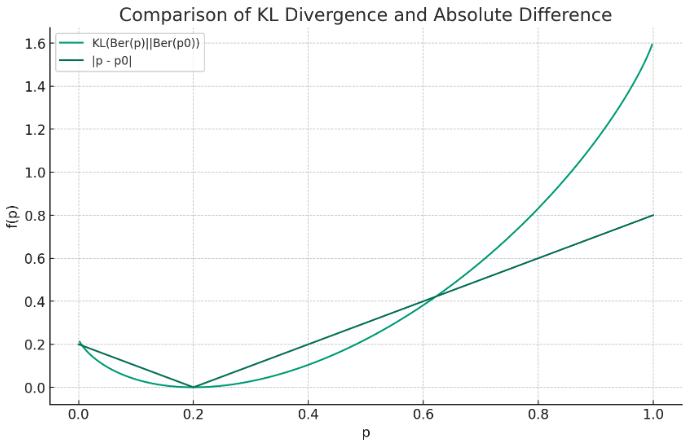}
        \caption{Comparison of KL regularization and pointwise centered $\ell_{1}$ regularization for a scalar value. Around the prior value $p_{0}$, the KL is much smoother than $\ell_{1}$ regularizer. $f'(p) = \log\left(\frac{\frac{p}{1-p}}{\frac{p_{0}}{1-p_{0}}}\right)$ for KL, which is very close to 0 when $p \in [p_{0}-\epsilon, p_{0}+\epsilon]$. However, for $\ell_{1}$  regularization, $f'(p) =1$ or $-1$ for all points except $p=p_{0}$.}
        \label{fig:klvsl1}
    \end{minipage}
\end{figure}

\begin{figure*}[!h]

    \begin{minipage}{0.5\textwidth}
        \centering
        \includegraphics[width=\textwidth]{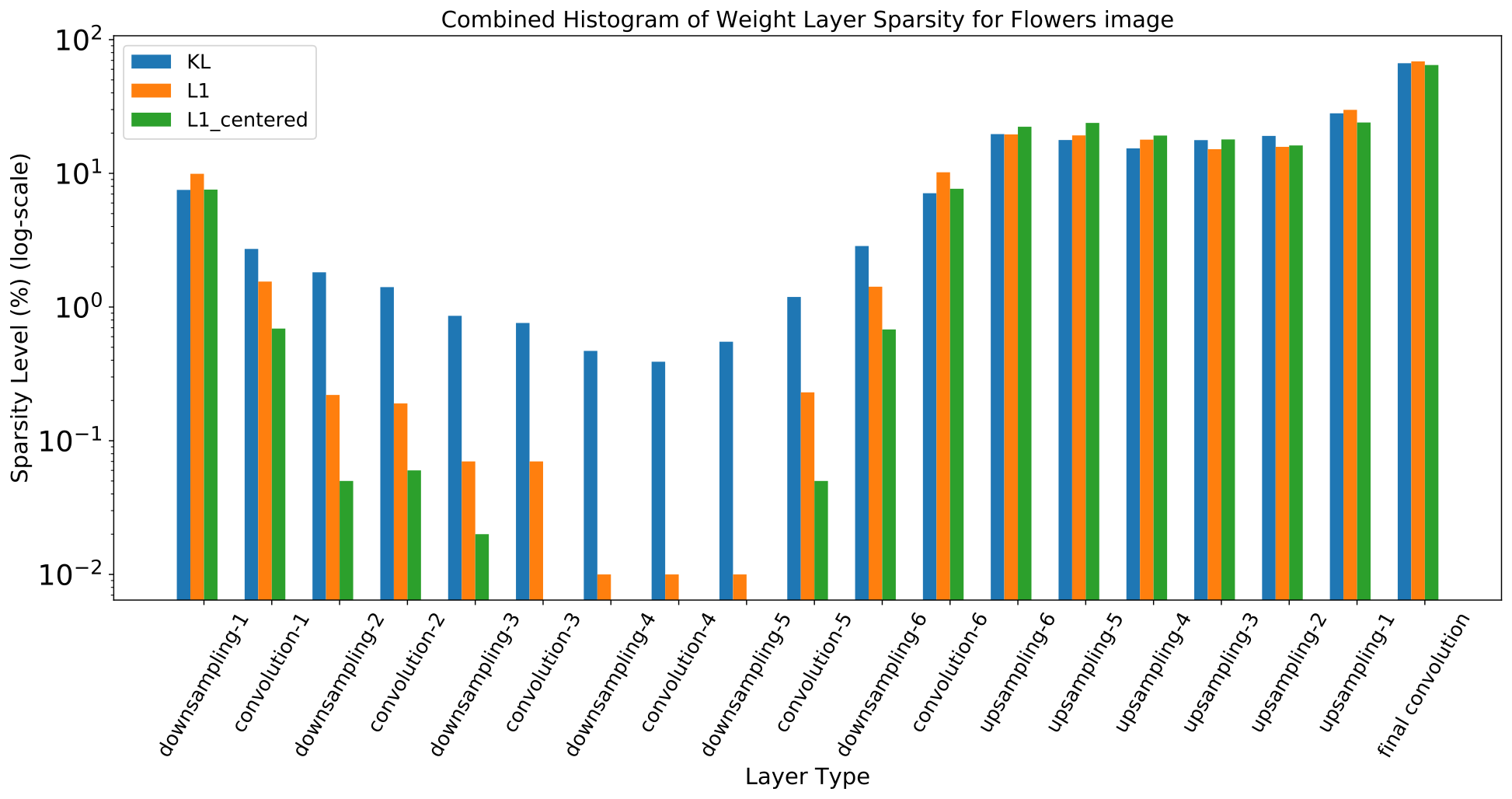}
        \subcaption{Flowers}
        \label{fig:Flowers}
    \end{minipage}\quad
        \begin{minipage}{0.5\textwidth}
        \centering
        \includegraphics[width=\textwidth]{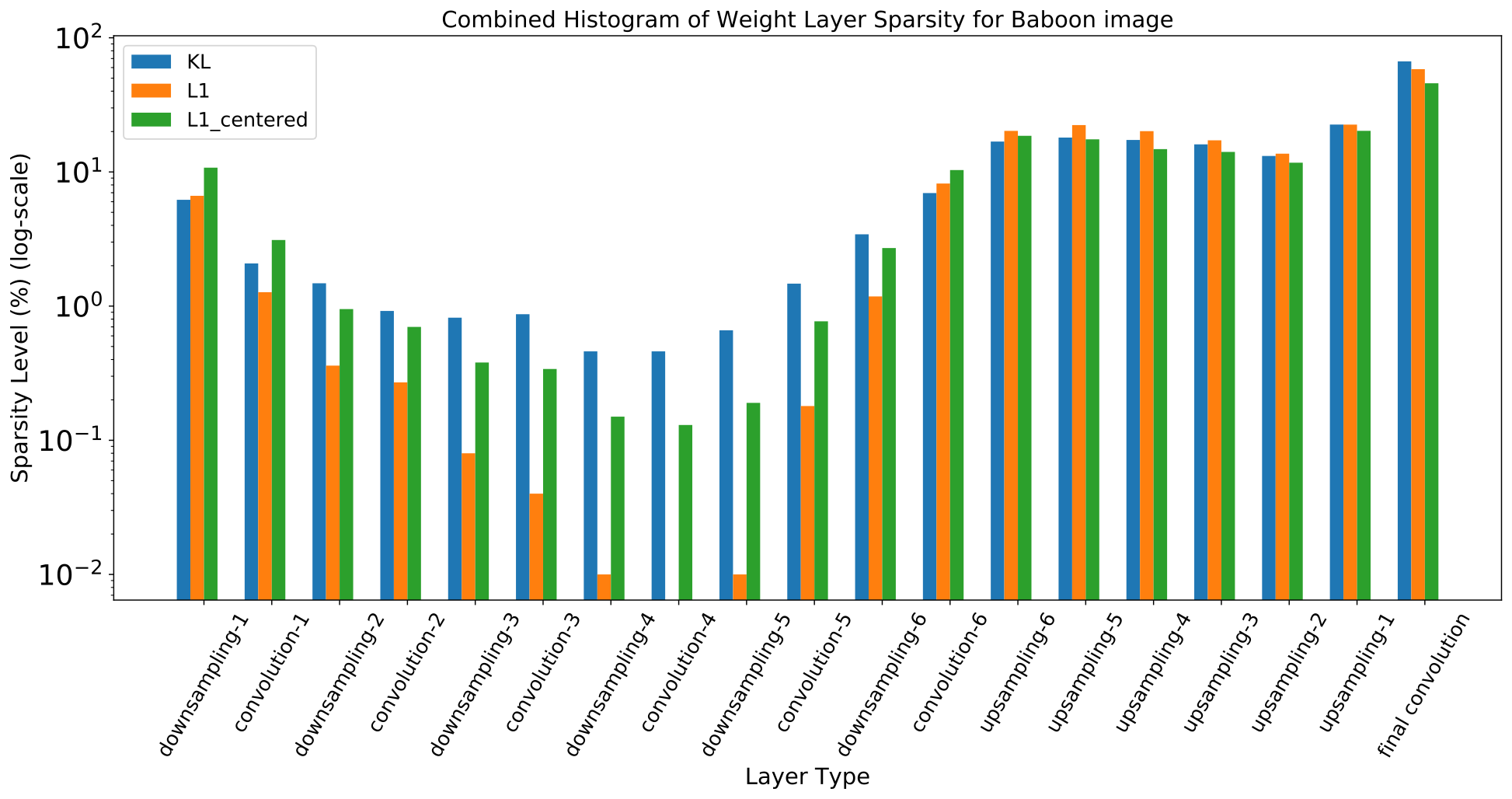}
        \subcaption{Baboon}
        \label{fig:Baboon}
    \end{minipage}\quad
       
        \caption{Layerwise architecure pruning (sparsity percentage in log-scale) by OES at initialziaiton using three different choices of regularization, KL, $\ell_{1}$ and centered $\ell_{1}$ for Baboon image and Flowers image in Set-14 dataset. Centered $\ell_{1}$ means the centered mean regularizer. }
    
    \label{architecture}
    
\end{figure*}

\begin{figure}
    \centering
    \includegraphics[width=0.7\textwidth]{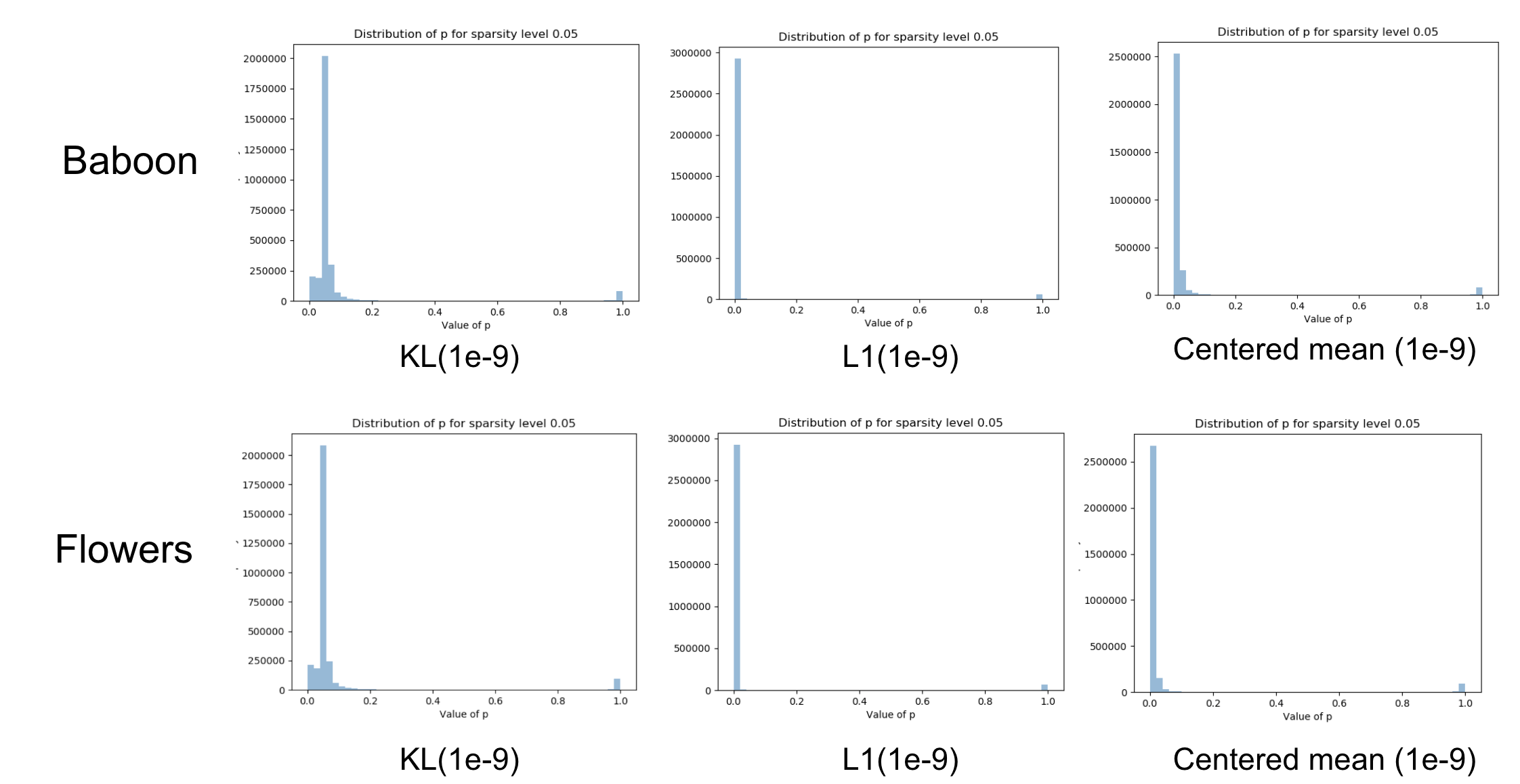}
    \caption{Histogram of logits of p when OES is ran across images with KL, $\ell_{1}$ and centered mean regularizer.  In our implementation we minimize $|\sum_{i} p_{i} - (\frac{s}{d}*numel(p)) |$, to both $\ell_{1}$ regularization and centered mean regularizer on the same scale.}
    \label{fig:histo_centered}
\end{figure}

\begin{figure}[h]
    \centering
    \includegraphics[width=0.5\textwidth]{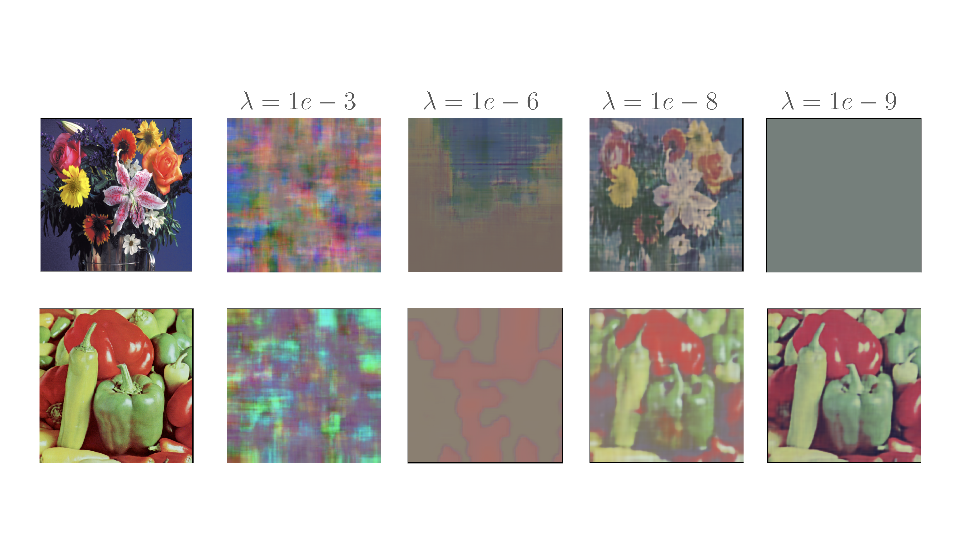}
    \caption{Sensitivity of hyperparameter $\lambda$ when mask is optimized by L1 regularization. Here, the best mask is obtained for different $\lambda$ for different images. For example, for the flower image, $\lambda = 1e-8$, is the best hyperparameter, but for the Pepper image $\lambda = 1e-9$. The masks here are $5\%$-sparse. }
    \label{fig:l1}
\end{figure}

From our experiments, we observed that using a pointwise regularizartion chosen with a proper regularization strength preserves the ranking. For KL regularization penalty, the ranking would remain preserved for very large range of moderate values of regularization coefficient $\lambda$, especially when compared to other pointwise regularization choices like $mean| p-p_{0}|$. This is because for KL regularization, the regularizer takes very low values around a large window $[p_{0}-\epsilon,p_{0} +\epsilon]$ (Figure-\ref{fig:klvsl1}). This is not true for linear pointwise regularizers such as $\ell_{1}$.  

We want to emphasize that pointwise regularization may allow the implementation of non-uniform prior $p_{0}$ across various weights in Unet. That's why we presented a more generic implementation, so if the user has some prior knowledge on what parameters are more important, they have the flexibility to modify the corresponding prior value. 

\section{Sensitivity of Masks to Weight Initialization}
\label{sense_weight_init}
\begin{figure}[h]
    \centering
    \includegraphics[width=0.4\textwidth]{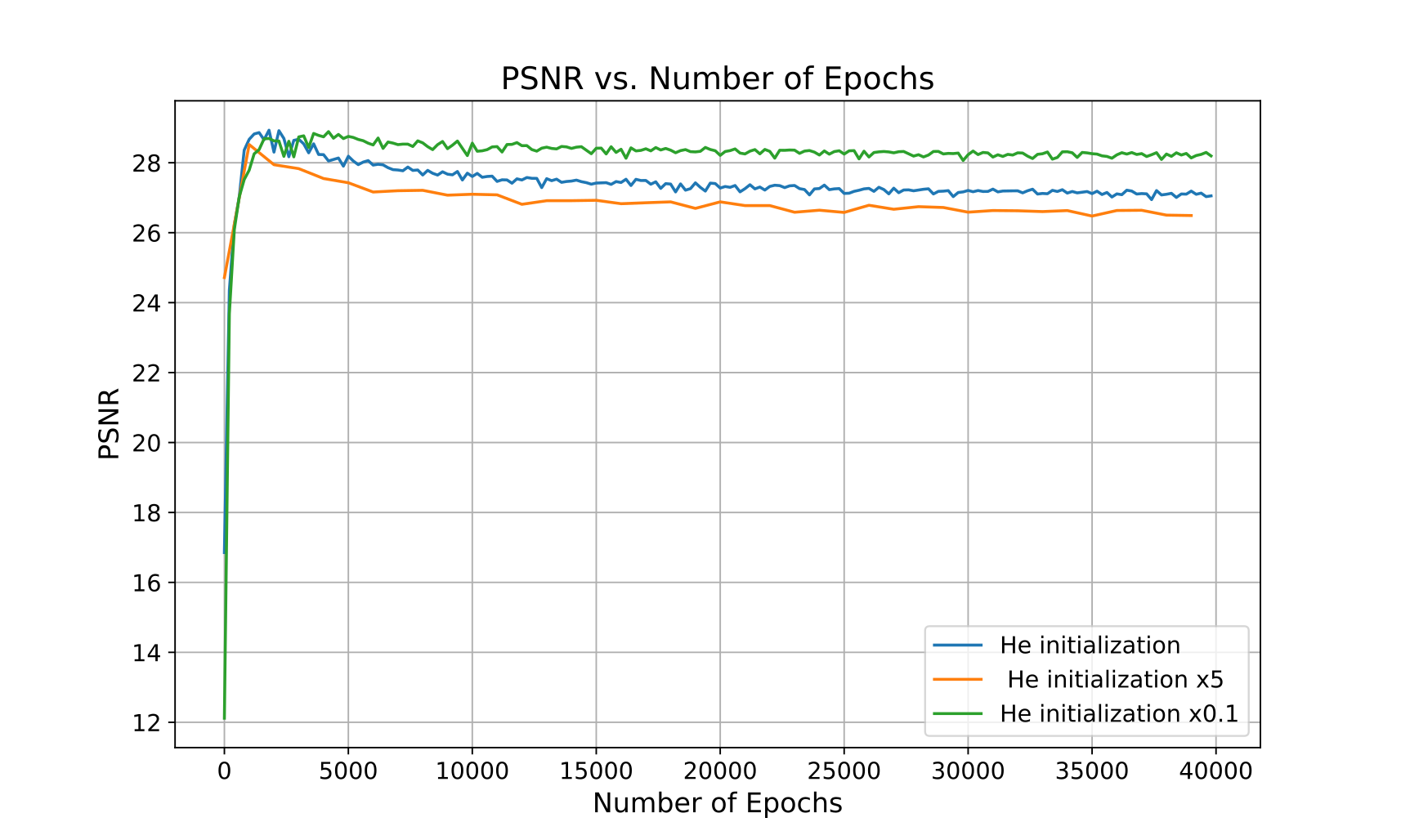}
    \caption{Comparison of denoising performance of OES masks at different initialization scales. }
    \label{fig:init_Scales}
\end{figure}

\subsection{Change in weight distribution (Uniform/ Normal initialization)}
\begin{figure*}[ht]
    \centering
    \begin{minipage}{0.4\textwidth}
        \centering
        \includegraphics[width=\textwidth]{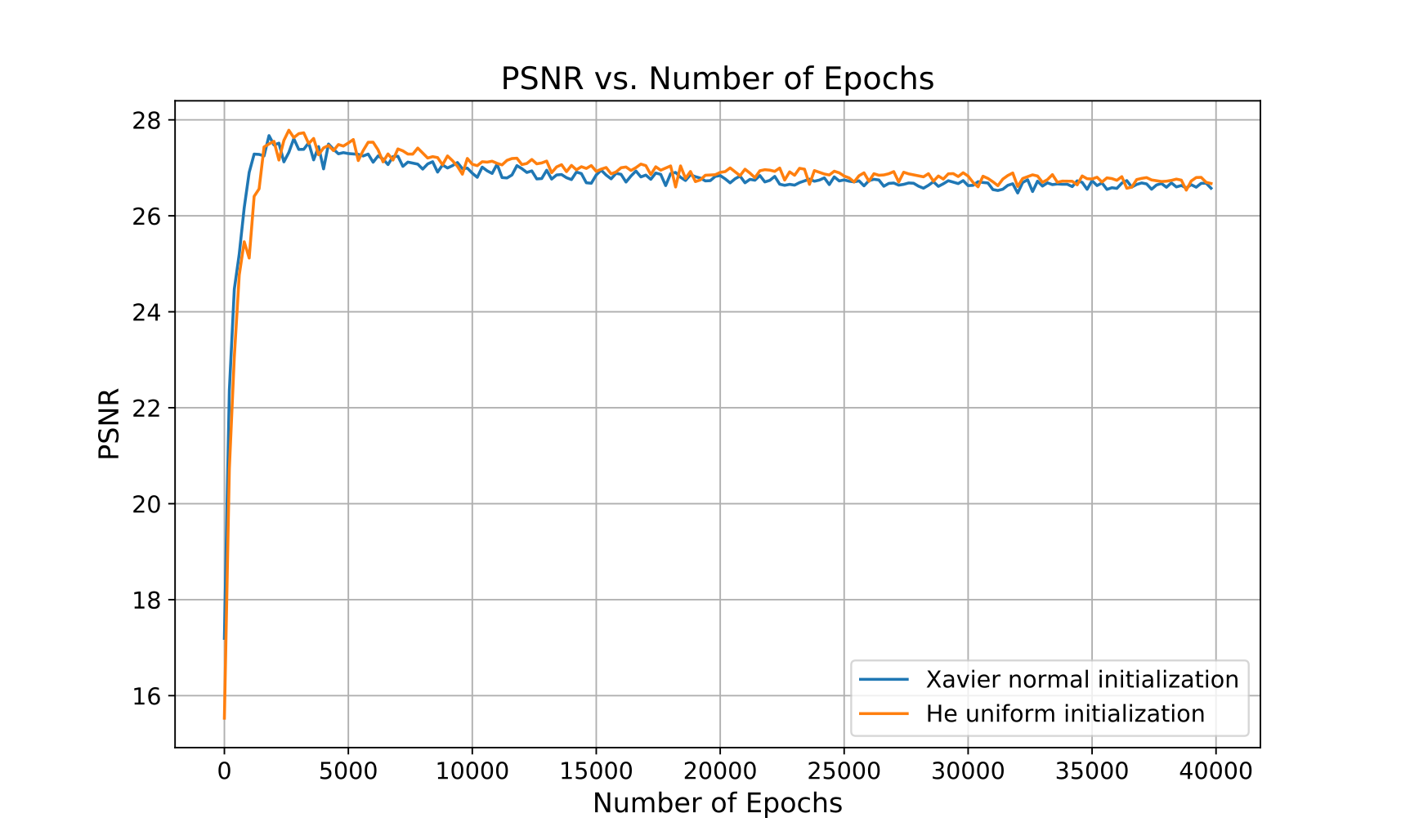}
        \subcaption{Flowers}
        \label{fig:bridge}
    \end{minipage}\quad
    \begin{minipage}{0.4\textwidth}
        \centering
        \includegraphics[width=\textwidth]{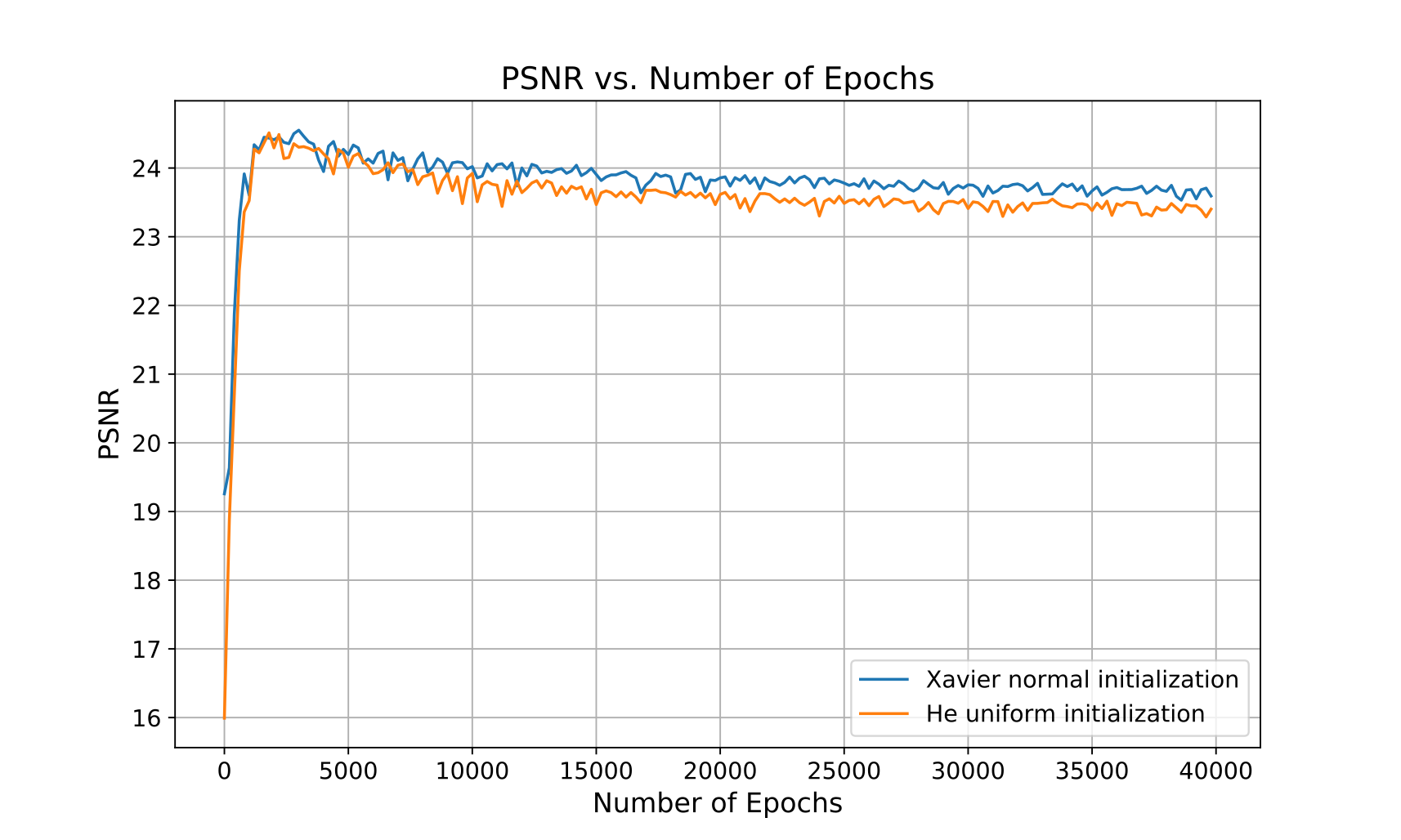}
        \subcaption{Comic}
        \label{fig:monarch}
    \end{minipage}\quad
    \begin{minipage}{0.4\textwidth}
        \centering
        \includegraphics[width=\textwidth]{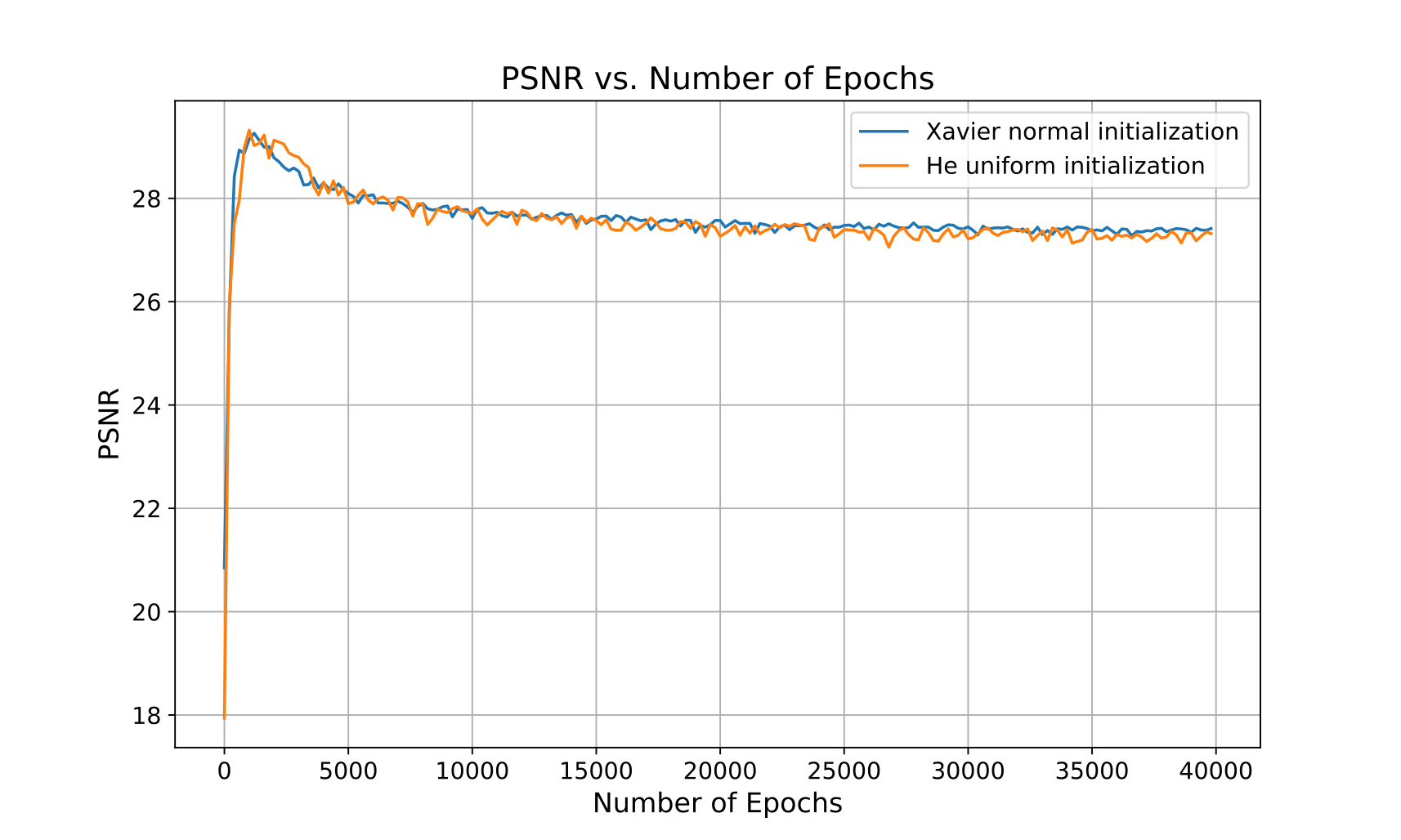}
        \subcaption{Lena}
        \label{fig:monarch}
    \end{minipage}\quad
    \begin{minipage}{0.4\textwidth}
        \centering
        \includegraphics[width=\textwidth]{arxiv_figures/distribution_compare/init_compare4.png}
        \subcaption{Barbara}
        \label{fig:monarch}
    \end{minipage}
    \caption{ Denoising performance of OES masks learned at He (uniform) initialization vs at Xavier initialization (Gaussian initialization). The initialization distriubtion does not seem to play a big role in learning the mask. }
    \label{init_oes}
\end{figure*}

Unlike other methods, OES learns the mask at initialization where the parameters are drawn from random Uniform distribution (He/Kaiming initializaiton) by Pytorch's default implementation. We check that the denoising performance of the mask is not affected by the distribution of the initialization. Changing the distribution to Normal Xavier distribution does not significantly affect the denoising performance of the OES masks. In Figure-\ref{init_oes}, we show that across 4 various images, the performance of masks learned either at uniform initialization or Normal Gaussian initialization remains the same. 

\subsection{Scale of Initiailization}

We observe that the scale of the initialization seems to affect the learned mask. So, in the experiment in Figure-\ref{fig:init_Scales}, we scale the original He initialization by $0.1$ and $5$ times respectively and then learn the mask by OES on the Pepper image. We observe that with a smaller scale of initialization, the learned OES mask seems to perform better in terms of denoising. On the contrary, masks learned at $5 \times$ initialization, seem to overfit slightly. 

\subsection{Initial Weights are at Early-Stopping Point}

Finding an early stopping point is a challenging task without the knowledge of the ground truth image. So performing IMP based pruning at the early-stopping point is too ambitious. In the following experiment, we show that even if we had an estimate of the early-stopping point, IMP based pruning may not be the best option. In Figure-\ref{fig:imp_early_oes}, we compare the denoising performance of three different masks: 1) IMP masks obtained at convergence on training with Pepper, 2) OES at initialization when Pepper is used in the loss function and 3) IMP masks obtained at the early stopping point also trained on Pepper (with the assumption that early stopping point is known). In this particular setting, when the target $\y$ is the corrupted Pepper image, we observe that IMP obtained at early-stopping point performs as well as OES. However, when the same 3 masks are used for denoising the Flower-image in Figure-\ref{fig:imp_early_oes}, we observe that the performance of IMP (at early stopping) degrades with respect to the OES mask.

\section{Pruning Deep Decoders by OES}
\label{deepdec}
\begin{figure}[!h]
    \centering
    \includegraphics[width=0.8\linewidth]{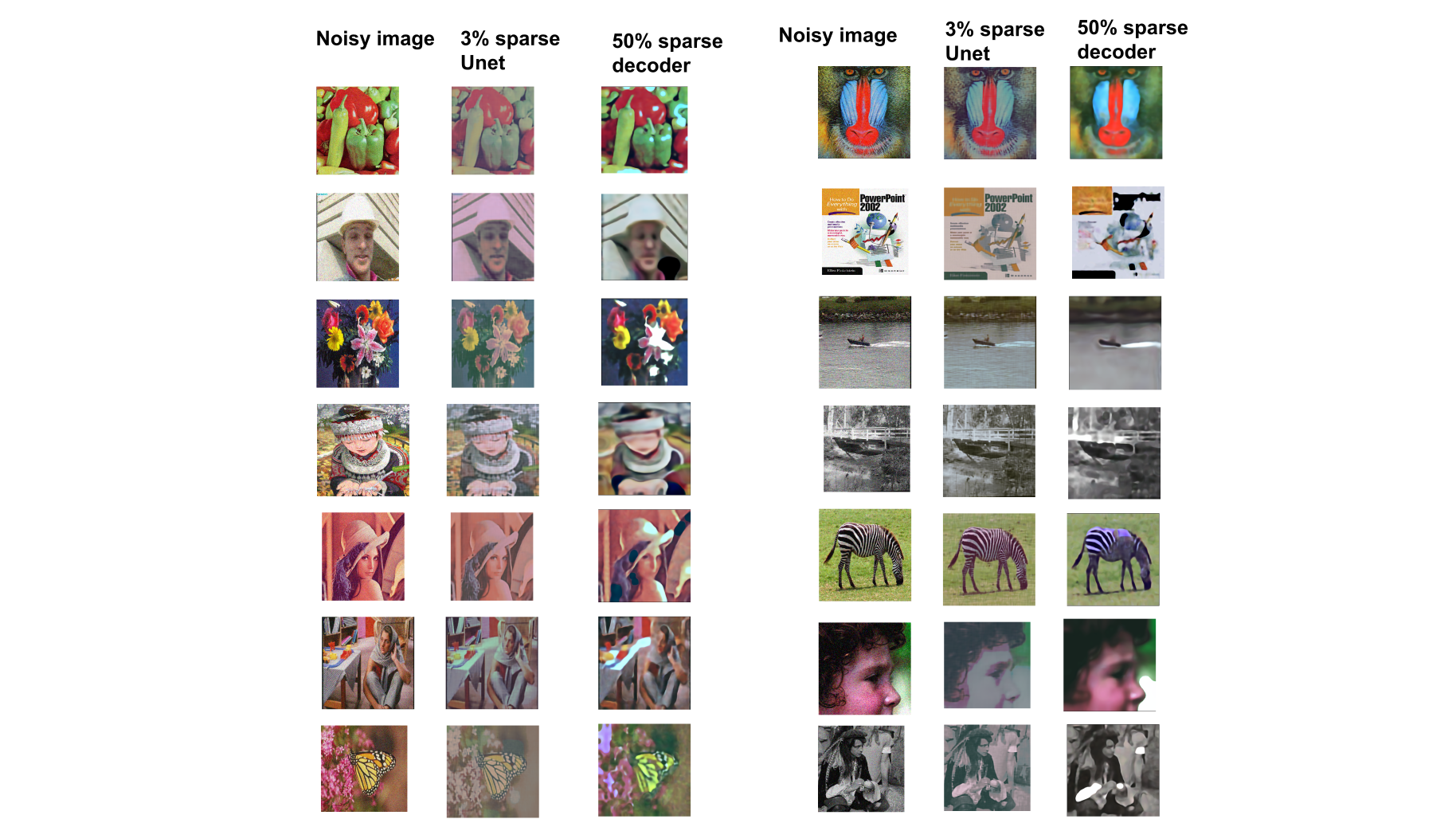}
    \caption{Comparison of OES masking in deep Unet vs in deep decoder. }
    \label{fig:decoder_mask}
\end{figure}

\begin{figure}[ht]
    \centering
    \includegraphics[width=1.0\textwidth]{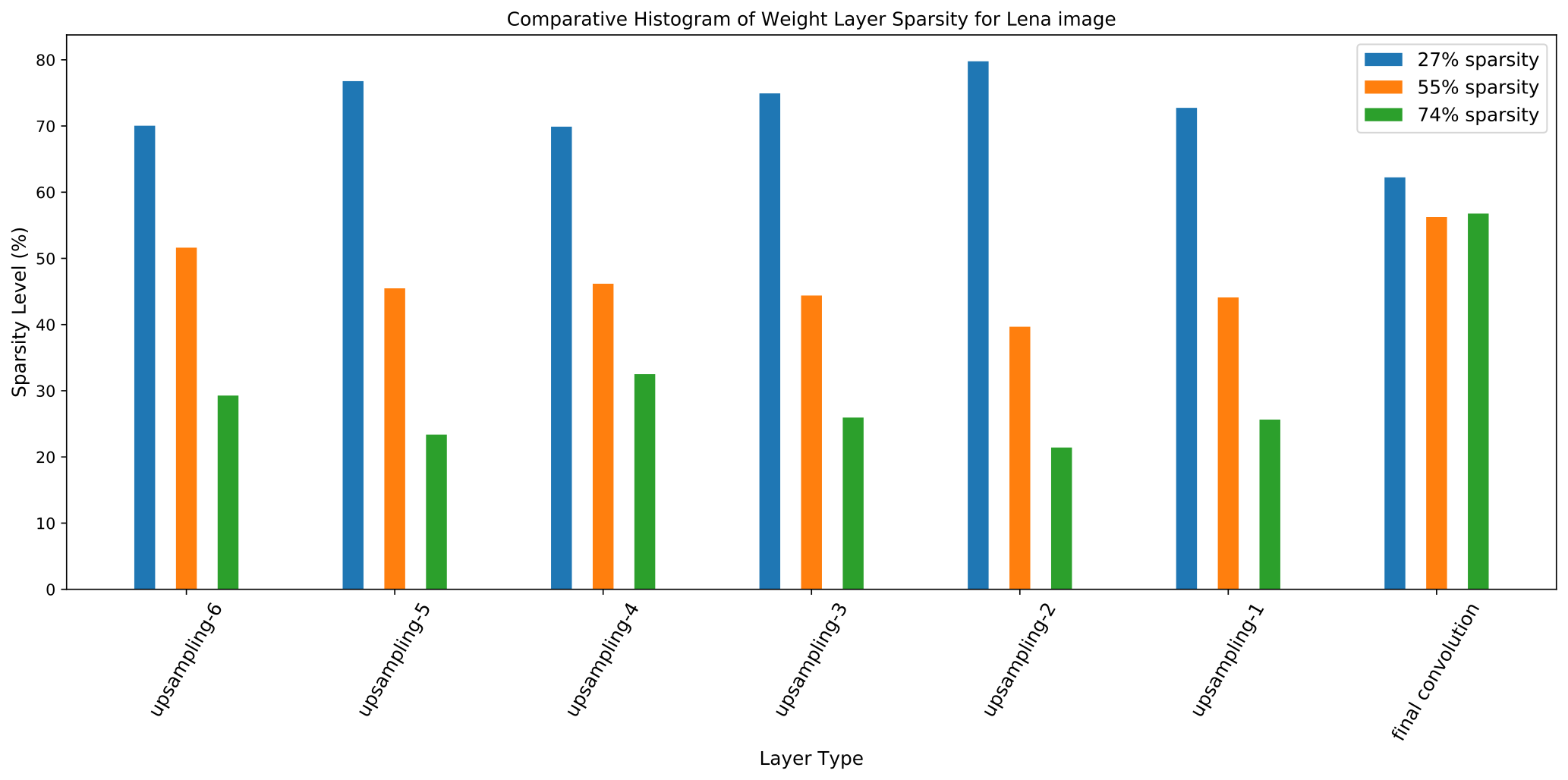}
    \caption{Layerwise pruning percentage for a deep decoder at various level sparsity levels.}
    \label{fig:decoder_histogram}
\end{figure}

In the manuscript, we showed that pruning a random-initialized Unet with 6 layers can give good starting point for further doing image reconstruction using just the masked subnetwork. Here we apply the OES methodology on the deep decoder architecture \citep{heckel2018deep}. Deep decoders only consist of upsampling operations as the source of getting low-frequency components in an image. In Figure-\ref{fig:decoder_mask}, we compare the images produced by the masked decoder at $55\%$ sparsity and compare it with images with masked Unet at $3\%$ sparsity, along with the corrupted versions. Since, the decoder is already underparameterized and acting as a natural image prior, masking at initialization seems to oversmoothen the image. There seems to be patches of bright and dark areas in the sparse decoder output when the parameters are just masked. On the contrary, for sparse Unets, the information lost due to oversmoothing is not that drastic. This is because decoders are already underparameterized, constraining the output space of decoder to have low frequency componenets. Further pruning by OES at masking leads to oversmoothing and loss of information. We observe that these sparse decoders are compressible by OES upto $74\%$ after which the output image is failed to produce due to layer collapse. In Table-\ref{mask_decoder_denoise}, we perform denoising using the masked decoder subnetworks for three different images. We observe that at $27\%$ sparsity level deep decoder performs comparably with it's original dense counterpart. However, for higher sparsity levels like $55\%$ and $74\%$, the performance starts to detoriate. When we observe the layer-wise sparsity pattern produced in deep decoder at 3 different sparsity levels in Figure-\ref{fig:decoder_histogram}, we observe that the first and last layer seems to the most important. The importance of parameter layers seems to be gradually diminishing towards the middle. This is similar to the finding in Figure-\ref{fig:subhist2} where the middle of the encoder and the decoder architecture was pruned the most. With the study of masking deep decoder, we motivate one fundamental question :
\paragraph{Q3}\label{q3}:  \textit{Should we start with a highly overparameterized model (dense Unet) to find a subnetwork or should we start with a smaller model (dense deep decoder) to find a subnetwork?}\\

Our experiments suggest that we should start with a highly overparameterized model. Starting with a smaller model, imposes the prior assumption that some architecture parts are not useful. However, this might not be always the case as we see that Sparse-DIP often outperforms deep decoders at the same level of sparsity. In all our experiments, we fix $\lambda=1e-9$ and fix prior $\pr_{0} =0.5$ for all the weights in all layers. A realization of the decoder can be obtained by fixing $\pr_{0}=1$ for the decoder part and $\pr_{0}=0$ for the encoder part. 

\begin{table}[ht]
\centering
\small
\caption{Comparison of deep decoder performance across various pruning levels.}
\begin{tabular}{|c|c|c|c|c|}
\hline
& Deep Decoder & Sparse Decoder ($27\%$) & Sparse Decoder ($55\%$) & Sparse Decoder ($74\%$) \\ \hline
Pepper & 27.01 & 27.06 & 26.17 & 26.35\\ \hline
Lena & 26.80 & 26.94 & 25.35 &  25.15\\ \hline
Barbara & 25.30 & 25.14 & 24.58 & 24.42\\ \hline
\end{tabular}
\label{mask_decoder_denoise}
\end{table}

\begin{figure}[h]
    \centering
    \includegraphics[width=0.8\textwidth]{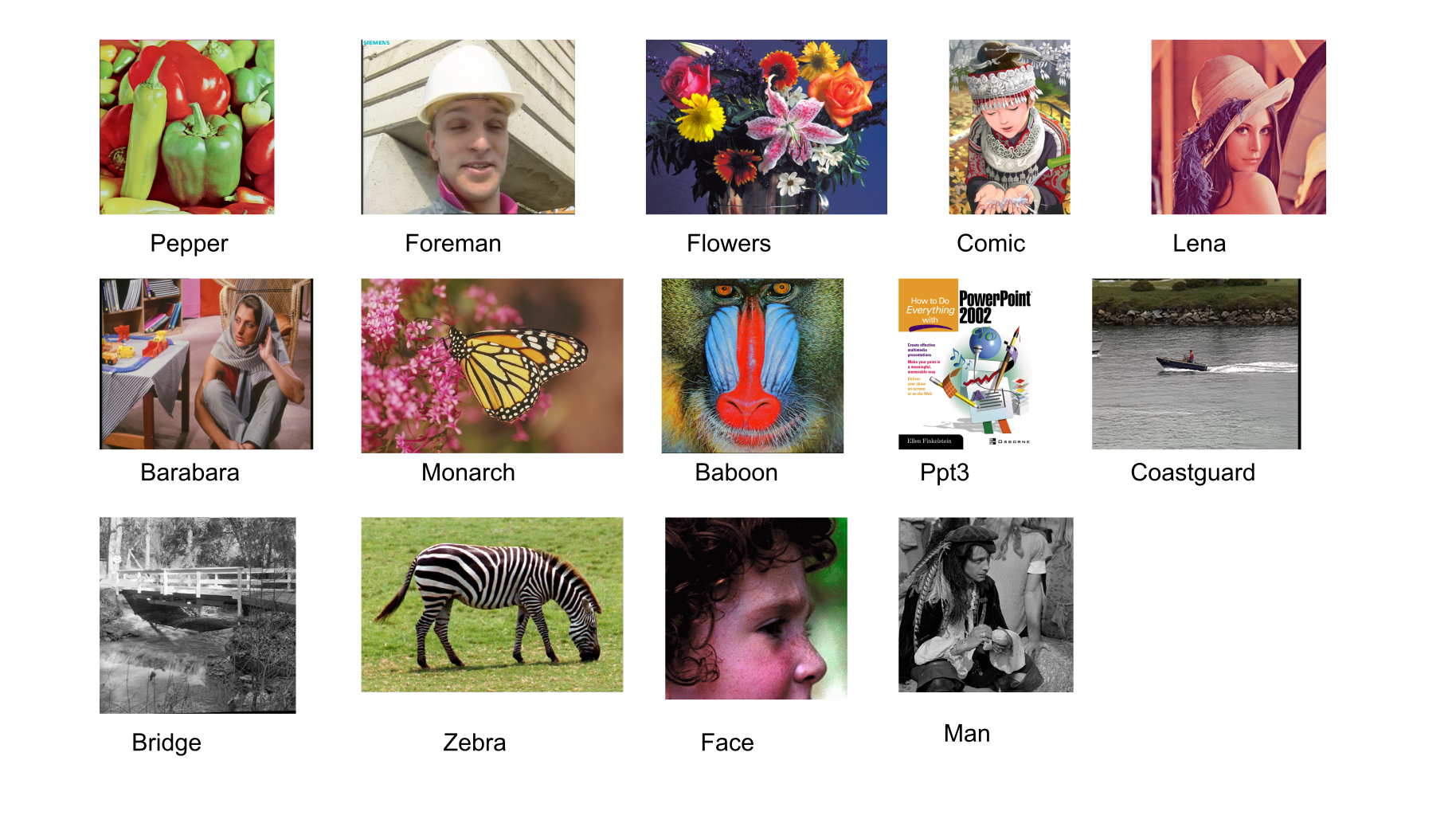}
    \caption{Set-14 dataset images used in this paper.}
    \label{fig:set14all}
\end{figure}

\begin{figure}[h]
    \centering
    \includegraphics[width=0.8\textwidth]{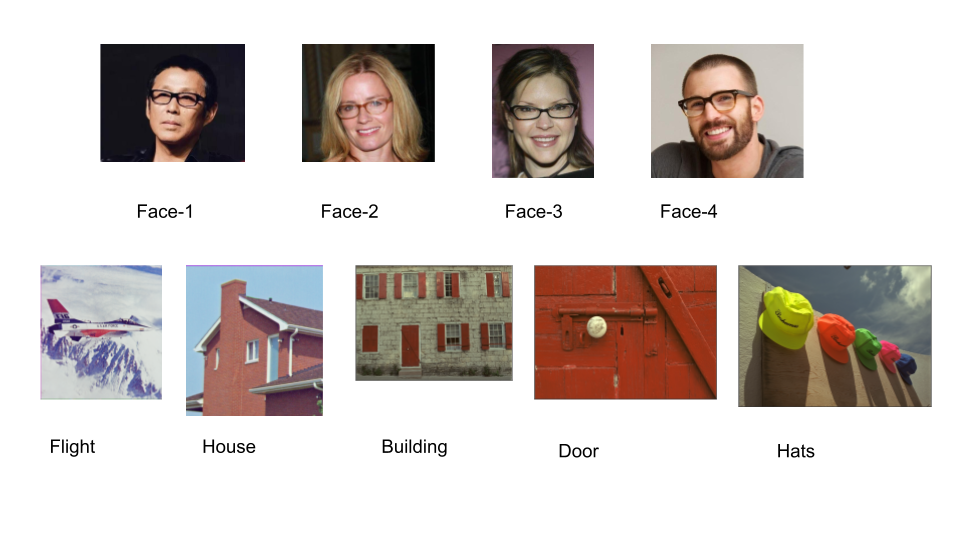}
    \caption{Images in face and standard dataset used in this paper.}
    \label{fig:face_standard_all}
\end{figure}


\section{Comparison of Pruning in Image Classification and Image Reconstruction}
\label{comp_class_rec}
In Table-\ref{comparingclassvsrec}, we discuss the many differences in pruning networks for image classification and image reconstruction. Pruning for image classification tasks, dates back to the early 90's with a recent surge of works being done after the popularity of Lottery Ticket Hypothesis \citep{frankle2018lottery}. To the best of our knowledge, \citet{wu2023chasing}, is the first work to propose pruning network for image reconstruction tasks. In our work, we show the drawbacks of just applying LTH on image reconstruction tasks and propose OES that mitigates the problem. Our work also shows the Strong Lottery Ticket Hypothesis in image reconstruction networks for the first time. In Figure-\ref{fig:mask_rep}, we highlight the representation capability of OES. With no mask and or all masked, we get two extremes. In the middle ground, we can approximate any image by just masking. In Figure-\ref{fig:transfer_flow}, we show the progression of transferred subnetwork through intermediate epochs, showing that the subnetwork output image is always constrained in the manifold of image priors. The images we used in this paper are shown in Figure-\ref{fig:set14all} and Figure-\ref{fig:face_standard_all}.
\begin{table}[htbp]
\centering
\caption{Pruning for Image Classification vs Image Reconstruction}
\label{tab:image_pruning_comparison}
\begin{tabular}{|m{0.11\textwidth}|m{0.35\textwidth}|m{0.35\textwidth}|}
\hline
\textbf{Criterion} & \textbf{Image Classification} & \textbf{Image Reconstruction (DIP)} \\ \hline
Task & The pruned network is learned based on ERM loss over a set of given image/label pairs. Usually, $0$-$1$ loss is used. & Pruned network is learned over a single image instance (extreme data-diet) and regression loss (MSE) loss is used. \\ \hline
Validity of LTH & LTH is essential to obtain matching subnetworks at non-trivial sparsities. & LTH is suboptimal as network overfits to image noise at convergence (post-training). \\ \hline
Transferability & Transferability is difficult to attain. \cite{mehta2019sparse} & Reasonable transferability can be attained. Better transferability can be achieved through OES when compared to LTH. \\ \hline
Performance of matching subnetworks & Matching subnetworks can attain almost the same level of test accuracy (or slightly higher in intermediate sparsity levels \citep{jin2022pruning}. \textit{Sparsity may not be necessary to get good generalization.} & Sparse subnetworks alleviate the problem of overfitting. \textit{Sparsity is necessary to alleviate overfitting.}   \\ \hline
Strong Lottery Ticket Hypothesis & \citet{ramanujan2020s} showed that masking a wide Resnet50 can give similar test accuracy as training a Resnet-34 on Imagenet classification. \citet{malach2020proving} proved that if a ReLU fully-connected neural network with depth $d$ and width $n$ can fit a target by normal training, then masking a Relu network at depth $2d$ and polynomial width can approximate the same performance. For CNN's \citep{da2021proving}, the width required was logarithmic in depth and number of parameters.    &  \textit{Our work is the first to show that Strong Lottery Ticket Hypothesis can also be observed for image reconstruction tasks.} We see that the network output can give low frequency representation of the clean image by just masking the network parameters by OES. Proving it for image reconstruction problems will be future work. \\ \hline
\end{tabular}
\label{comparingclassvsrec}
\end{table}

\end{document}